\documentclass[journal,twoside]{IEEEtran}

\ifCLASSINFOpdf
  \usepackage[pdftex]{graphicx}
  \graphicspath{{figures/}}
  \DeclareGraphicsExtensions{.pdf,.jpeg,.png}
\else
  \usepackage[dvips]{graphicx}
  \graphicspath{{figures/}}
  \DeclareGraphicsExtensions{.eps}
\fi

\usepackage{diagbox}

\usepackage{cite}

\usepackage{amsmath}
\usepackage{amsfonts}

\usepackage{array}

\ifCLASSOPTIONcompsoc
  \usepackage[caption=false,font=normalsize,labelfont=sf,textfont=sf]{subfig}
\else
  \usepackage[caption=false,font=footnotesize]{subfig}
\fi
\usepackage{float}

\begin{document}
\title{A Directionally Selective Small Target Motion Detecting Visual Neural Network in Cluttered Backgrounds}

\author{Hongxin Wang, Jigen Peng, Shigang Yue, \IEEEmembership{Senior Member,~IEEE} \thanks{Manuscript received April 5, 2018; revised July 6, 2018 and July 18, 2018; accepted August 29, 2018. This work was supported in part by EU FP7 Project HAZCEPT under Grant 318907, in part by HORIZON 2020 project STEP2DYNA under Grant 691154, and in part by the National Natural Science Foundation of China under Grant 11771347. This paper is recommended by Associate Editor Q. Meng. (Hongxin Wang and Jigen Peng contributed equally to this work.) (\emph{Corresponding author: Shigang Yue.})} 
	\thanks{H. Wang and S. Yue are with the Computational Intelligence Lab (CIL), School of Computer Science, University of Lincoln, Lincoln LN6 7TS, U.K. (email: syue@lincoln.ac.uk).}
	\thanks{J. Peng is with School of Mathematics and Information Science, Guangzhou University, Guangzhou, 510006, China (email: jgpeng@gzhu.edu.cn).}
	\thanks{This paper has supplementary downloadable material available at
		http://ieeexplore.ieee.org, provided by the author.}
	\thanks{Color versions of one or more of the figures in this paper are available
		online at http://ieeexplore.ieee.org.}
    \thanks{Digital Object Identifier 10.1109/TCYB.2018.2869384}}
   

\markboth{IEEE TRANSACTIONS ON Cybernetics}
{Wang \MakeLowercase{\textit{et al.}}: Directionally Selective Small Target Motion Detecting Visual Neural Network in Cluttered Backgrounds}

%



\maketitle

\begin{abstract}
Discriminating targets moving against a cluttered background is a huge challenge, let alone detecting a target as small as one or a few pixels and tracking it in flight. In the insect's visual system, a class of specific neurons, called small target motion detectors (STMDs), have been identified as showing exquisite selectivity for small target motion. Some of the STMDs have also demonstrated direction selectivity which means these STMDs respond strongly only to their preferred motion direction. Direction selectivity is an important property of these STMD neurons which could contribute to tracking small targets such as mates in flight. However, little has been done on systematically modeling these directionally selective STMD neurons. In this paper, we propose a directionally selective STMD-based neural network for small target detection in a cluttered background. In the proposed neural network, a new correlation mechanism is introduced for direction selectivity via correlating signals relayed from two pixels. Then, a lateral inhibition mechanism is implemented on the spatial field for size selectivity of the STMD neurons. Finally, a population vector algorithm is used to encode motion direction of small targets. Extensive experiments showed that the proposed neural network not only is in accord with current biological findings, i.e., showing directional preferences, but also worked reliably in detecting small targets against cluttered backgrounds.
\end{abstract}


%
\IEEEpeerreviewmaketitle

\section{Introduction}
\IEEEPARstart{I}{ntelligent} robots have shown great potential in reshaping human life in the future. However, artificial visual systems so far are still struggling to provide a robot with the required capacity to respond to the dynamic visual world in real-time, like many animal species do. Among many visual functionalities, detecting small moving targets is one of the most important abilities for many animal species, e.g.,  finding mates in the distance, and it is also critical for a robot to track small targets in a cluttered background.

Small target motion detection in visual cluttered backgrounds is always considered as a challenging problem for artificial visual systems. The difficulty is reflected in two aspects: first, when a target is far away from the observer, it always appears as a small dim speckle whose size may vary from one pixel to a few pixels in the field of view. In this size, shape, color and texture information cannot be used for target detection. Second, small targets are often buried in cluttered backgrounds and difficult to separate from noise. In addition, ego-motion may bring in further difficulties to small target motion detection.

Nature has provided a rich source of inspiration for small target motion detection. Detecting small targets in naturally cluttered backgrounds is critical for many insect species to search for mates or track prey. As the result of millions of years of evolution, the small target motion detection visual systems in insects are both efficient and reliable \cite{yue2006collision,nordstrom2006insect}. For example, dragonflies can pursue small flying insects with successful capture rates as high as $97\%$ relying on their well evolved vision system \cite{olberg2000prey}. Compared to the visual systems of primate animals, insects' visual systems achieve amazing capability using relatively simple structures and a small number of neurons. Insects' visual pathways are practical models for designing artificial vision systems for small target motion detection.

In the insect's visual system, a class of specific neurons, called small target motion detectors (STMDs), have been identified as showing exquisite selectivity for small targets (size selectivity) \cite{nordstrom2006insect,nordstrom2012neural,nordstrom2006small}. To be more precise, the STMD neurons give peak responses to targets subtending $1^{\circ}-3^{\circ}$ of the viusal region, with no response to larger bars (typically $> 10^{\circ}$) or to wide-field grating stimuli. In addition, some STMD neurons are directionally selective (direction selectivity) \cite{o1993feature,barnett2007retinotopic}. They respond strongly to small target motion oriented along a preferred direction, but show weak or no, even fully opponent response to null-direction motion. Null direction is $180^{\circ}$ from the preferred direction. Although the postsynaptic pathways of the STMD neurons are still under investigation \cite{gonzalez2013eight}, it is clear that knowing the small target motion and its direction at the same time is an advantage in tasks such as tracking mates or intercepting prey.

The electrophysiological knowledge about the STMD neurons and their afferent pathways revealed in the past few decades makes it possible to propose quantitative STMD models, however, little has been done on systematically modeling these directionally selective STMD neurons. As pioneers, Wiederman \emph{et al.} \cite{wiederman2008model} developed elementary small target motion detector (ESTMD) to account for size selectivity of the STMD neurons. The ESTMD showed strong responses to small target motion, but much weaker or even no responses to wide-field motion. However, it did not consider direction selectivity and showed no different responses to small target motion oriented along different directions. Wiederman and O'Carroll \cite{wiederman2013biologically} mentioned that two hybrid models, i.e., elementary motion detector (EMD)-ESTMD and ESTMD-EMD, could exhibit both size and direction selectivities. In the further research \cite{bagheri2017autonomous,bagheri2017performance,bagheri2015properties}, these two models are successfully used  for target tracking. Although direction selectivity was noted in these models, the direction selectivity in an STMD model should be systematically investigated.
\begin{enumerate}
	\item The existing STMD-based models, including ESTMD\cite{wiederman2008model}, EMD-ESTMD\cite{wiederman2013biologically}, and ESTMD-EMD\cite{wiederman2013biologically}, have not provided unified and rigorous mathematical description.
	\item Wiederman and O'Carroll \cite{wiederman2013biologically} and Bagheri \emph{et al.} \cite{bagheri2017autonomous,bagheri2017performance,bagheri2015properties} focused on the size selectivity, tracking mechanisms and non-directionally selective properties, e.g., velocity and contrast tuning. Since direction selectivity has not been systematically studied, characteristics and performance of the directionally selective STMD models, are unclear. 
	\item The existing models have not shown the capacity for encoding motion direction of small targets.
\end{enumerate}
In this paper, we propose a neural network to model the specific STMD neurons with direction selectivity called DSTMD. It can detect not only small target motion but also the motion direction in cluttered backgrounds. The proposed neural network incorporates a new correlation mechanism which correlates signals relayed from two pixels so as to introduce directional selectivity. Then, a lateral inhibition mechanism acting on correlation outputs is used for size selectivity. Finally, a population vector algorithm is used to encode motion direction of small targets. Systematic experiments are carried out to validate the proposed neural network in complex environments.

The main contributions of this paper can be summarized as follows.
\begin{enumerate}
	\item We develop a new directionally selective STMD-based neural network (DSTMD) with unified and rigorous mathematical description.
	\item We systematically study and test both directionally selective and non-directionally selective properties of the developed neural network.
	\item We propose a population vector algorithm to encode motion direction of small targets.
\end{enumerate}

The remainder of this paper is organized as follows. In Section \ref{Related-Work}, the related work will be reviewed. In Section \ref{Formulation-of-the-model}, the proposed neural network is described in detail. In Section \ref{Results-and-Discussions}, the experiments are carried out to test the performances of the proposed neural network. We give further discussions in Section \ref{Further-Discussions} and finally in Section \ref{Conclusion}, we conclude this paper.

\section{Related Work}
\label{Related-Work}
In this section, we review the related work on three motion sensitive neurons, including the lobula giant movement detector (LGMD) \cite{judge1997locust,rind1992orthopteran,simmons1997responses}, lobula plate tangential cell (LPTC) \cite{lee2015spatio,borst1990direction,borst1995mechanisms} and STMD \cite{nordstrom2006insect,nordstrom2012neural,nordstrom2006small,o1993feature,barnett2007retinotopic}. These three neurons are all found in insects' visual systems and have been extensively studied. 

\subsection{Lobula Giant Movement Detector}
LGMDs are collision sensitive neurons found in locusts (certain species of short-horned grasshoppers) \cite{judge1997locust,rind1992orthopteran,simmons1997responses}. They respond strongly to the objects approaching the insect on a direct collision course while exhibiting little or no response to receding objects. A great number of LGMD-based neural networks \cite{rind1996neural,yue2006collision,yue2013redundant,yue2006bio,LGMD2-BMVC,hu2016bio} have been developed. These neural networks showed the same collision sensitivity as the LGMD neuron and can detect collisions cheaply and reliably in a complex background. Nevertheless, they are incapable of detecting small target motion, and do not show size and direction selectivities.

\subsection{Lobula Plate Tangential Cell}
LPTCs exhibit strong responses to wide-field motion, but also to the motion of local, salient features \cite{lee2015spatio,borst1990direction,borst1995mechanisms}. The first LPTC model which is the spatial integration of elementary motion detectors (EMDs), was originally inferred from behavioral investigation of insects  \cite{hassenstein1956systemtheoretische}. In the past decade, considerable progress has been made in identifying the afferent pathways and the characteristics of the LPTCs. To incorporate these new biological findings, the EMD was adapted, giving rise to several models, such as two-quadrant-detector (TQD) \cite{franceschini1989directionally,eichner2011internal} and weighted-quadrant-detector \cite{clark2011defining}. The above-mentioned models respond to objects' motion,  but they are not size selective.

\subsection{Small Target Motion Detector}
Small target motion detectors are characterized by exquisite size selectivity \cite{nordstrom2006insect,nordstrom2012neural,nordstrom2006small}, some of which are also directionally selective  \cite{o1993feature,barnett2007retinotopic}. Wiederman \emph{et al.} \cite{wiederman2008model} proposed the ESTMD, to model an STMD neuron with spatially integrated multiple ESTMDs. Although the ESTMD shows size selectivity, it is not directionally selective.Wiederman and O'Carroll \cite{wiederman2013biologically} and Bagheri \emph{et al.} \cite{bagheri2017autonomous,bagheri2017performance,bagheri2015properties} mentioned that two hybrid models, i.e., EMD-ESTMD and ESTMD-EMD, could exhibit both size and direction selectivities. However, characteristics and performance of these directionally selective STMD models, are unclear, since direction selectivity has not been systematically studied.

\section{Formulation of the model}
\label{Formulation-of-the-model}
\begin{figure}[t]
	\centering
	\includegraphics[width=0.45\textwidth]{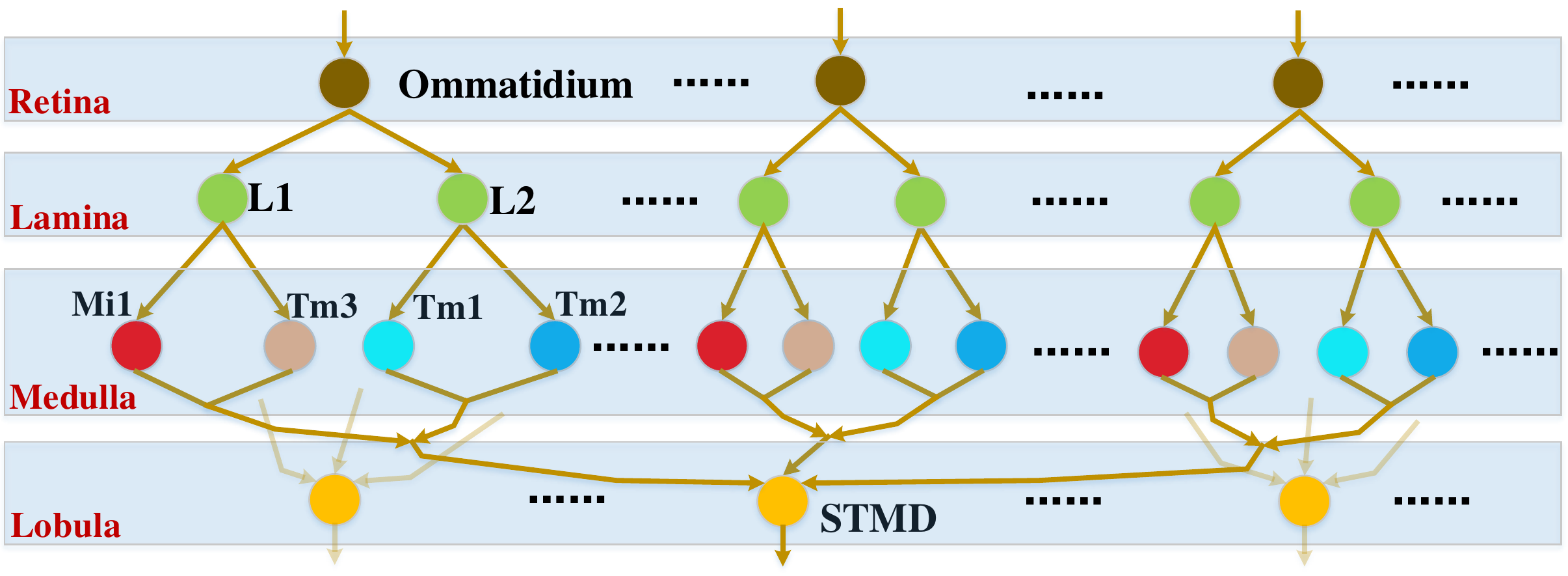}
	\caption{Schematic illustration of the insect's visual system. The insect's visual system consists of four neural layers, including retina, lamina, medulla and lobula (from top to bottom). Each neural layer contains numerous specialized neurons illustrated by colored circular nodes. Luminance signals are first perceived by ommatidia, further processed by LMCs (i.e., L1 and L2) and medulla neurons (Mi1, Tm1, Tm2, Tm3), finally integrated in STMD neurons. Note that the connection between the four medulla neurons and the STMD neuron is speculative.}
	\label{Schematic-of-Fly-Visual-System}
\end{figure}

\begin{figure}[t]
	\centering
	\includegraphics[width=0.40\textwidth]{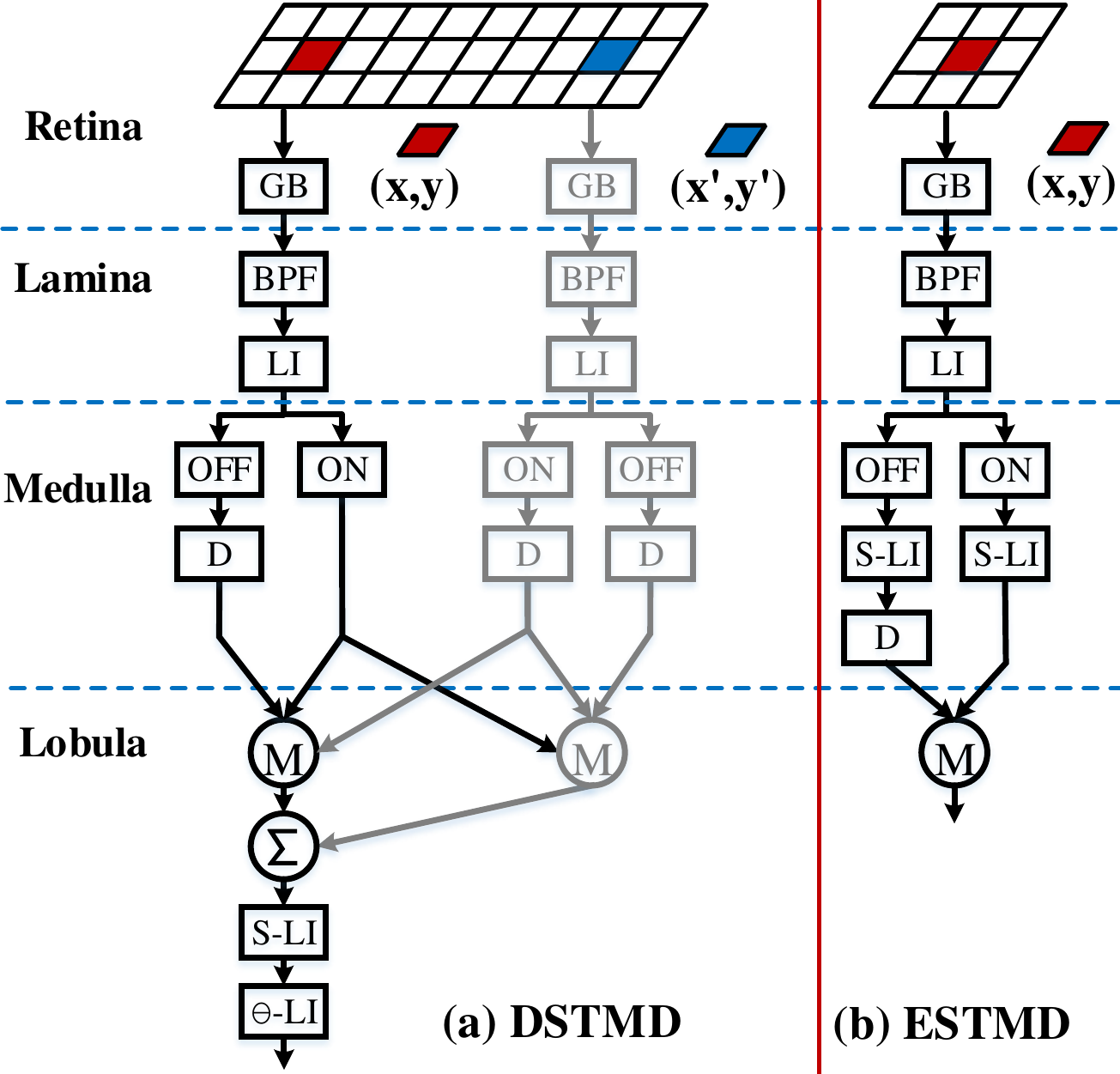}
	\caption{Schematic illustration of the proposed DSTMD and the existing ESTMD models which exhibit selectivity for dark small targets. (a) Schematic illustration of one DSTMD located at $(x,y)$ with a preferred direction $\theta$.  (b) Schematic illustration of one ESTMD located at $(x,y)$. The most significant difference between the DSTMD and ESTMD is that the DSTMD integrates signals from two different positions $(x,y)$ and $(x',y')$ where $x' = x + \alpha_1 \cos \theta$, $y' = y + \alpha_1 \sin \theta$, $\alpha_1$ is a constant. However, the ESTMD integrates signals from a single position $(x,y)$. Therefore, for each position $(x,y)$, the DSTMD has multiple model outputs corresponding to different preferred directions $\theta$ while the ESTMD just has a single output without direction selectivity. \emph{Abbreviation}, GB: Gaussian blur, BPF: band-pass filter, LI: lateral inhibition, ON/OFF: ON/OFF signals, D: time delay, M and $\sum$: multiplier and adder, S-LI: second-order lateral inhibition, $\theta$-LI: lateral inhibition implemented on $\theta$.}
	\label{Schematic-of-DSTMD-ESTMD}
\end{figure}

Following the typical multi-stage view of motion detection in the insect's visual system (schematically illustrated in Fig. \ref{Schematic-of-Fly-Visual-System}), we devised a DSTMD in this paper. Fig. \ref{Schematic-of-DSTMD-ESTMD}(a) shows the schematic of one DSTMD cell and its presynaptic neural network. The proposed neural network is composed of four neural layers including the retina, lamina, medulla, and lobula. These four sequentially arranged neural layers have specific functions and cooperate together for small target motion detection. In the following sections, we will elaborate on the components and functions of each layer.

\subsection{Retina Layer}
In the insect's visual system, the retina layer contains a great number of ommatidia (see Fig. \ref{Schematic-of-Fly-Visual-System}). Each ommatidium is composed of eight photoreceptors. Each photoreceptor views a small region of the whole viusal filed and supplies a 'pixel' of luminance information to ommatidia \cite{warrant2016matched}.

\begin{figure}[t]
	\centering
	\includegraphics[width=0.15\textwidth]{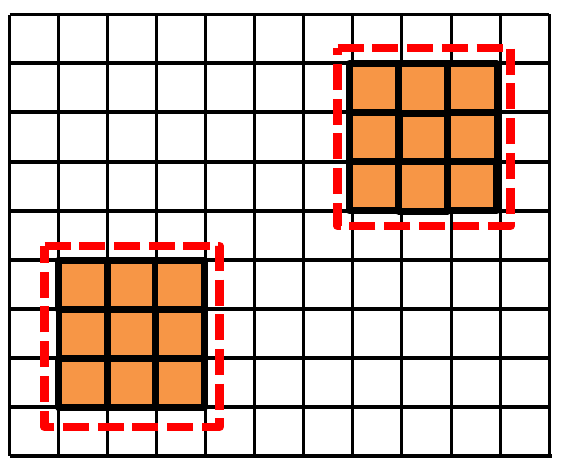}
	\caption{Schematic illustration of the mapping from pixels to photoreceptors. Each small square denotes a pixel, corresponding to a photoreceptor. Each red dotted rectangle which contains multiple pixels (photoreceptors), represents the visual region of an ommatidium.}
	\label{Mapping-from-pixels-to-photoreceptors-C1}
\end{figure}

In the proposed neural network, image sequences are network inputs, so we first construct a mapping from pixels to photoreceptors. As depicted in Fig. \ref{Mapping-from-pixels-to-photoreceptors-C1}, each small square denotes a pixel, corresponding to a photoreceptor. The red dotted rectangle which contains multiple pixels (photoreceptors), represents the visual region of an ommatidium. 

Specifically, let $I(x,y,t) \in \mathbb{R}$ denote varying luminance values captured by photoreceptors where $x,y$ and $t$ are spatial and temporal field positions. Then the response of an ommatidium is approximated by Gaussian blur. That is, the output of an ommatidium with viusal region centered at $(x,y)$ denoted by $P(x,y,t)$ is given by,
\begin{equation}
P(x,y,t) =  \iint I(u,v,t)G_{\sigma_1}(x-u,y-v)dudv
\label{Photoreceptors-Gaussian-Blur}
\end{equation}
where $G_{\sigma_1}(x,y)$ is a Gaussian function, defined as
\begin{equation}
G_{\sigma_1}(x,y)= \frac{1}{2\pi\sigma_1^2}\exp(-\frac{x^2+y^2}{2\sigma_1^2}).
\label{Photoreceptors-Gauss-blur-Kernel}
\end{equation}

\subsection{Lamina Layer}
In the insect's visual system, large monopolar cells (LMCs), such as L1 and L2, are postsynaptic neurons of the ommatidia (see Fig. \ref{Schematic-of-Fly-Visual-System}). They receive signals from the ommatidia and show strong responses to luminance increments and decrements, i.e., luminance changes \cite{freifeld2013gabaergic,behnia2014processing}. 

\begin{figure}[t!]
	\centering
	\subfloat[]{\includegraphics[width=0.23\textwidth]{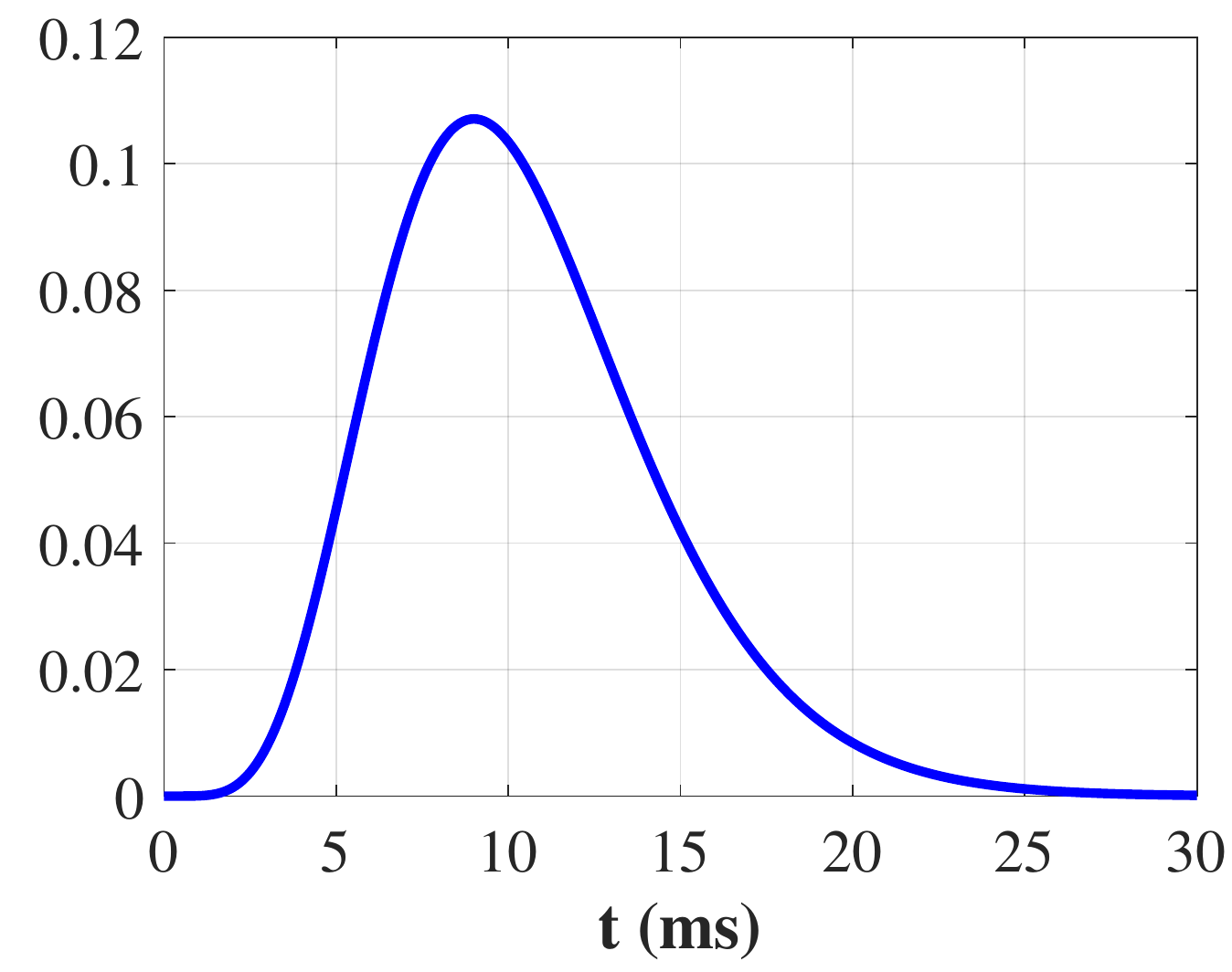}
		\label{GammaFun}}
	\hfil
	\subfloat[]{\includegraphics[width=0.23\textwidth]{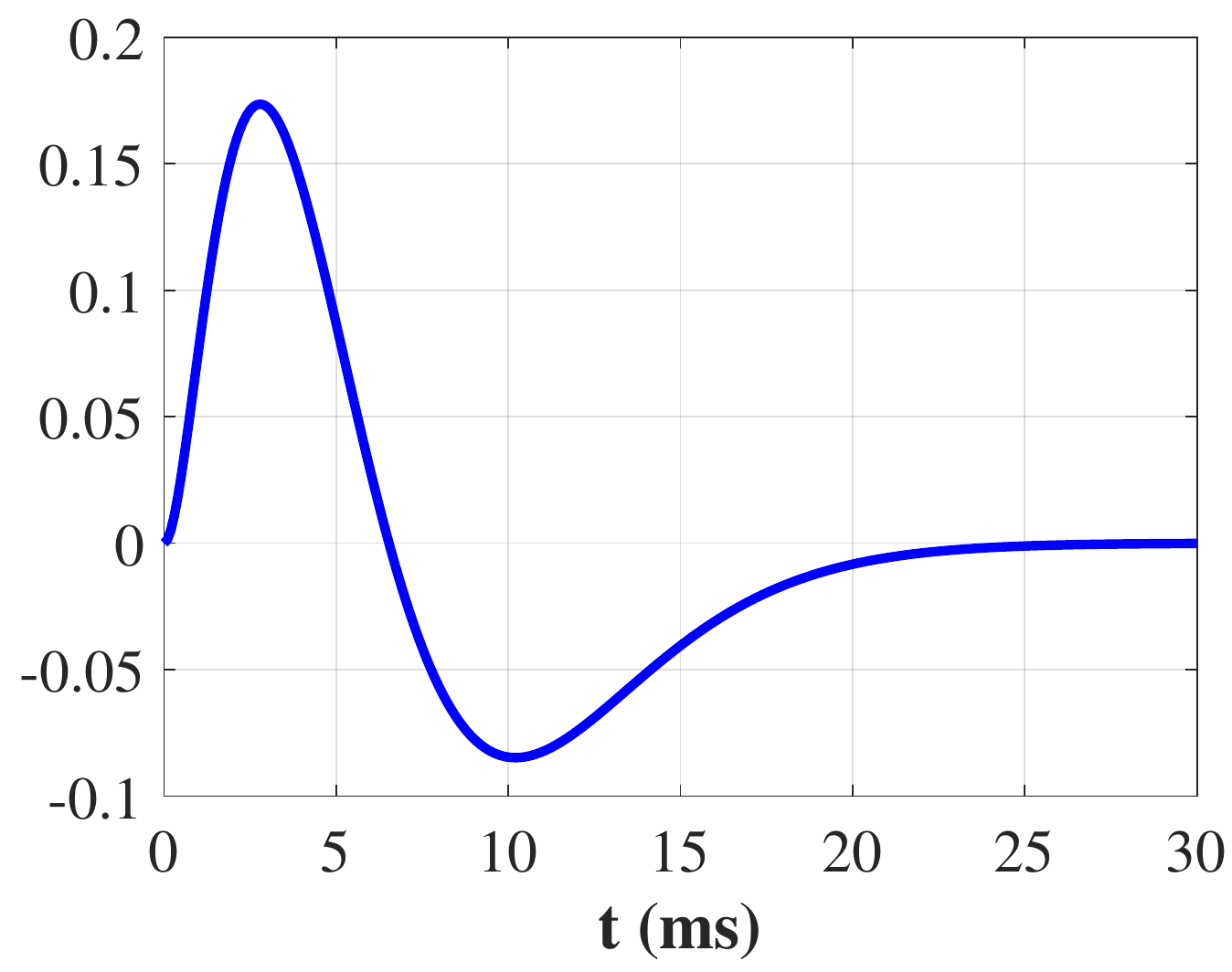}
		\label{DiffofGamma}}
	\caption{(a) Gamma kernel $\Gamma_{n,\tau}(t)$ where $n=6$, $\tau = 9$. (b) Temporal band-pass filter $H(t)$ where $n_1=2$, $\tau_1 = 3$, $n_2=6$, $\tau_2 = 9$.}
	\label{Schematic-of-Gamma}
\end{figure}

In the proposed neural network, each LMC is modeled as a temporal band-pass filter to extract luminance changes from input signals. Let $L(x,y,t)$ denote the output of a LMC located at $(x,y)$. Then $L(x,y,t)$ is defined by convolving the ommatidium output $P(x,y,t)$ with a temporal band-pass filter $H(t)$. That is,
\begin{align}
L(x,y,t) &= \int P(x,y,s)H(t-s) ds \label{LMCs-Conv} \\
H(t) &= \Gamma_{n_1,\tau_1}(t) - \Gamma_{n_2,\tau_2}(t)
\label{LMCs-HPF}
\end{align}
where $\Gamma_{n,\tau}(t)$ is a Gamma kernel, defined as
\begin{equation}
\Gamma_{n,\tau}(t) = (nt)^n \frac{\exp(-nt/\tau)}{(n-1)!\tau^{n+1}}.
\end{equation}
The illustration of the Gamma kernel $\Gamma_{n,\tau}(t)$ and temporal band-pass filter $H(t)$ is presented in Fig. \ref{Schematic-of-Gamma}.

In the insect's visual system, the LMC receives lateral inhibition from adjacent neurons before relaying its output to the next layer. In the proposed neural network, $L(x,y,t)$ is convolved with an inhibition kernel $W_1(x,y,t)$ so as to implement the lateral inhibition mechanism. That is,
\begin{equation}
L_{I}(x,y,t) = \iiint L(u,v,s)W_1(x-u,y-v,t-s) du dv ds
\label{LMCs-Lateral-Inhibition-Mechanism}
\end{equation}
where $L_{I}(x,y,t)$ is the signal after lateral inhibition and $W_1(x,y,t)$ is defined as,
\begin{equation}
W_1(x,y,t) = W_{S}^{P}(x,y)W_{T}^{P}(t) +W_{S}^{N}(x,y)W_{T}^{N}(t)
\end{equation}
where $W_{S}^{P}(x,y)$, $W_{S}^{N}(x,y)$, $W_{T}^{P}(t)$, $W_{T}^{N}(t)$ are set as
\begin{align}
W_{S}^{P} &= [G_{\sigma_2}(x,y) - G_{\sigma_3}(x,y)]^+ \label{LMCs-Lateral-Inhibition-Kernel-1}\\
W_{S}^{N} &= [G_{\sigma_2}(x,y) - G_{\sigma_3}(x,y)]^- ,  \ \sigma_3 = 2 \cdot \sigma_2 \label{LMCs-Lateral-Inhibition-Kernel-2}\\
W_{T}^{P} &= \frac{1}{\lambda_1}\exp(-\frac{t}{\lambda_1})  \label{LMCs-Lateral-Inhibition-Kernel-3}\\
W_{T}^{N} &= \frac{1}{\lambda_2}\exp(-\frac{t}{\lambda_2}),  \ \lambda_2 > \lambda_1 \label{LMCs-Lateral-Inhibition-Kernel-4}.
\end{align}
Here, $[x]^+, [x]^-$ denote $\max (x,0)$ and $\min (x,0)$, respectively.

\subsection{Medulla Layer}
\label{Modeling-Medulla-Layer}
In the insect's visual system, medulla neurons including Tm1, Tm2, Tm3 and Mi1, are downstream neurons of the LMCs (see Fig. \ref{Schematic-of-Fly-Visual-System}). The Mi1 and Tm3 respond selectively to luminance increments, with the response of the Mi1 delayed relative to the Tm3 \cite{behnia2014processing}. Conversely, the Tm1 and Tm2 respond selectively to luminance decrements, with the response of the Tm1 delayed relative to the Tm2 \cite{yang2016subcellular}.

In the proposed DSTMD and the existing ESTMD \cite{wiederman2008model}, the modeling methods for these four medulla neurons, are different. These two modeling methods are introduced as follows, respectively.

\textit{1) Medulla Neuron Modeling of DSTMD:}
As the Tm3 and Tm2 neurons respond strongly to luminance increments and decrements, we use the positive and negative parts of the LMC output $L_{I}(x,y,t)$ to define the outputs of the Tm3 and Tm2, denoted by $S^{\text{Tm3}}$ and $S^{\text{Tm2}}$, respectively. That is, 
\begin{align}
S^{\text{Tm3}}(x,y,t) &= S^{\text{ON}}(x,y,t)  \label{DS-STMD-Tm3}\\
S^{\text{Tm2}}(x,y,t) &= S^{\text{OFF}}(x,y,t)  \label{DS-STMD-Tm2}
\end{align}
where $S^{\text{ON}}$ and $S^{\text{OFF}}$ represent the positive and negative parts of $L_{I}(x,y,t)$, respectively. That is,
\begin{align}
S^{\text{ON}}(x,y,t) &= [L_{I}(x,y,t)]^{+} \label{ON-Channel} \\ 
S^{\text{OFF}}(x,y,t) &= -[L_{I}(x,y,t)]^{-}  \label{OFF-Channel}
\end{align}
where $S^{^{ON}}(x,y,t)$ and $S^{^{OFF}}(x,y,t)$ are also called ON and OFF signals (see the ON and OFF in Fig. \ref{Schematic-of-DSTMD-ESTMD}), which reflect luminance increase and decrease, respectively.

Since the Mi1 (or Tm1) is a temporally delayed version of the Tm3 (or Tm2), the output of the Mi1 (or Tm1) is defined by convolving $S^{\text{Tm3}}(x,y,t)$ (or $S^{\text{Tm2}}(x,y,t)$) with a Gamma kernel. That is,
\begin{align}
S_{{D(n_{N},\tau_{N})}}^{\text{Mi1}}(x,y,t) &= \int S^{\text{Tm3}}(x,y,s)\Gamma_{n_{N},\tau_{N}}(t-s) ds \label{ON-Channel-Delay} \\
S_{{D(n_{F},\tau_{F})}}^{\text{Tm1}}(x,y,t) &= \int S^{\text{Tm2}}(x,y,s)\Gamma_{n_{F},\tau_{F}}(t-s) ds
\label{OFF-Channel-Delay}
\end{align}
where $S_{{D(n_{N},\tau_{N})}}^{\text{Mi1}}$ and $S_{{D(n_{F},\tau_{F})}}^{\text{Tm1}}$ represent the outputs of the Mi1 and Tm1, respectively. $n_{N}, n_{F}$ are orders of the Gamma kernels while $\tau_{N}, \tau_{F}$ are time constants.

\textit{2) Medulla Neuron Modeling of ESTMD:}
The most significant difference between the medulla neuron modeling methods of the DSTMD and ESTMD is that the ESTMD uses laterally inhibited ON and OFF signals to define the outputs of the medulla neurons. This can be seen in Fig. \ref{Schematic-of-DSTMD-ESTMD} that the ESTMD implements a second-order lateral inhibition mechanism following ON and OFF signals while the DSTMD does not. In the ESTMD, the outputs of the Tm3 and Tm2 denoted by $\tilde{S}^{\text{Tm3}}$ and $\tilde{S}^{\text{Tm2}}$, are defined as,
\begin{align}
\tilde{S}^{\text{Tm3}}(x,y,t) &= \Big{[}\iint S^{\text{ON}}(u,v,t) W_2(x-u,y-v) du dv\Big{]}^{+}
\label{ESTMD-Tm3-Lateral-Inhibition}  \\
\tilde{S}^{\text{Tm2}}(x,y,t) &= \Big{[}\iint S^{\text{OFF}}(u,v,t) W_2(x-u,y-v) du dv\Big{]}^{+} 
\label{ESTMD-Tm2-Lateral-Inhibition}
\end{align}
where $S^{\text{ON}}$ and $S^{\text{OFF}}$ are the ON and OFF signals defined in (\ref{ON-Channel}) and (\ref{OFF-Channel}); $W_2(x,y)$ is the second-order lateral inhibition kernel, defined as
\begin{equation}
W_2(x,y) = A[g(x,y)]^{+} + B[g(x,y)]^{-}
\label{ESTMD-Mdeulla-Lateral-Inhibition-Kernel-W2}
\end{equation}
where $A,B$ are constant, and $g(x,y)$ is given by
\begin{equation}
g(x,y)  = G_{\sigma_4}(x,y) - e \cdot G_{\sigma_5}(x,y) - \rho
\label{ESTMD-Mdeulla-Lateral-Inhibition-Kernel-W2-2}
\end{equation}
where $G_{\sigma}(x,y)$ is a Gaussian function and $e,\rho$ are constant.

Similarly, the outputs of the Tm1 and Mi1 are defined as the temporally delayed outputs of the Tm3 and Tm2, which are obtained by convolving $\tilde{S}^{\text{Tm3}}(x,y,t)$ and $\tilde{S}^{\text{Tm2}}(x,y,t)$ with a Gamma kernel. That is,
\begin{align}
\tilde{S}_{{D(n_{N},\tau_{N})}}^{\text{Mi1}}(x,y,t) &= \int \tilde{S}^{\text{Tm3}}(x,y,s)\Gamma_{n_{N},\tau_{N}}(t-s) ds \label{ON-Channel-Delay-ESTMD} \\
\tilde{S}_{{D(n_{F},\tau_{F})}}^{\text{Tm1}}(x,y,t) &= \int \tilde{S}^{\text{Tm2}}(x,y,s)\Gamma_{n_{F},\tau_{F}}(t-s) ds
\label{OFF-Channel-Delay-ESTMD}
\end{align}
where $\tilde{S}_{{D(n_{_N},\tau_{_N})}}^{\text{Mi1}}$ and $\tilde{S}_{{D(n_{_F},\tau_{_F})}}^{\text{Tm1}}$ stand for the outputs of the Mi1 and Tm1, respectively.

In the following, we discuss the implementation of the second-order lateral inhibition mechanism. Existing biological research \cite{bolzon2009local} asserts that the size selectivity of the STMD neurons is shaped by the second-order lateral inhibition mechanism. However, where this second-order lateral inhibition mechanism occurs remains elusive. Although the ESTMD implements this second-order lateral inhibition mechanism on medulla neurons, it is just speculative and there is no neuroanatomical evidence for it. On the other hand, we notice that the LPTC neurons also receive signals from medulla neurons \cite{tuthill2016four,behnia2014processing}. If the medulla neurons which provide signals to the LPTC neurons, are laterally inhibited, the LPTC neurons would show strong size selectivity (this will be demonstrated in the experiment Section \ref{Tuning-Properties}). This may conflict with the finding that the LPTCs do not exhibit size selectivity \cite{lee2015spatio,borst1990direction,borst1995mechanisms}. To satisfy both size selectivity of the STMDs and size insensitivity of the LPTCs, we infer that the second-order lateral inhibition mechanism may happen on the STMD pathway rather than medulla neurons in the implementation of our proposed neural network.

\subsection{Lobula Layer}
In the insect's visual system, the STMD neurons integrate signals from the medulla layer. They respond strongly to small target motion, but show weak or no response to wide-field motion (size selectivity) \cite{nordstrom2006insect,nordstrom2012neural,nordstrom2006small}. Besides, some STMDs exhibit strong responses to small target motion oriented along a preferred direction, but show weak or no response to opposite-direction motion (direction selectivity) \cite{o1993feature,barnett2007retinotopic}.

In the proposed neural network (DSTMD), a new correlation mechanism and a second-order lateral inhibition mechanism are introduced for direction and size selectivities, respectively. For comparison with the proposed neural network, the existing non-directionally selective ESTMD \cite{wiederman2008model} is also presented in the following.

\textit{1) ESTMD:}
In the ESTMD, the output of an STMD neuron located at $(x,y)$, denoted by $\tilde{D}(x,y,t)$, is defined as,
\begin{equation}
\tilde{D}(x,y,t) = \tilde{S}^{\text{Tm3}}(x,y,t)\times \tilde{S}^{\text{Tm1}}_{{D(n_3,\tau_3)}}(x,y,t).
\label{ESTMD-Correlation}
\end{equation}

As we can see from (\ref{ESTMD-Correlation}), the output of an STMD neuron located at $(x,y)$ is defined by the multiplication of the Tm1 and Tm3 outputs at the same position. Since medulla neural signals from at least two different positions are needed for detecting motion direction \cite{escobar2013mathematical}, the ESTMD is able to detect the presence of target motion, but not the target's motion direction.

\textit{2) DSTMD:}
In the DSTMD, the correlation output of an STMD neuron located at $(x,y)$ with a preferred motion direction $\theta$, denoted by $D(x,y,t;\theta)$, is defined as,
\begin{equation}
\begin{split}
D(x,y,t;\theta) = S^{\text{Tm3}}(x,y,t) \times \Big \{&S^{\text{Tm1}}_{{D(n_{5},\tau_{5})}}(x,y,t) \\   &+S^{\text{Mi1}}_{{D(n_{4},\tau_{4})}}(x',y',t)\Big\} \\
{\times  S^{\text{Tm1}}_{{D(n_{6},\tau_{6})}}(x',y',t)}
\end{split}
\label{DS-STMD-Signal-Correlation}
\end{equation}
where
\begin{equation}
\begin{split}
x' &= x+\alpha_1\cos\theta \\ 
y' &= y + \alpha_1\sin\theta
\end{split}
\label{DS-STMD-Signal-Correlation-Distance}
\end{equation}
and $\alpha_1$ is a constant, $\theta \in \{0, \frac{\pi}{4}, \frac{\pi}{2}, \frac{3\pi}{4}, \pi, \frac{5\pi}{4}, \frac{3\pi}{2}, \frac{7\pi}{4}\}$.


\begin{figure}[t!]
	\centering
	\subfloat[]{\includegraphics[width=0.175\textwidth]{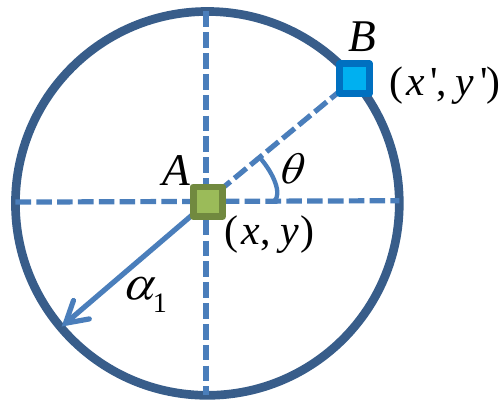}
		\label{Schematic-Relative-Position-of-Photoreceptor-A-and-B}}
	\hfil
	\subfloat[]{\includegraphics[width=0.13\textwidth]{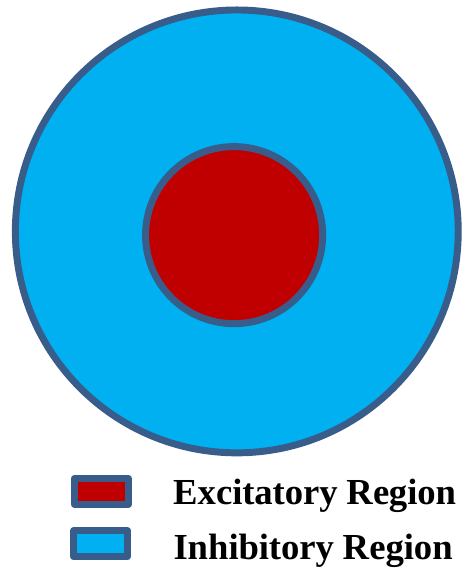}
		\label{Schematic-Second-Order-Inhibition-Mechanism}}
	\caption{(a) Schematic illustration of relative position between A $(x,y)$ and B $(x',y')$. $\alpha_1$ is the distance between A and B while $\theta$ is the angle between line segment AB and the horizontal line. (b) Schematic illustration of excitatory and inhibitory regions of the second-order lateral inhibition mechanism.}
	\label{Schematic-Correlation-Mechanism-and-Lateral-Inhibition-Mechanism}
\end{figure}

As we can see from (\ref{DS-STMD-Signal-Correlation}), four medulla neural signals from two different positions, i.e., $(x,y)$ and $(x',y')$, are used to define the output of an STMD neuron located at $(x,y)$ (see Fig. \ref{Schematic-of-DSTMD-ESTMD}, two multipliers and one adder). These four medulla neural signals include the outputs of the Tm1 and Tm3 located at position $(x,y)$, i.e., $S^{\text{Tm1}}_{{D(n_{_5},\tau_{_5})}}(x,y,t)$ and $S^{\text{Tm3}}(x,y,t)$, the outputs of the
Tm1 and Mi1 located at position $(x',y')$, i.e., $S^{\text{Tm1}}_{{D(n_{_6},\tau_{_6})}}(x',y',t)$ and $S^{\text{Mi1}}_{{D(n_{_4},\tau_{_4})}}(x',y',t)$ (the full derivation of (\ref{DS-STMD-Signal-Correlation}) is shown in the supplementary materials). The schematic illustration of relative position between $(x,y)$ and $(x',y')$ is presented in Fig. \ref{Schematic-Correlation-Mechanism-and-Lateral-Inhibition-Mechanism}(a). For a given position $(x,y)$, we can choose a series of $(x',y')$, corresponding to different directions $\theta$. Thus, a series of correlation outputs $D(x,y,t;\theta)$ with different preferred motion directions $\theta$ can be defined. For a given direction $\theta_0$, $D(x,y,t; \theta_0)$ gives the strongest output to small target motion oriented along direction $\theta_0$, with weak or no outputs to motion oriented along other directions. That is, $D(x,y,t;\theta)$ shows direction selectivity.

After the signal correlation, the DSTMD implements the second-order lateral inhibition mechanism on $D(x,y,t;\theta)$ for size selectivity. That is,
\begin{equation}
D_{I}(x,y,t;\theta) = \Big{[}\iint D(u,v,t;\theta) W_2(x-u,y-v) du dv\Big{]}^{+}
\label{DS-STMD-Lateral-Inhibition}
\end{equation}
where $D_{I}(x,y,t;\theta)$ is the signal after lateral inhibition and $[x]^+$ denotes $\max (x,0)$, $W_2(x,y)$ is defined in (\ref{ESTMD-Mdeulla-Lateral-Inhibition-Kernel-W2}). 

The schematic illustration of inhibition kernel $W_2(x,y)$ is shown in Fig. \ref{Schematic-Correlation-Mechanism-and-Lateral-Inhibition-Mechanism}(b). As can be seen, the inhibition kernel $W_2(x,y)$ contains two components, i.e., excitatory and inhibitory regions. For the kernel $W_2(x,y)$, its surround inhibition is set as three times as strong as the center excitation. In this case, once the target's size exceeds the excitatory region, it will receive strong inhibition. When the target is smaller than the excitatory region, the amount of excitation will increase as the rise of target size. That is, the DSTMD prefers the target whose size is equal to the excitatory region and exhibits size selectivity.

Following the second-order lateral inhibition mechanism, the DSTMD inhibits model output $D_{_I}(x,y,t;\theta)$ at directions more than $45^{\circ}$ apart by convolving $D_{_I}(x,y,t;\theta)$ with an inhibition kernel $W_3(\theta)$. That is,
\begin{equation}
E(x,y,t;\theta) = \Big{[}\int D_{_I}(x,y,t;\varphi)W_3(\theta - \varphi) d\varphi \Big{]}^{+}
\label{DS-STMD-Lateral-Inhibition-Direction-W3}
\end{equation}
where $[x]^+$ denotes $\max (x,0)$ and $W_3(\theta)$ is defined as 
\begin{equation}
W_3(\theta) =  G_{\sigma_6}(\theta) - G_{\sigma_7}(\theta).
\label{DS-STMD-Lateral-Inhibition-Direction-W3-2}
\end{equation}
where $G_{\sigma}(x,y)$ is a Gaussian function. 

In the DSTMD, $E(x,y,t;\theta)$ is regarded as the output of the STMD neurons.

\subsection{Motion Direction Estimation}
In the insect's visual system, the STMD neurons are believed to be upstream of target selective descending neurons (TSDNs) \cite{o1993feature,nordstrom2012neural,gonzalez2013eight}. Further biological research \cite{gonzalez2013eight} found that eight pairs of the TSDNs are able to encode motion direction of targets by a population vector algorithm.

In the proposed neural network, we estimate motion directions of targets by populating the model output $E(x,y,t;\theta)$ along different directions $\theta$. That is,
\begin{equation}
MD(t) = \sum_{(x,y)\in \text{Target}} \sum_{\theta} (E(x,y,t;\theta) \cos \theta, E(x,y,t;\theta)\sin \theta)
\label{DS-STMD-Estimated-Direction}
\end{equation}
where $MD(t)$ denotes the motion direction of the small target at time $t$, $(x,y) \in \text{Target}$ stands for the position of the STMD neurons which respond to the small target motion.

\subsection{Parameter Setting}
Parameters of the proposed neural network (DSTMD) and ESTMD are given in Table \ref{Table-Parameter-STMD}. These parameters are tuned manually based on empirical experiences and will not be changed in the following experiments unless stated.

The proposed neural network is written in Matlab (The MathWorks, Inc., Natick, MA). The computer used in the experiments is a PC with one $2.50$ Ghz CPU (Core i7 4710MQ) and windows $7$ operating system. The source code can be found at https://github.com/wanghongxin/DSTMD.

\begin{table}[t!]
	\renewcommand{\arraystretch}{1.3}
	\caption{Parameters of the DSTMD and ESTMD}
	\label{Table-Parameter-STMD}
	\centering
	\begin{tabular}{cc}
		\hline
		Eq. & Parameters \\	
		\hline
		(\ref{Photoreceptors-Gaussian-Blur}) & $\sigma_1 = 1$ \\
		
		(\ref{LMCs-HPF}) & $n_1 = 2, \tau_1= 3, n_2 = 6,\tau_2 = 9$\\
		
		(\ref{LMCs-Lateral-Inhibition-Kernel-1}),  (\ref{LMCs-Lateral-Inhibition-Kernel-2})  &  $\sigma_2 = 1.5,  \sigma_3 = 3$ \\
		
		(\ref{LMCs-Lateral-Inhibition-Kernel-3}), 
		(\ref{LMCs-Lateral-Inhibition-Kernel-4})& $\lambda_1 = 3, \lambda_2 = 9$ \\
				
		(\ref{ESTMD-Mdeulla-Lateral-Inhibition-Kernel-W2})& $A = 1, B = 3$ \\
		
		 (\ref{ESTMD-Mdeulla-Lateral-Inhibition-Kernel-W2-2})  & $\sigma_4 = 1.5, \sigma_5 = 3, e = 1, \rho = 0$ \\
		 
		(\ref{ESTMD-Correlation})  & $n_3 = 5, \tau_3 = 25$                 \\
		(\ref{DS-STMD-Signal-Correlation}) & $n_4 = 3, \tau_4 = 15, n_5 = 5, \tau_5 = 25, n_6 = 8, \tau_6 = 40$ \\
		
		(\ref{DS-STMD-Signal-Correlation-Distance}) & $\alpha_1 = 3$ \\

		(\ref{DS-STMD-Lateral-Inhibition-Direction-W3-2}) & $\sigma_6 = 1.5, \sigma_7 = 3$ \\
				
		\hline
	\end{tabular}
\end{table}

\section{Results and Discussions}
\label{Results-and-Discussions}

The proposed neural network is tested on image sequences produced by Vision Egg \cite{straw2008vision}. The Vision Egg is a open-source programming library that allows scientists to produce arbitrary visual stimuli (http://visionegg.org/). Such stimuli involve traditional stimuli such as sinusoidal gratings, or may be more complex, 3-D, and naturalistic scenes. The image sequences used in this paper can be divided into two categories depending on background types. The first category contains image sequences showing small target motion against white backgrounds. This category is used to test the basic properties of the proposed neural network, such as tuning properties (see Sections \ref{Tuning-Properties} and \ref{Parameter-Sensitivity}), direction selectivity (see Section \ref{Direction-Selcetivity-and-Motion-Direction-Estimation}). The other category contains image sequences showing small target motion against naturally cluttered backgrounds. This category is used to test the detection performance of the proposed neural network in complex backgrounds (see Section \ref{Contribution-of-Various-Neurons} and \ref{Target-Detection-in-Cluttered-Backgrounds}). All image sequences can be reproduced by the Vision Egg with the same parameters (given before each experiment). The video images are $500$ (in horizontal) by $250$ (in vertical) pixels and temporal sampling frequency is set as $1000$ Hz.

\subsection{Contribution of Various Neurons}
\label{Contribution-of-Various-Neurons}
\begin{figure}[t]
	\centering
	\includegraphics[width=0.30\textwidth]{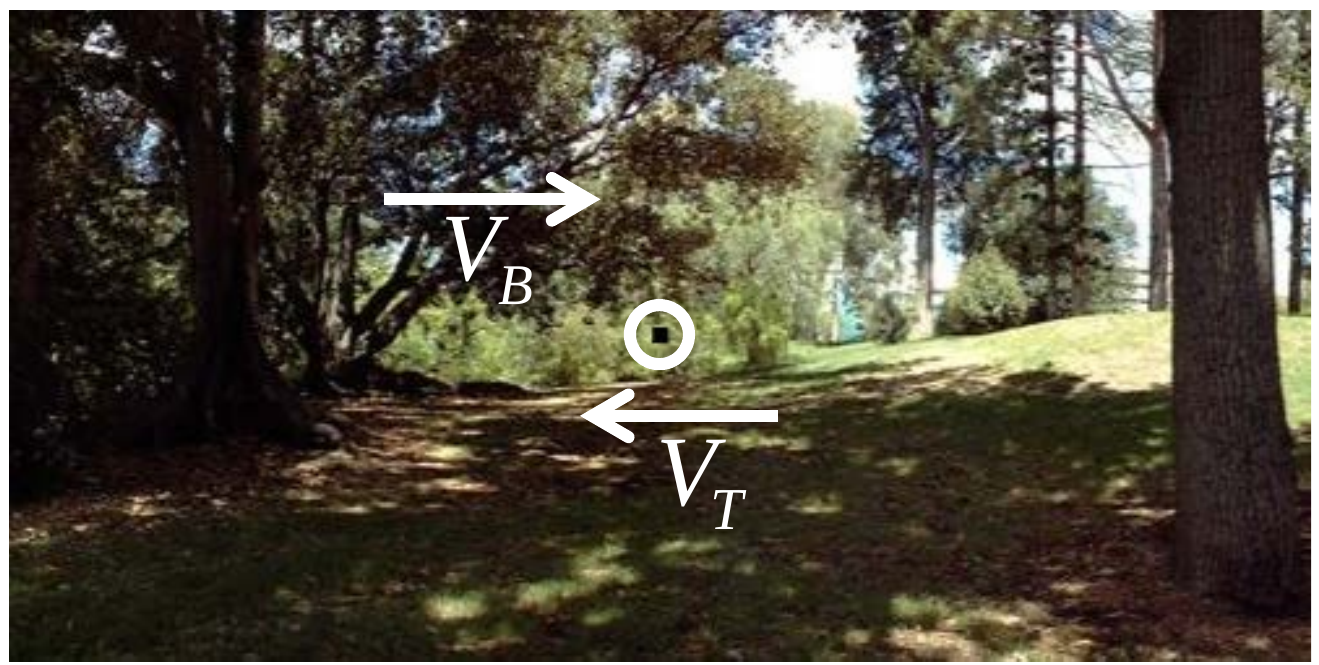}
	\caption{Representative frame of the input image sequence. A small rectangle highlighted by the white circle, is moving against the cluttered background. This rectangle whose size and luminance are set as $5 \times 5$ pixels and 0, is the small target needed to be detected. The arrows $V_T$ and $V_B$ denote the motion directions of the small target and the background, respectively. The velocities of the small target and the background are all set as $250$ pixel/s.}
	\label{Input-Image-Frame-Middle-Line-Highlighted}
\end{figure}

\begin{figure*}[!t]
	\centering
	{\vspace{-5pt}
	\subfloat[]{\includegraphics[width=0.32\textwidth]{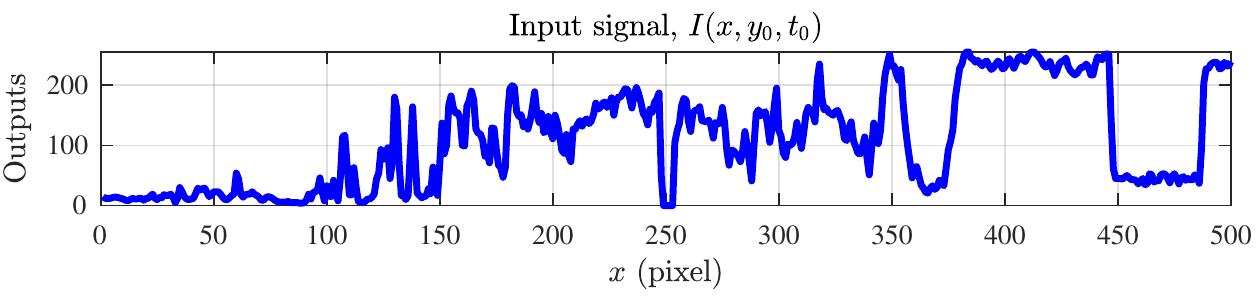}
		\label{Layer-Output-Intensity}}
	\hfil
		\subfloat[]{\includegraphics[width=0.32\textwidth]{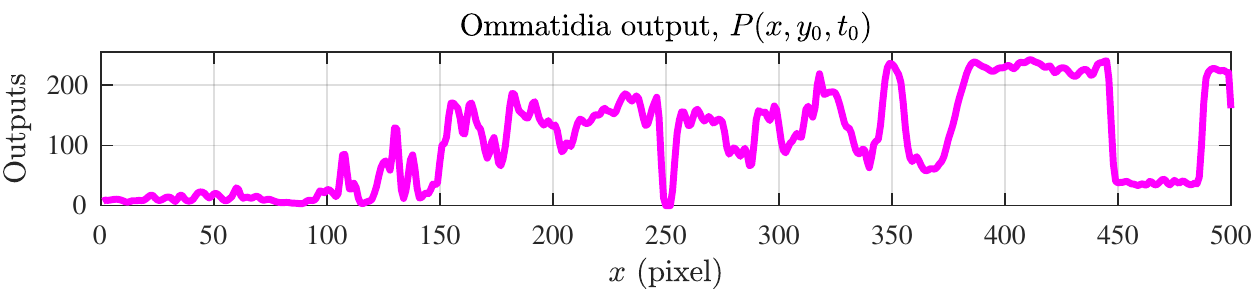}
		\label{Layer-Output-Photoreceptors}}
	\hfil
	\subfloat[]{\includegraphics[width=0.32\textwidth]{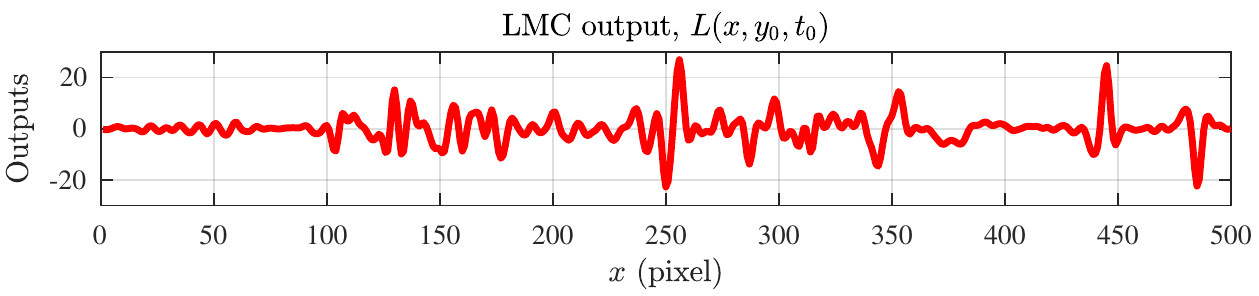}
		\label{Layer-Output-LMCs}}	}

{ 
	\subfloat[]{\includegraphics[width=0.40\textwidth]{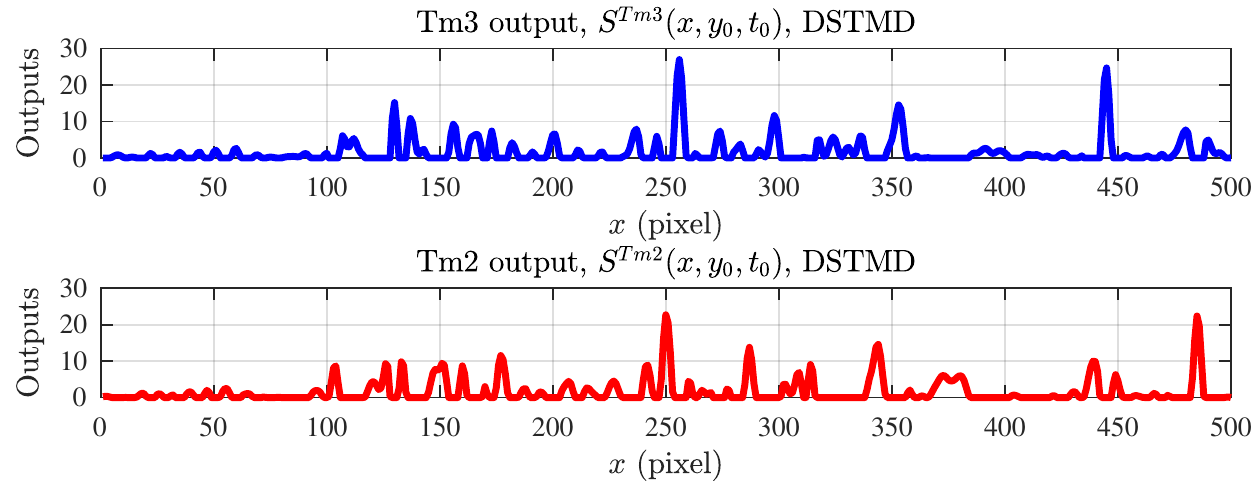}
		\label{Layer-Output-ON-OFF-Channels}}
	\hfil
	\subfloat[]{\includegraphics[width=0.40\textwidth]{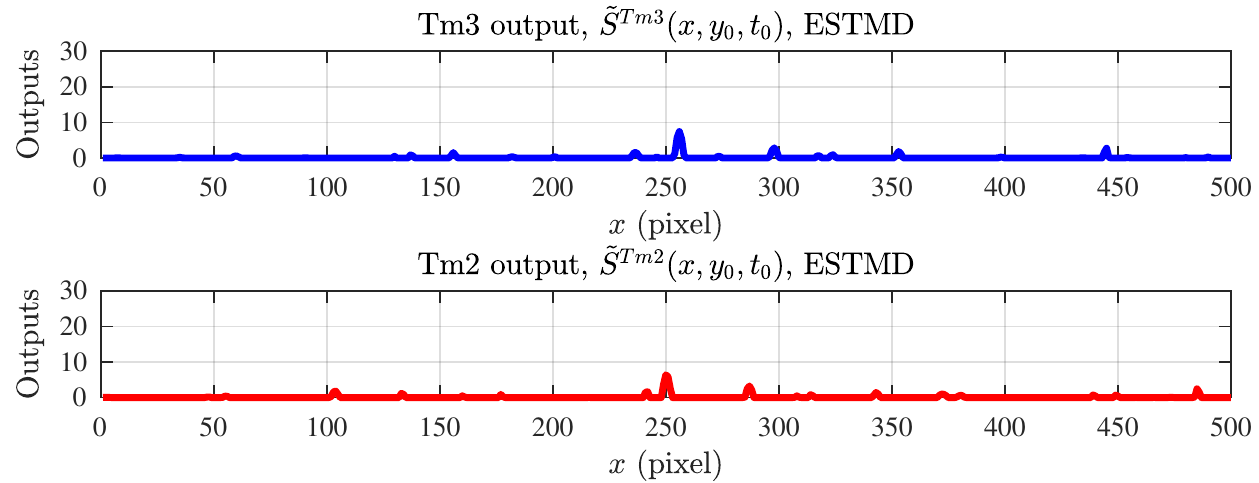}
		\label{Layer-Output-ON-OFF-Channels-LI}}
	}

{ 
	\subfloat[]{\includegraphics[width=0.40\textwidth]{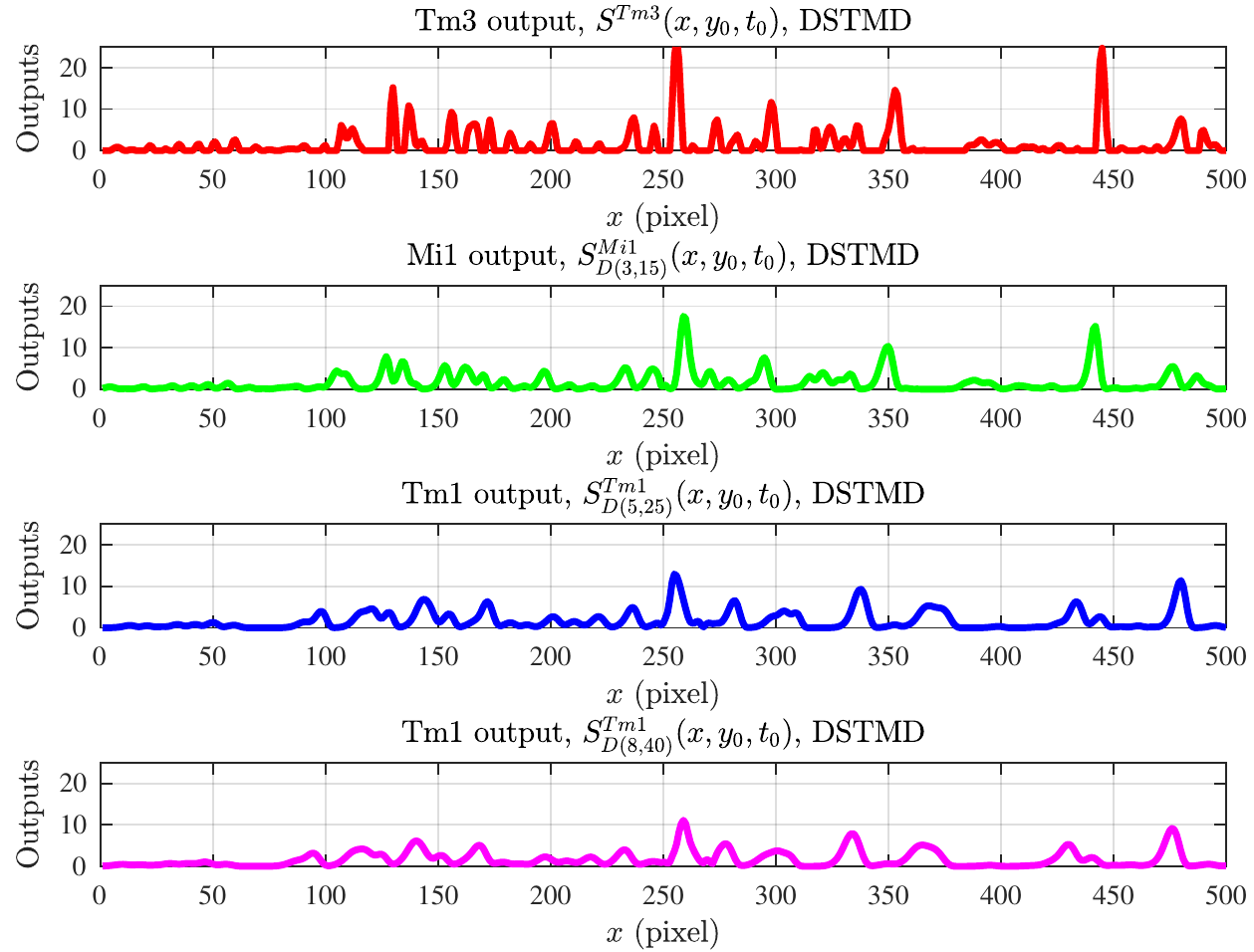}
		\label{Layer-Output-Medulla-Neuron-Signal-DS-STMD}}
	\hfil
	\subfloat[]{\includegraphics[width=0.40\textwidth]{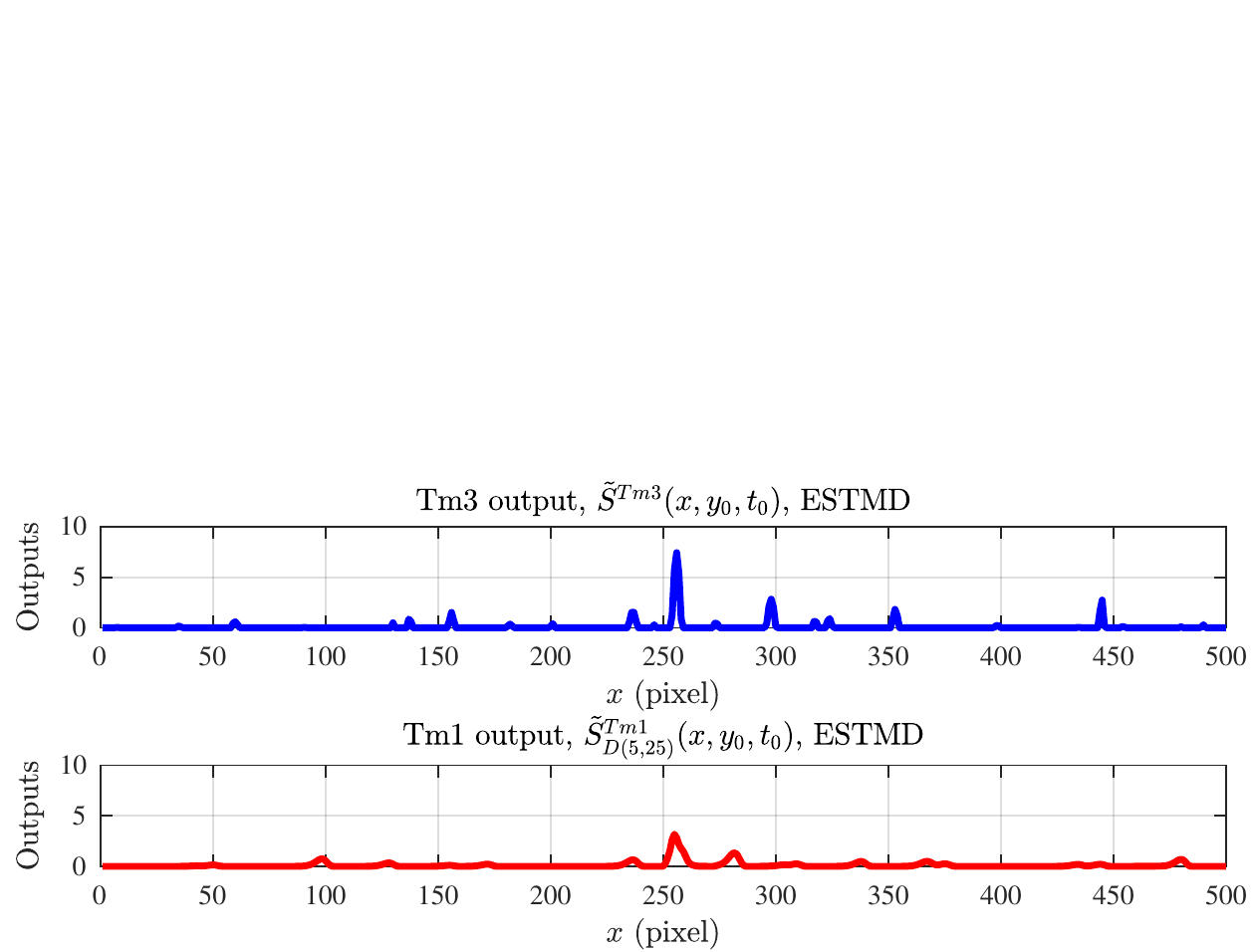}
		\label{Layer-Output-Medulla-Neuron-Signal-ESTMD}}
}
{
	\subfloat[]{\includegraphics[width=0.40\textwidth]{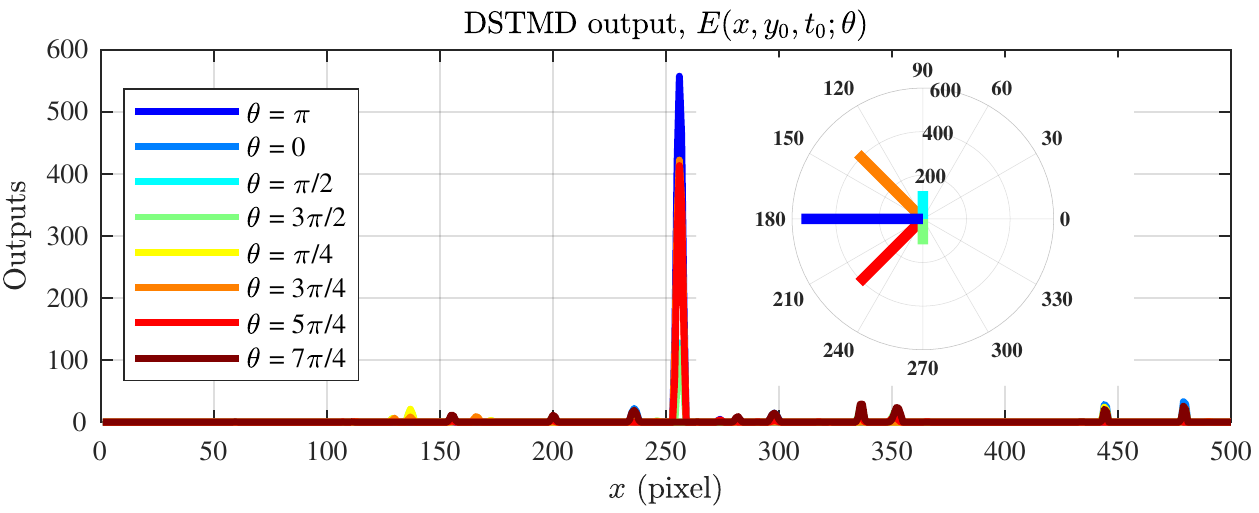}
		\label{Layer-Output-DS-STMD}}
	\hfil
	\subfloat[]{\includegraphics[width=0.40\textwidth]{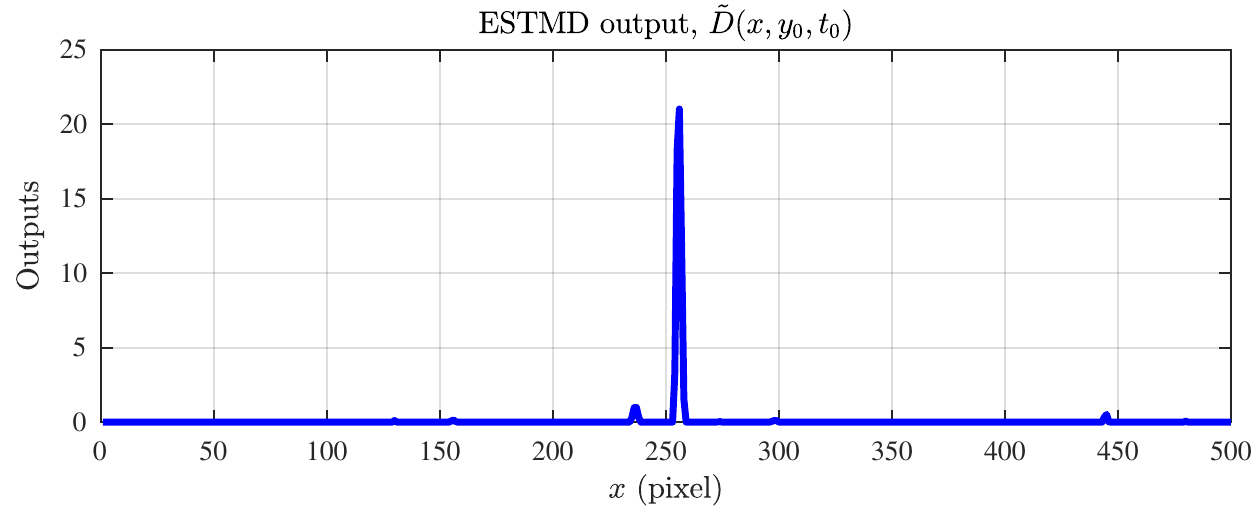}
		\label{Layer-Output-ESTMD}}
	}
{
	\caption{Outputs of various neurons in the DSTMD and ESTMD models where $x \in [0,500]$ pixel, $y_0=125$ pixel, $t_0 = 1000$ ms. In each subplot, the horizontal axis denotes the location of neurons ($x$ coordinate) while the vertical axis represents neural outputs. (a) Input signal $I(x,y_0,t_0)$. (b) Ommatidium output $P(x,y_0,t_0)$. (c) LMC output $L(x,y_0,t_0)$. (d) Outputs of the Tm3 and Tm2 modeled by the DSTMD, i.e., $S^{\text{Tm3}}(x,y_0,t_0)$ and $S^{\text{Tm2}}(x,y_0,t_0)$. (e) Outputs of the Tm3 and Tm2 modeled by the ESTMD, i.e., $\tilde{S}^{\text{Tm3}}(x,y_0,t_0)$ and $\tilde{S}^{\text{Tm2}}(x,y_0,t_0)$. (f) Medulla neural outputs used for the signal correlation in the DSTMD, i.e., $S^{\text{Tm3}}(x,y_0,t_0)$, $S^{\text{Mi1}}_{D(3,15)}(x,y_0,t_0)$, $S^{\text{Tm1}}_{D(5, 25)}(x,y_0,t_0)$ and $S^{\text{Tm1}}_{D(8,40)}(x,y_0,t_0)$. (g) Medulla neural outputs used for the signal correlation in the ESTMD, i.e., $\tilde{S}^{\text{Tm3}}(x,y_0,t_0)$ and $\tilde{S}^{\text{Tm1}}_{D(5,25)}(x,y_0,t_0)$. (h) DSTMD output $E(x,y_0,t_0;\theta)$. In this subplot, the DSTMD outputs to the small target are further shown in the polar coordinate system, where the angular coordinate denotes the preferred motion direction $\theta$ while the radial coordinate stands for the model output along this preferred direction. (i) ESTMD output $\tilde{D}(x,y_0,t_0)$.}
	\label{Neural-Layer-Outputs}}
\end{figure*}

To evaluate the characteristics of the neurons in the proposed neural network, we observe and analyze their outputs. For an input image sequence $I(x,y,t)$, where $x \in [0,500]$ pixel, $y \in [0,250]$ pixel, $t \in [0,1000]$ ms (see Fig. \ref{Input-Image-Frame-Middle-Line-Highlighted}), we first fix $y$ and $t$ as $y_0=125$ pixel and $t_0 = 1000$ ms, then illustrate $I(x,y_0,t_0)$ with respect to $x$ in Fig. \ref{Neural-Layer-Outputs}(a). Similarly, the outputs of other neurons are presented in the subplots below.

Fig. \ref{Neural-Layer-Outputs}(a)-(c) shows the input luminance signal $I(x,y_0,t_0)$, ommatidium output $P(x,y_0,t_0)$ and LMC output $L(x,y_0,t_0)$, respectively. Compared to the input signal, the ommatidium output demonstrates little difference and is just slightly smoothed. This is because the ommatidium is modeled as a spatial Gaussian filter to smooth the input luminance signals. The LMC output displays significant difference from the ommatidium output. More precisely, the LMC output becomes the band-pass-filtered version of the ommatidium output. From the other perspective, the LMC output reveals the luminance changes of pixels, where the positive values correspond to luminance increase while the negative values suggest luminance decrease. 

Fig. \ref{Neural-Layer-Outputs}(d) and (e) illustrates the outputs of the Tm3 and Tm2 modeled by the DSTMD and ESTMD, respectively. Compared to Fig. \ref{Neural-Layer-Outputs}(d), the outputs of the Tm3 and Tm2 are largely suppressed in Fig. \ref{Neural-Layer-Outputs}(e). This is because the ESTMD uses the laterally inhibited ON and OFF signals to define the outputs of the Tm3 and Tm2 (see (\ref{ESTMD-Tm3-Lateral-Inhibition}) and (\ref{ESTMD-Tm2-Lateral-Inhibition})), while the DSTMD utilizes the ON and OFF signals directly (see (\ref{DS-STMD-Tm3}) and (\ref{DS-STMD-Tm2})). Fig. \ref{Neural-Layer-Outputs}(f) and (g) demonstrate the medulla signals used for the signal correlation in the DSTMD and ESTMD, respectively. As can be seen, four medulla signals are used for the signal correlation in the DSTMD whereas only two medulla signals are utilized in the ESTMD.

Fig. \ref{Neural-Layer-Outputs}(h) and (i) displays the outputs of the proposed DSTMD model and the existing non-directionally selective ESTMD model, respectively. From these two subplots, we can see that both the DSTMD and ESTMD show the strongest response at $x=250$ which is the location of the small moving target. At the other positions, both models exhibit much weaker or even no response. For example, both models demonstrate little response to the tree trunk located between $x =450$ and $x=480$, which is regarded as a large object. The above results indicate that both the DSTMD and ESTMD are only interested in small target motion. 

Comparing Fig. \ref{Neural-Layer-Outputs}(h) with \ref{Neural-Layer-Outputs}(i), we can find that the major difference between the DSTMD and ESTMD is direction selectivity. More precisely, in Fig. \ref{Neural-Layer-Outputs}(h), the DSTMD has eight outputs $E(x,y_0,t_0;\theta)$ corresponding to eight preferred directions $\theta$, $\theta \in \{0, \frac{\pi}{4}, \frac{\pi}{2}, \frac{3\pi}{4}, \pi, \frac{5\pi}{4}, \frac{3\pi}{2}, \frac{7\pi}{4}\}$. However, in Fig. \ref{Neural-Layer-Outputs}(i), the ESTMD only has one output $\tilde{D}(x,y_0,t_0)$ lacking of direction information. To clearly show direction selectivity, the DSTMD outputs to the small target are illustrated in polar coordinate (see the right part of Fig. \ref{Neural-Layer-Outputs}(h)). As can be seen, the DSTMD exhibits the strongest output along direction $\theta = \pi$ which is consistent with the motion direction of the small target. The other seven outputs of the DSTMD decrease as the corresponding direction $\theta$ deviates from the small target motion direction.

\subsection{Tuning Properties}
\label{Tuning-Properties}
\begin{figure}[t!]
	\centering
	\includegraphics[width=0.18\textwidth]{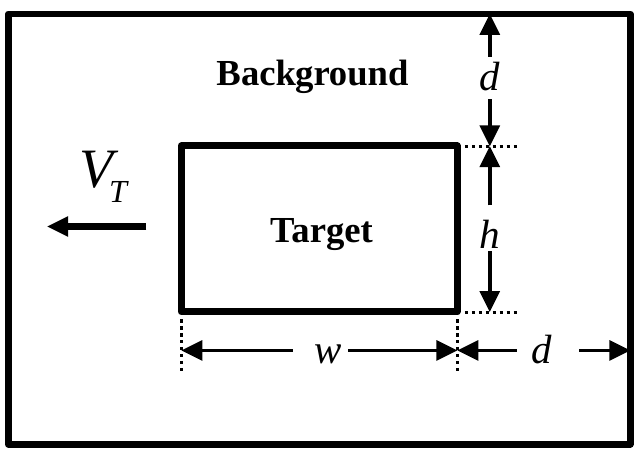}
	\caption{External rectangle and neighboring background rectangle of a small target. The arrow $V_T$ denotes the motion direction of the target. $w$ represents target width while $h$ stands for target height.}
	\label{The-External-Rectangle-and-Neighboring-Background-Rectangle-of-a-Small-Target}
\end{figure}
\begin{figure*}[!t]
	\vspace{-5pt}
	\centering
	\subfloat[]{\includegraphics[width=0.23\textwidth]{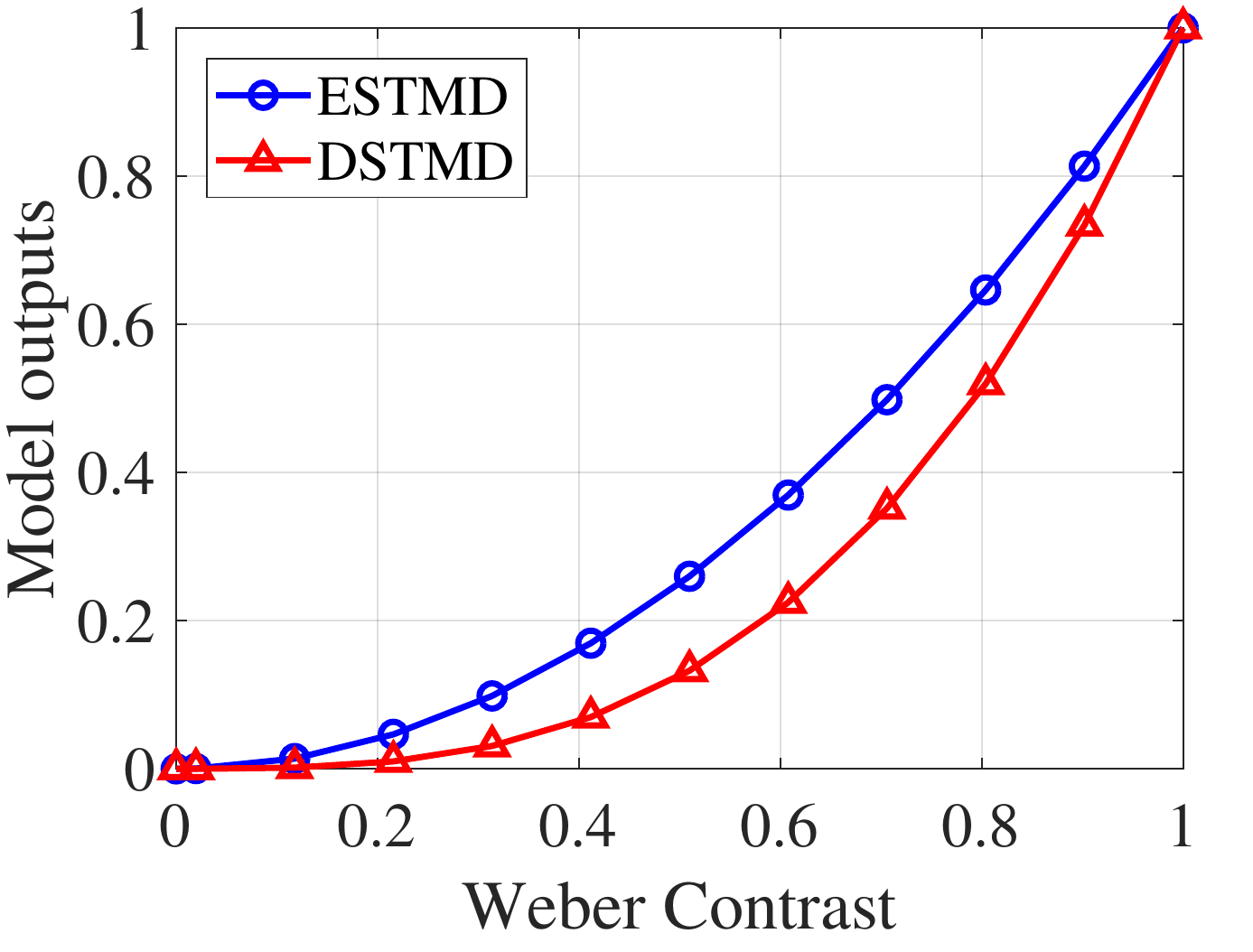}
		\label{Tuning-Properties-LDTB-ESTMD-DS-STMD}}
	\hfil
	\subfloat[]{\includegraphics[width=0.23\textwidth]{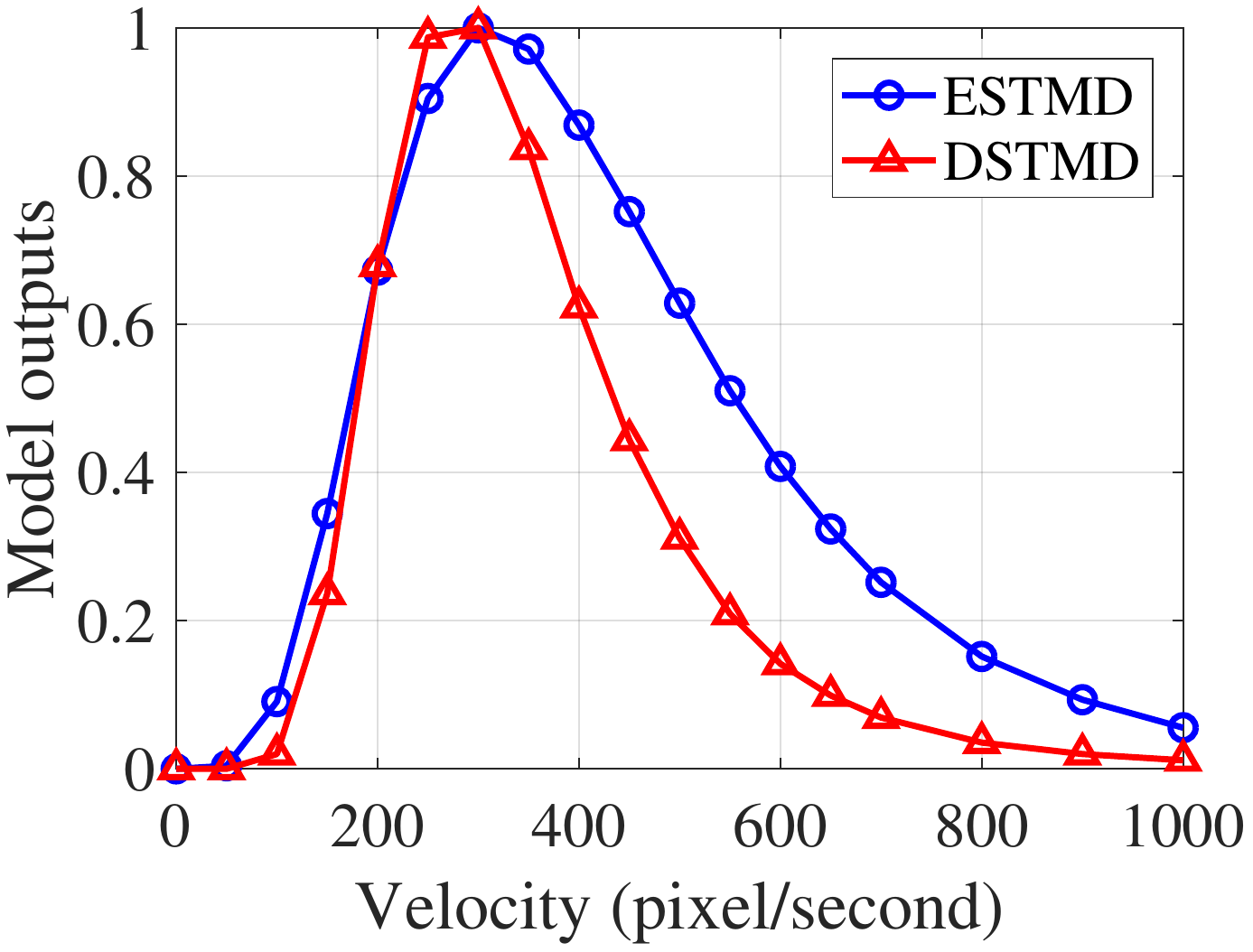}
		\label{Tuning-Properties-Velocity-ESTMD-DS-STMD}}
	\hfil
	\subfloat[]{\includegraphics[width=0.23\textwidth]{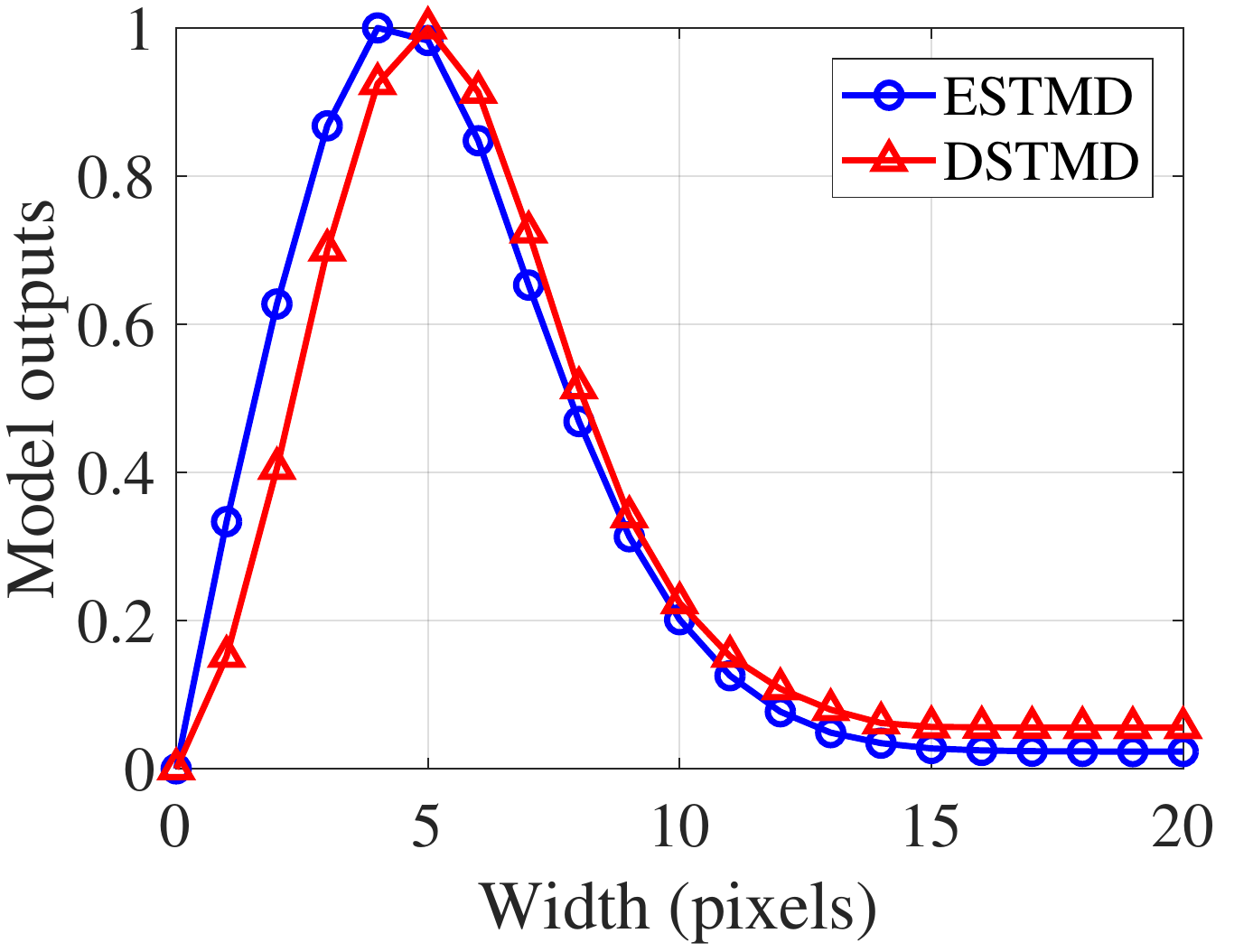}
		\label{Tuning-Properties-Width-ESTMD-DS-STMD}}
	\hfil
	\subfloat[]{\includegraphics[width=0.23\textwidth]{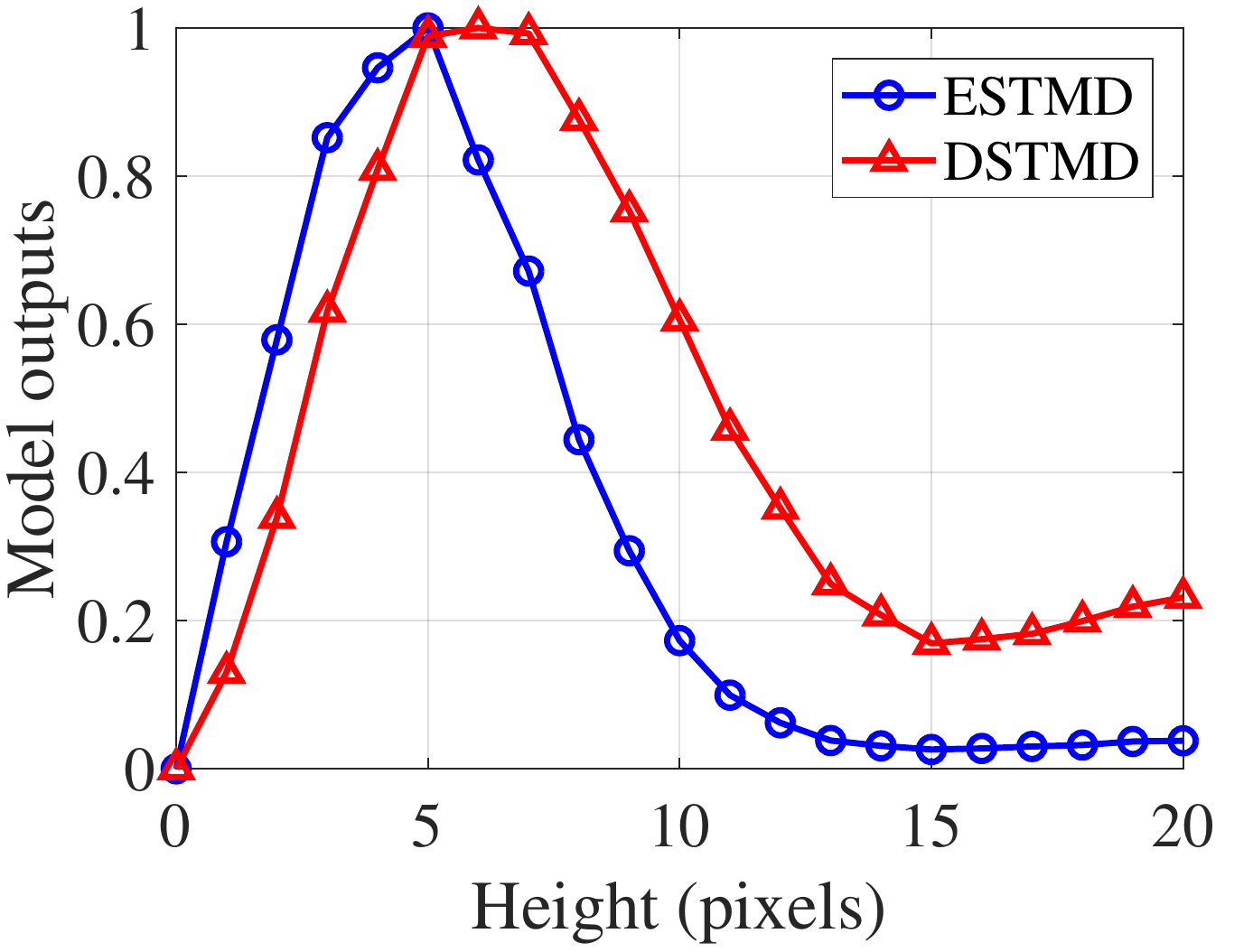}
		\label{Tuning-Properties-Height-ESTMD-DS-STMD}}
	
	\caption{Tuning properties of the proposed neural network (DSTMD) and ESTMD. In each subplot, the horizontal axis represents one of target parameters (Weber Contrast, velocity, width, and height) while the vertical axis denotes normalized model outputs. (a) Weber Contrast tuning curves. (b) Velocity tuning curves. (c) Width tuning curves. (d) Height tuning curves.}
	\label{Tuning-Properties-ESTMD-DS-STMD}
\end{figure*}

\begin{figure*}[!t]
	\vspace{-5pt}
	\centering
	\subfloat[]{\includegraphics[width=0.23\textwidth]{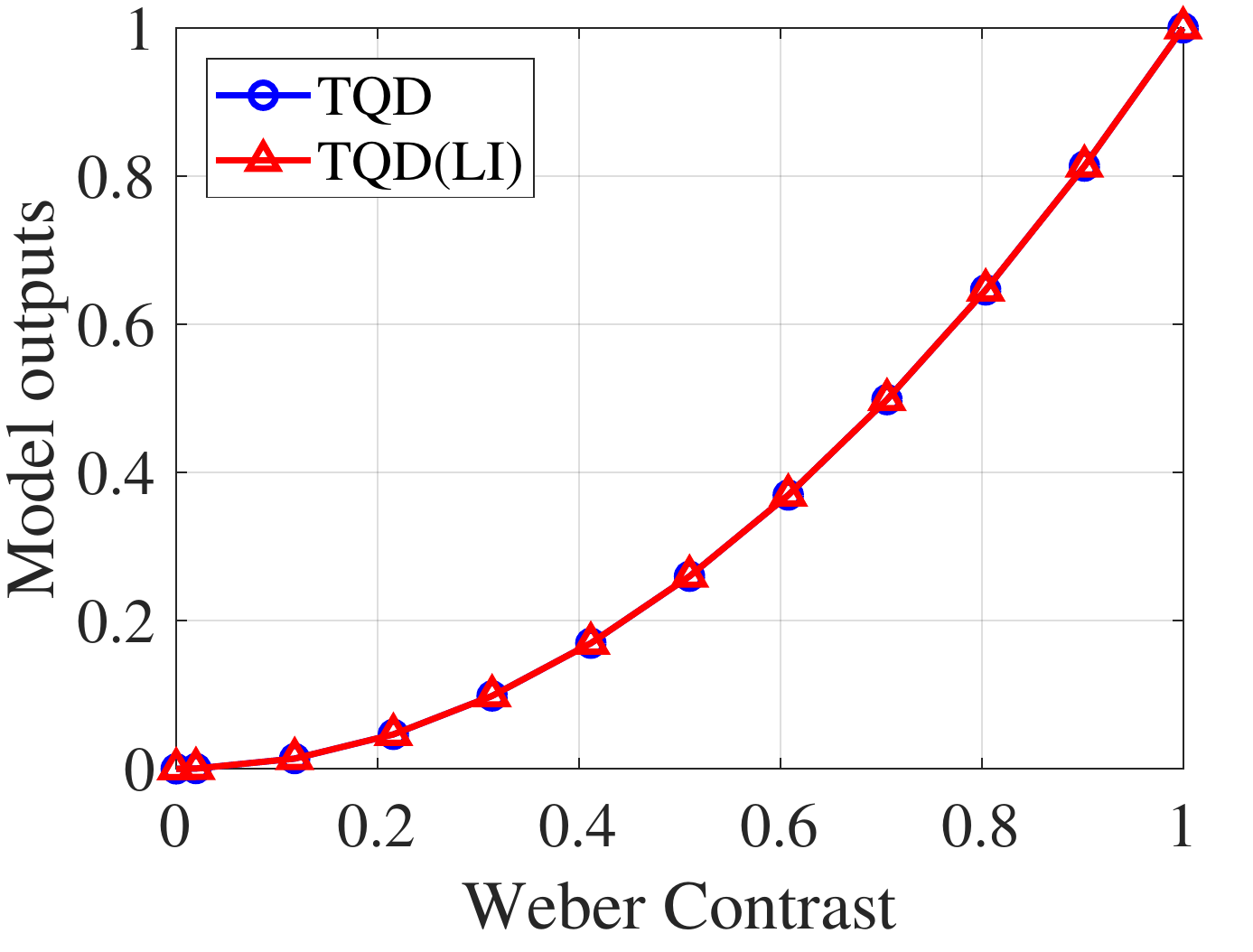}
		\label{Tuning-Properties-LDTB-TQD-TQD-LI}}
	\hfil
	\subfloat[]{\includegraphics[width=0.23\textwidth]{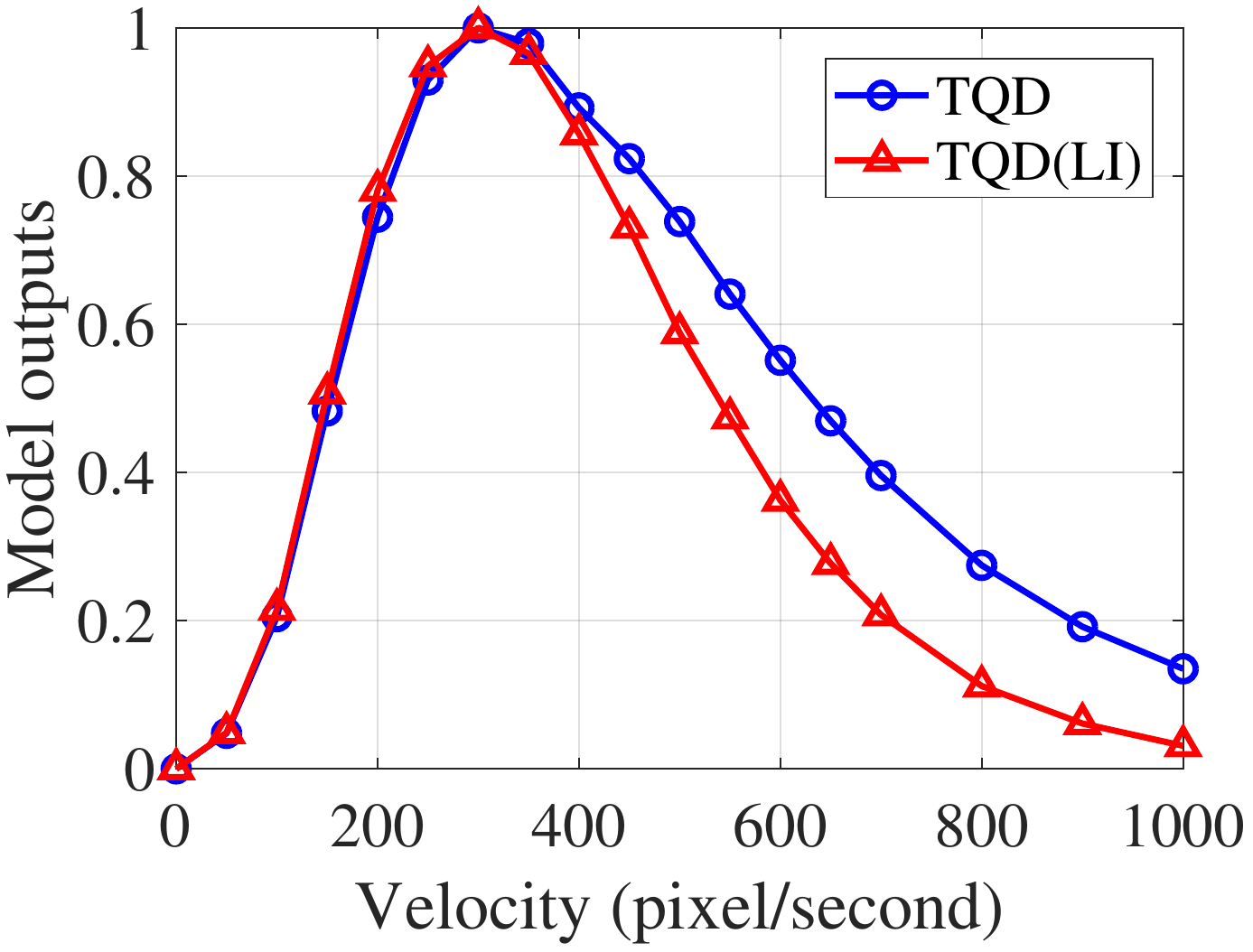}
		\label{Tuning-Properties-Velocity-TQD-TQD-LI}}
	\hfil
	\subfloat[]{\includegraphics[width=0.23\textwidth]{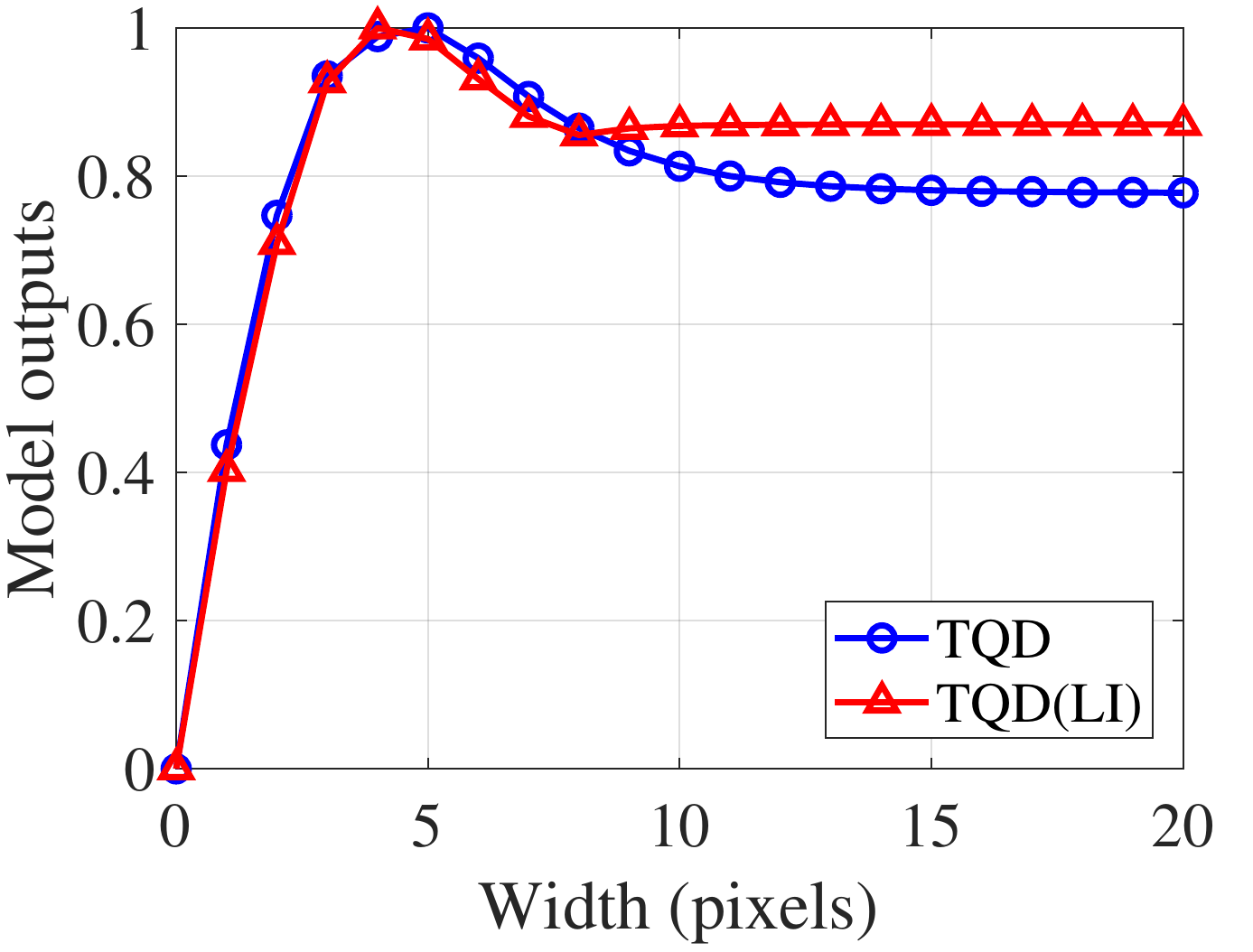}
		\label{Tuning-Properties-Width-TQD-TQD-LI}}
	\hfil
	\subfloat[]{\includegraphics[width=0.23\textwidth]{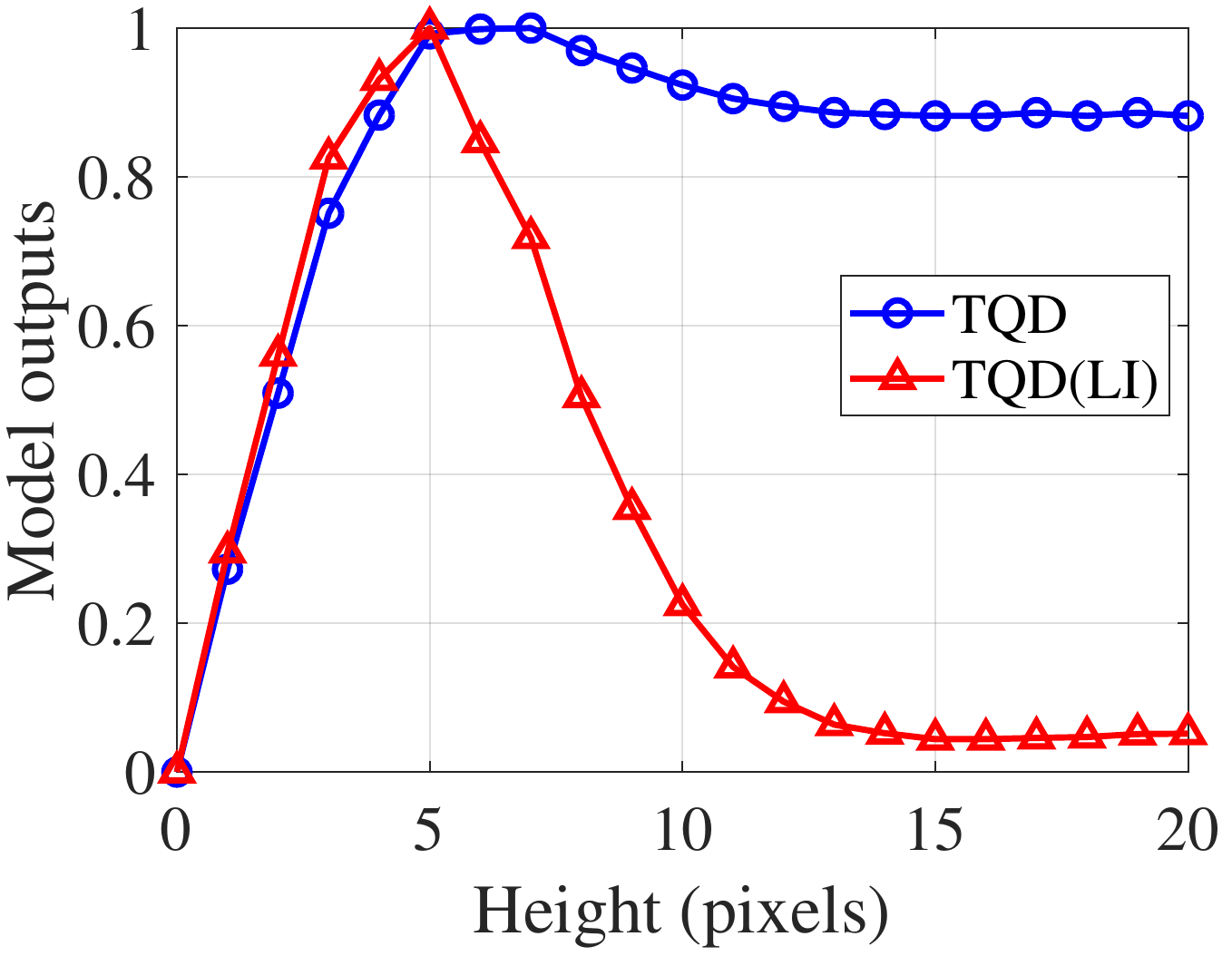}
		\label{Tuning-Properties-Height-TQD-TQD-LI}}
	
	\caption{Tuning properties of the TQD and TQD(LI). In each subplot, the horizontal axis represents one of target parameters (Weber Contrast, velocity, width, and height) while the vertical axis denotes normalized model outputs. (a) Weber Contrast tuning curves. (b) Velocity tuning curves. (c) Width tuning curves. (d) Height tuning curves.}
	\label{Tuning-Properties-TQD-TQD-LI}
\end{figure*}

We test four basic properties of the proposed neural network, including Weber Contrast sensitivity, velocity selectivity, width selectivity and height selectivity. These four properties are basic properties of the STMD neurons and are used to distinguish the  STMD neurons in biology \cite{nordstrom2006small,nordstrom2006insect,barnett2007retinotopic}. Here, Weber Contrast sensitivity refers to that  the STMD neural response increases with the increase of Weber Contrast. Velocity selectivity refers to that the STMD neurons show the strongest response to a specific velocity (optimal velocity). Above or below this optimal velocity will result in the significant decrease of neural responses. Width selectivity and height selectivity are similar to velocity selectivity. 

We first give definitions of Weber Contrast, width and height. As it is shown in Fig. \ref{The-External-Rectangle-and-Neighboring-Background-Rectangle-of-a-Small-Target}, width represents the target length extended parallel to the motion direction while height denotes the target length extended orthogonal to the motion direction. If the size of a target is $w \times h$, the size of its background rectangle is $(w+2d)\times(h+2d)$, where $d$ is a constant which equals to $10$ pixels in this paper. Weber Contrast is defined by the following equation,
\begin{equation}
\text{Weber Contrast} = \frac{|\mu_t - \mu_b|}{255}
\end{equation}
where $\mu_t$ is the average pixel value of the target, $\mu_b$ is the average pixel value in neighboring area around the target.

We perform four experiments to illustrate four basic properties of the proposed neural network. In these four experiments, image sequences which display a small target moving against the white background, are used as the model input. The initial parameters of the small target including luminance, velocity, width and height, are set as $0$, $250$ pixel/second, $5$ pixels and $5$ pixels, respectively. In each experiment, we change one of four target parameters while fix other three parameters, then record corresponding model outputs. The recorded tuning curves are displayed in Fig. \ref{Tuning-Properties-ESTMD-DS-STMD}. 

As it is shown in Fig. \ref{Tuning-Properties-ESTMD-DS-STMD}(a), the outputs of the DSTMD and ESTMD increase with the increase of Weber Contrast, until reach maximum at Weber Contrast $=1$. This reveals that the DSTMD and ESTMD exhibit Weber Contrast sensitivity. In Fig. \ref{Tuning-Properties-ESTMD-DS-STMD}(b), the outputs of the DSTMD and ESTMD all peak at velocity $=300$ pixel/s and decrease significantly when the target velocity is above or below $300$ pixel/s. This suggests that the DSTMD and ESTMD have a preferred velocity and exhibit velocity selectivity. Similar variation trends can be seen in Fig. \ref{Tuning-Properties-ESTMD-DS-STMD}(c) and (d) which reveal the width selectivity and height selectivity of the DSTMD, respectively. 

In the following, we carry out an experiment to demonstrate the hypothesis raised in Section \ref{Modeling-Medulla-Layer}. The hypothesis is that if the medulla neurons which provide signals to the LPTC neurons \cite{lee2015spatio,borst1990direction,borst1995mechanisms}, are laterally inhibited, the LPTC neurons would show strong size selectivity. In order to demonstrate this point, we first adopt TQD model \cite{franceschini1989directionally,eichner2011internal} to simulate the LPTC neurons. Then we use the medulla neuron modelling methods of DSTMD and ESTMD to simulate medulla neurons, respectively. For TQD which receives signals from medulla neurons modeled by DSTMD, we denote it as TQD. For TQD which receives signals from medulla neurons modeled by ESTMD, we denote it as TQD(LI). The only difference between the TQD and TQD(LI) is that medulla neurons providing signals to the TQD(LI), are laterally inhibited. Finally, we test the four basic properties of the TQD and TQD(LI). The recorded tuning curves are presented in Fig. \ref{Tuning-Properties-TQD-TQD-LI}.

As it can be seen from Fig. \ref{Tuning-Properties-TQD-TQD-LI}(a), (b) and (c), the TQD and TQD (LI) display little difference. They all exhibit Weber Contrast sensitivity and velocity selectivity, but do not show the width selectivity. In Fig. \ref{Tuning-Properties-TQD-TQD-LI}(d), although both TQD and TQD (LI) have a local maximum at height $= 5$, they show differences with increasing height. As the continuous increase of the height, the output of the TQD firstly has a slight drop and finally tends to be stable around $0.9$. In contrast, the output of the TQD (LI) decreases significantly and finally tends to be stable around $0.05$. Above results indicate that the TQD(LI) exhibits height selectivity. This contradicts with the biological finding that the LPTC neurons are not size selective \cite{lee2015spatio,borst1990direction,borst1995mechanisms}. To avoid conflict with the biological finding on the LPTC neurons, we adopt the new medulla neuron modeling method and implement the second-order lateral inhibition mechanism on the STMD neuron pathways.

\subsection{Parameter Sensitivity}
\label{Parameter-Sensitivity}
\begin{figure*}[t]
	\vspace{-5pt}
	\centering
	\subfloat[]{\includegraphics[width=0.23\textwidth]{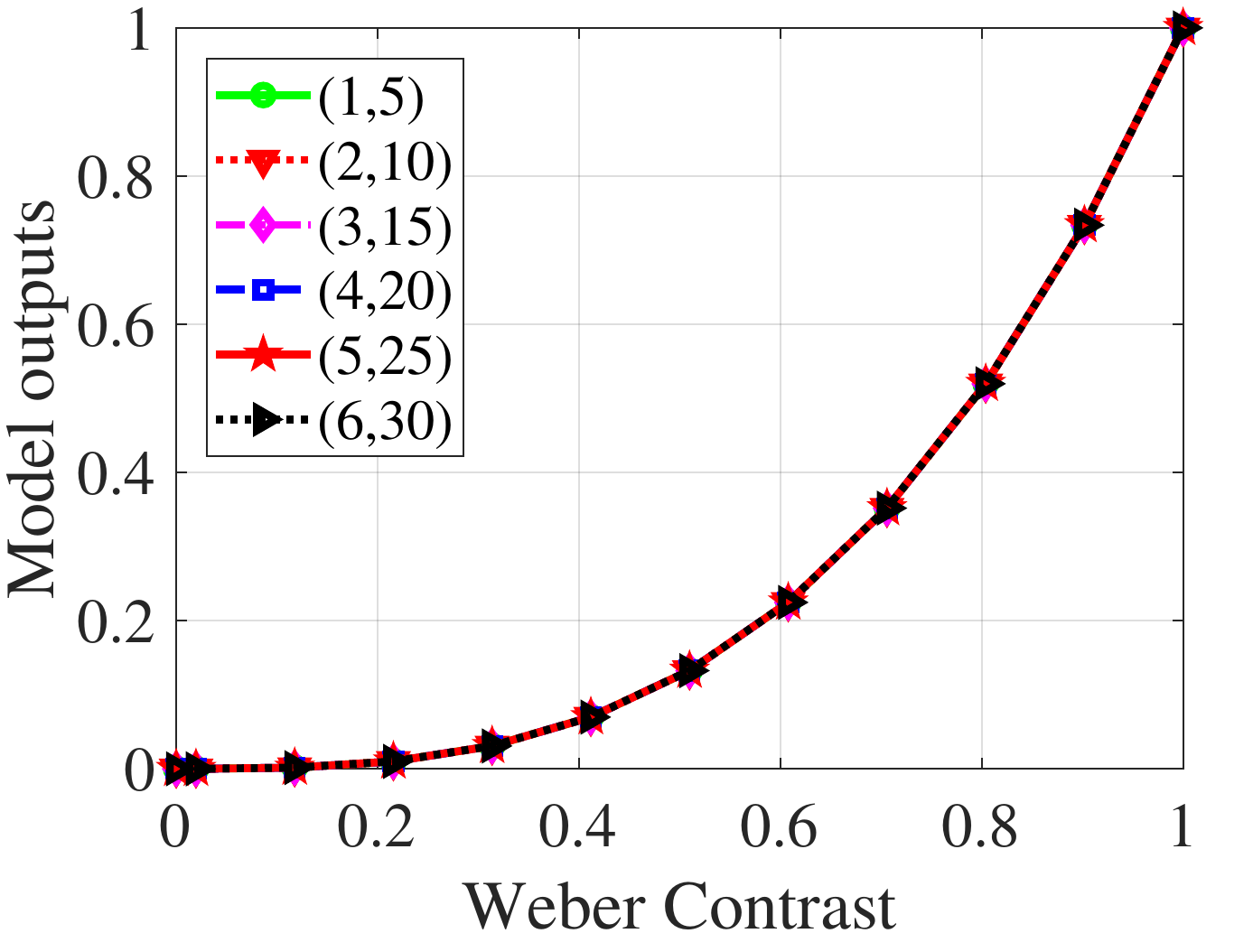}
		\label{Parameter-Sensitivity-Experiment-1-DS-STMD-LDTB-Tuning}}
	\hfil
	\subfloat[]{\includegraphics[width=0.23\textwidth]{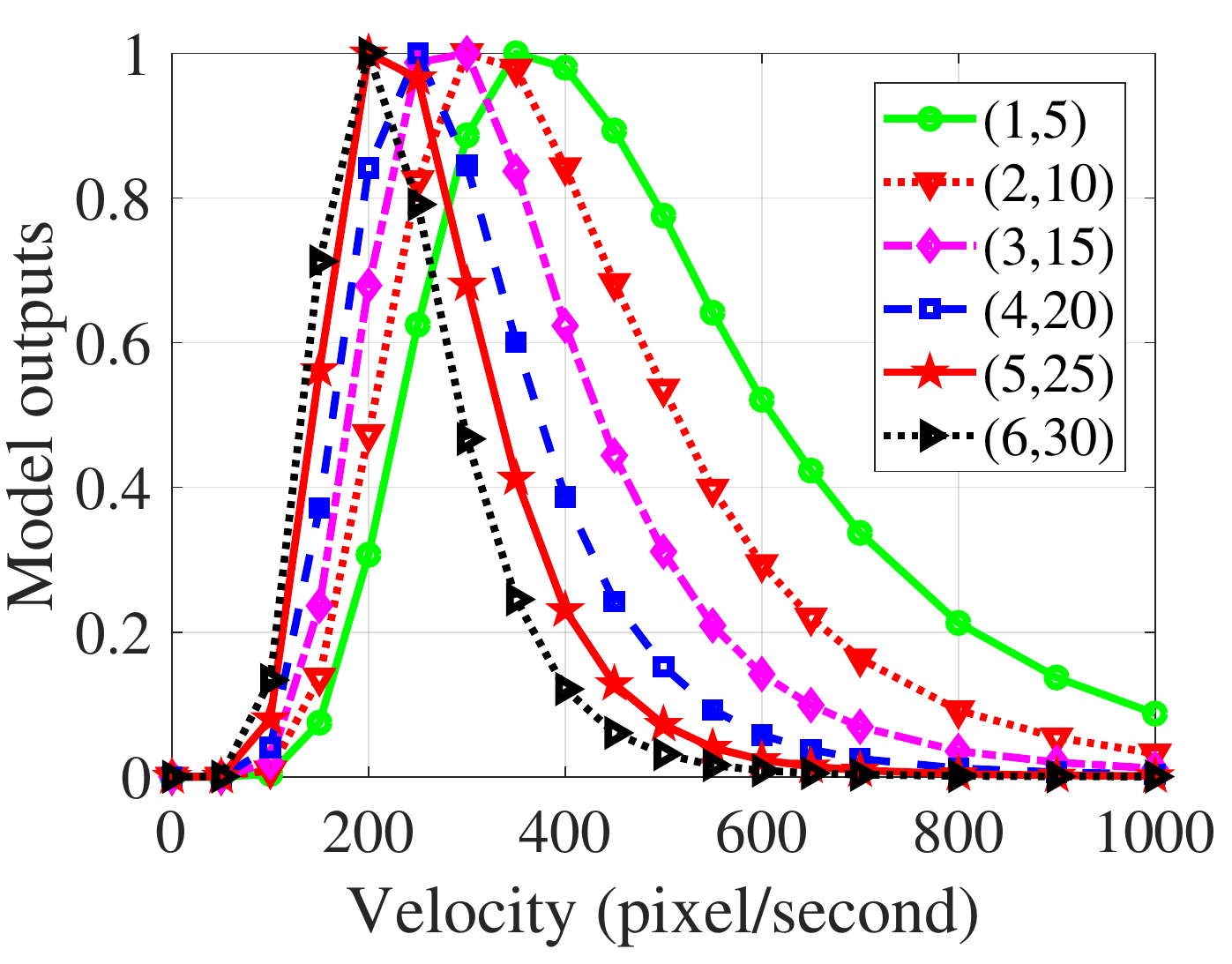}
		\label{Parameter-Sensitivity-Experiment-1-DS-STMD-Velocity-Tuning}}
	\hfil
	\subfloat[]{\includegraphics[width=0.23\textwidth]{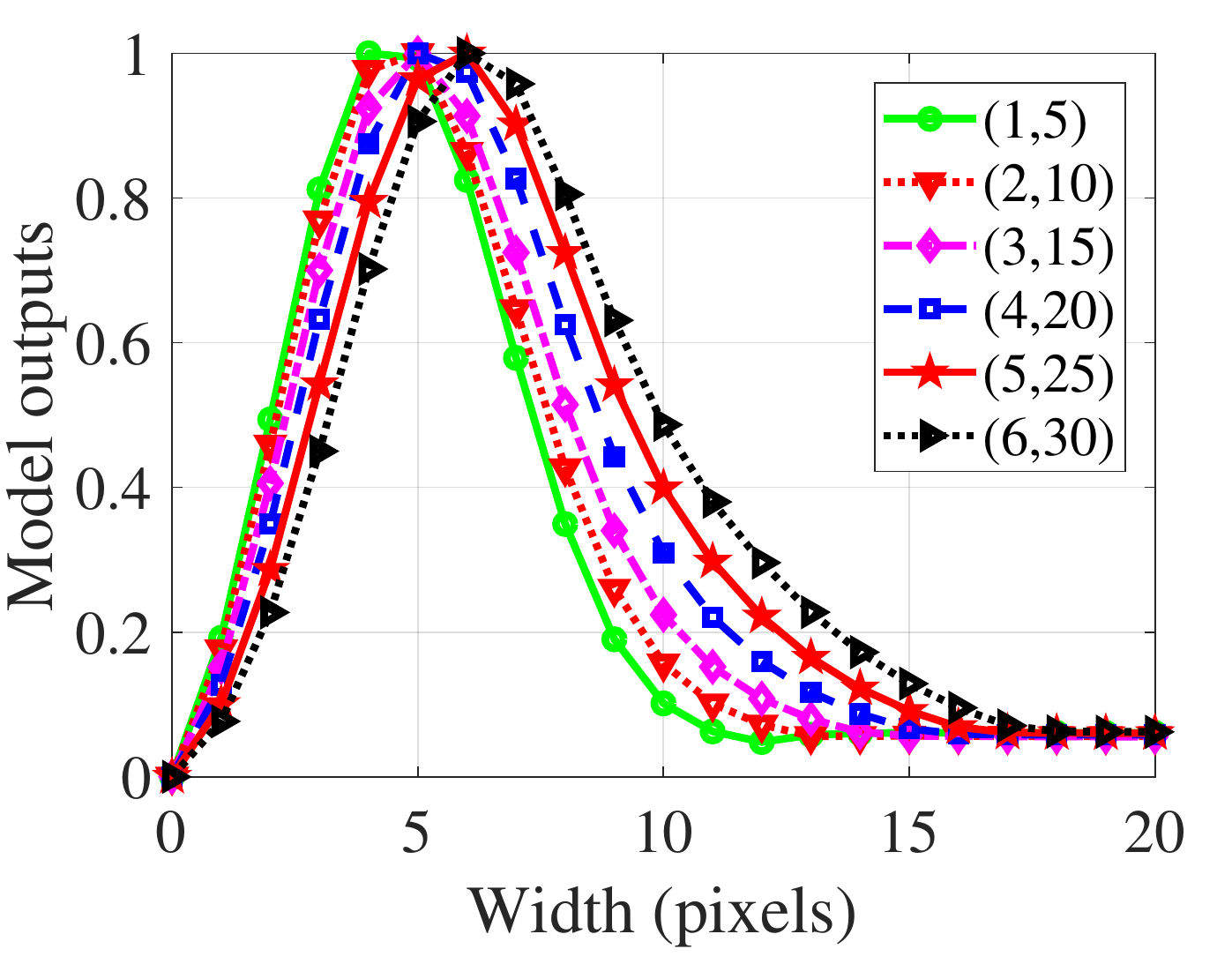}
		\label{Parameter-Sensitivity-Experiment-1-DS-STMD-Width-Tuning}}
	\hfil
	\subfloat[]{\includegraphics[width=0.23\textwidth]{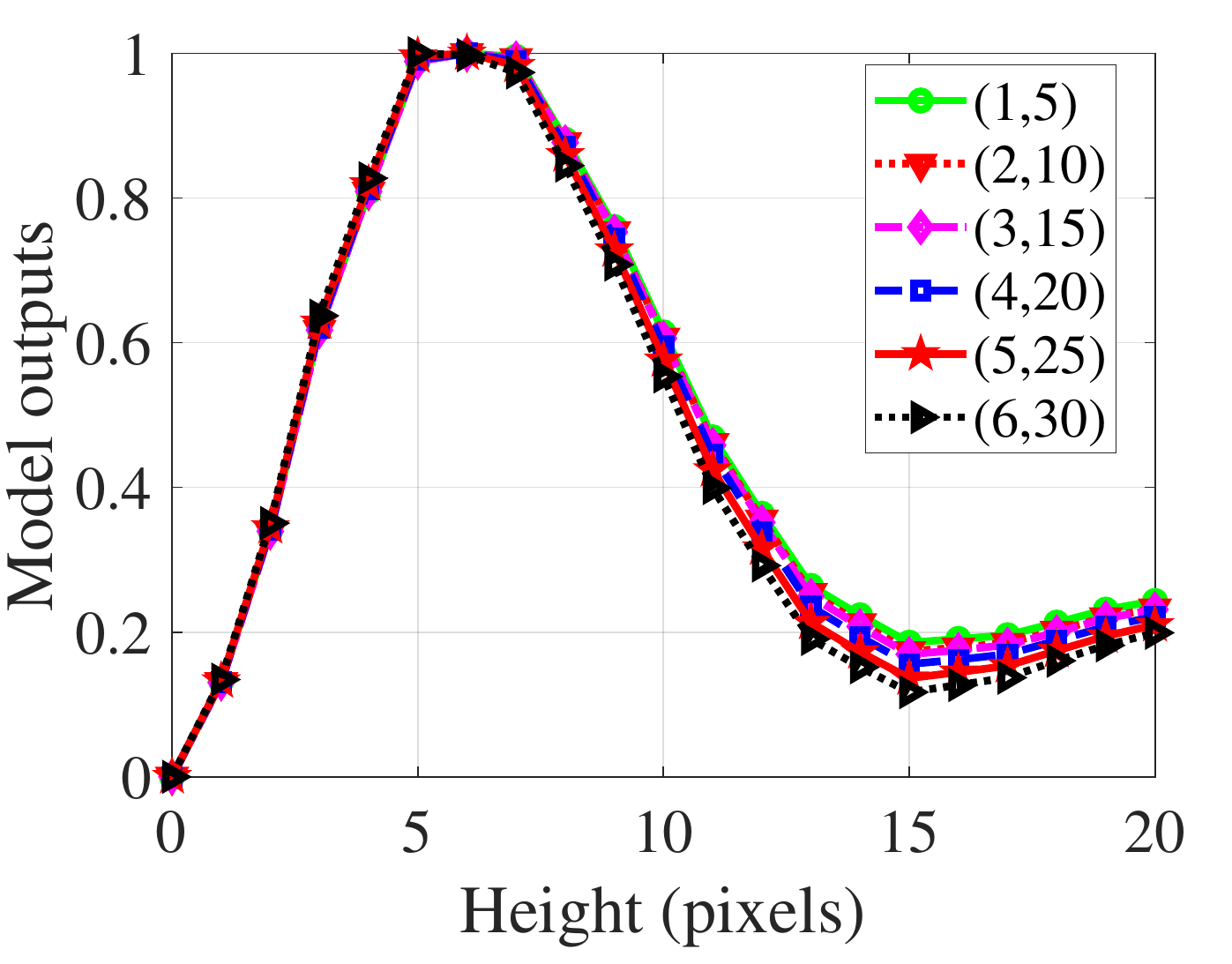}
		\label{Parameter-Sensitivity-Experiment-1-DS-STMD-Height-Tuning}}
	
	\caption{Tuning properties of the proposed neural network under different parameter $(n_4,\tau_4)$. In this experiment, $(n_4,\tau_4)$ is set as $(1,5)$, $(2,10)$, $(3,15)$, $(4,20)$, $(5,25)$, $(6,30)$ while the other parameters are fixed. In each subplot, the horizontal axis represents one of the target parameters (Weber Contrast, velocity, width, and height) while the vertical axis denotes normalized model outputs. (a) Weber Contrast tuning curves. (b) Velocity tuning curves. (c) Width tuning curves. (d) Height tuning curves.}
	\label{Parameter-Sensitivity-Experiment-1-DS-STMD}
\end{figure*}

\begin{figure*}[!t]
	\vspace{-5pt}
	\centering
	\subfloat[]{\includegraphics[width=0.23\textwidth]{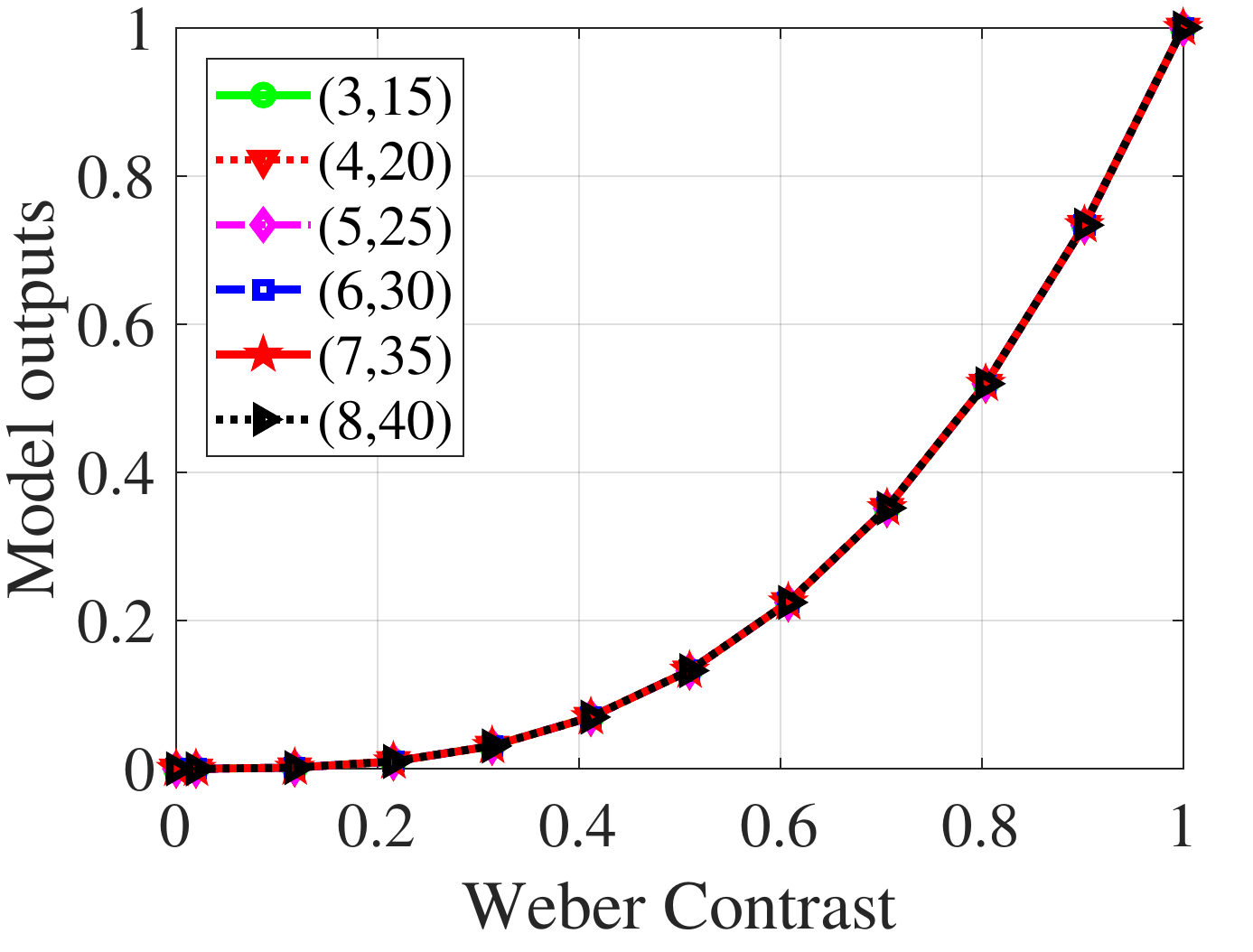}
		\label{Parameter-Sensitivity-Experiment-2-DS-STMD-LDTB-Tuning}}
	\hfil
	\subfloat[]{\includegraphics[width=0.23\textwidth]{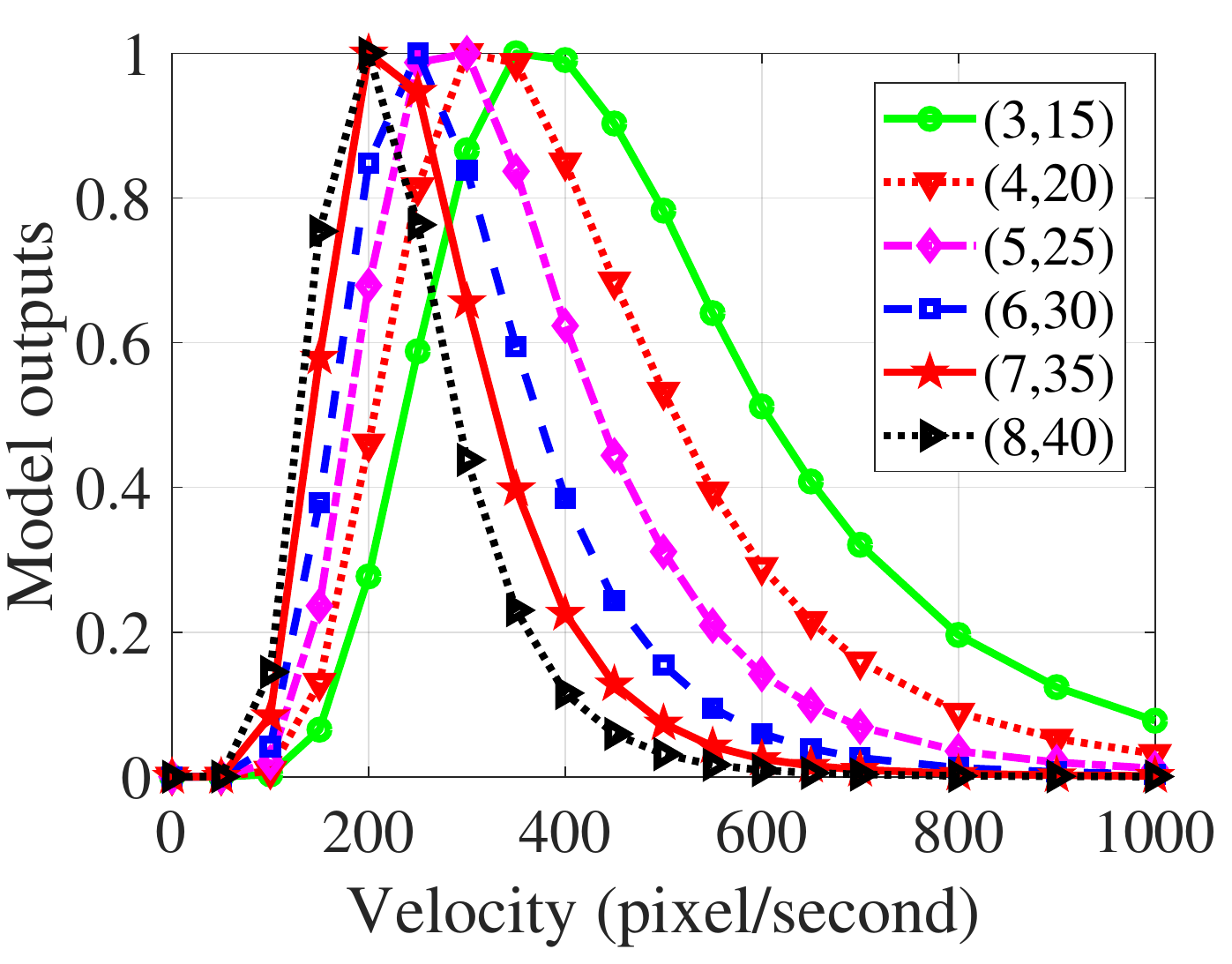}
		\label{Parameter-Sensitivity-Experiment-2-DS-STMD-Velocity-Tuning}}
	\hfil
	\subfloat[]{\includegraphics[width=0.23\textwidth]{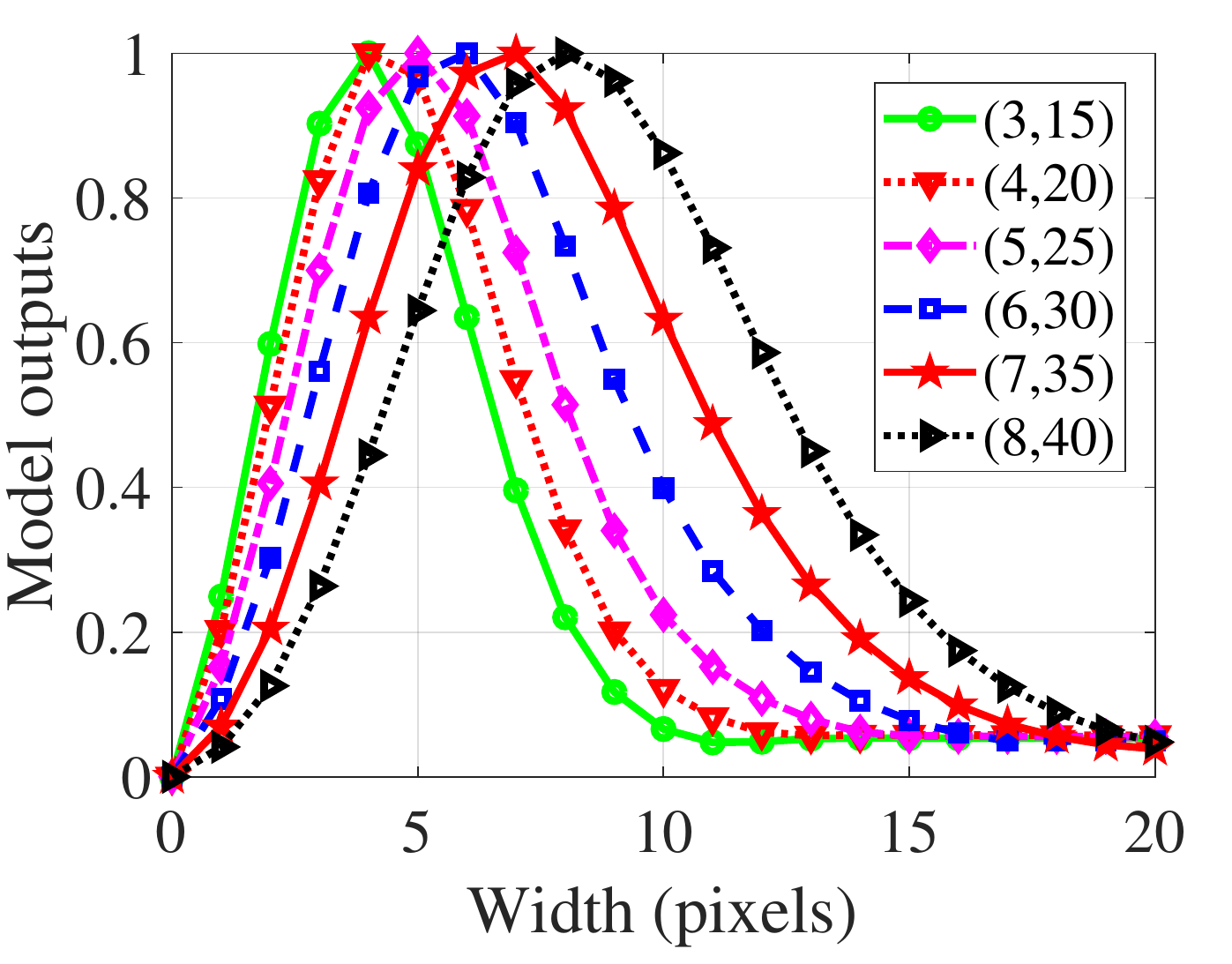}
		\label{Parameter-Sensitivity-Experiment-2-DS-STMD-Width-Tuning}}
	\hfil
	\subfloat[]{\includegraphics[width=0.23\textwidth]{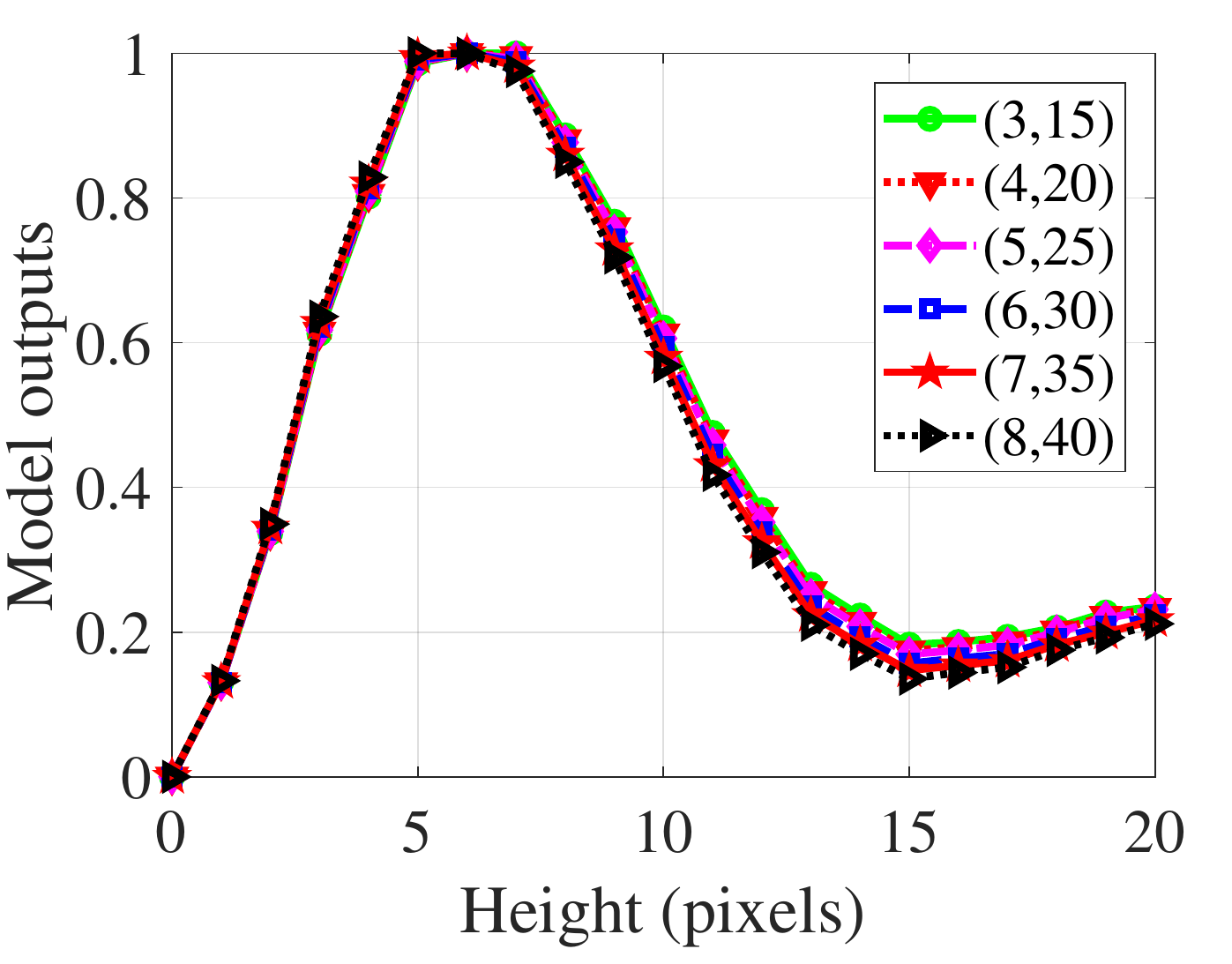}
		\label{Parameter-Sensitivity-Experiment-2-DS-STMD-Height-Tuning}}
	
	\caption{Tuning properties of the proposed neural network under different parameter $(n_5,\tau_5)$. In this experiment, $(n_5,\tau_5)$ is set as $(3,15)$, $(4,20)$, $(5,25)$, $(6,30)$, $(7,35)$, $(8,40)$ while the other parameters are fixed. In each subplot, the horizontal axis represents one of the target parameters (Weber Contrast, velocity, width, and height) while the vertical axis denotes normalized model outputs. (a) Weber Contrast tuning curves. (b) Velocity tuning curves. (c) Width tuning curves. (d) Height tuning curves.}
	\label{Parameter-Sensitivity-Experiment-2-DS-STMD}
\end{figure*}

\begin{figure*}[!t]
	\centering
	\subfloat[]{\includegraphics[width=0.23\textwidth]{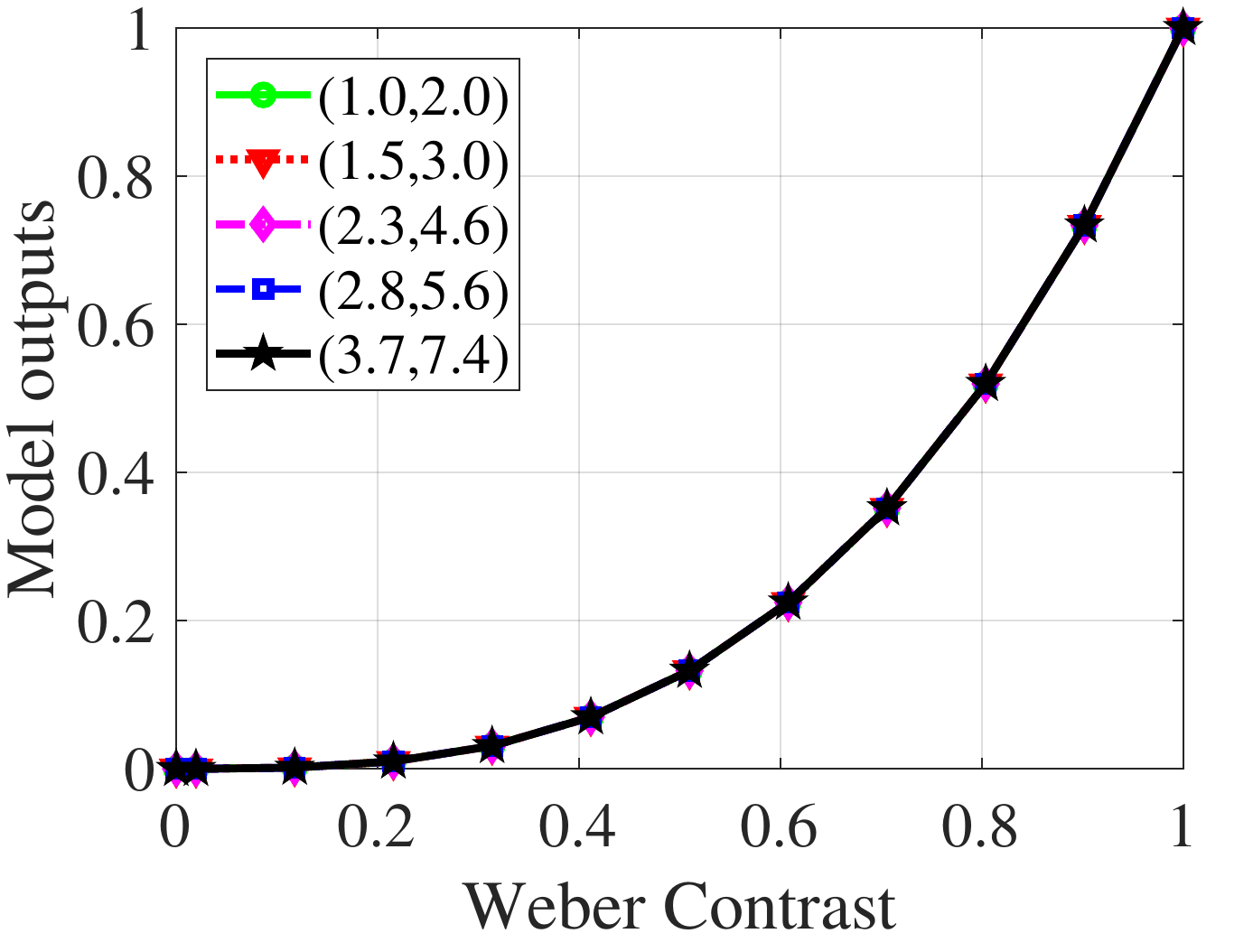}
		\label{Parameter-Sensitivity-Experiment-3-DS-STMD-LDTB-Tuning}}
	\hfil
	\subfloat[]{\includegraphics[width=0.23\textwidth]{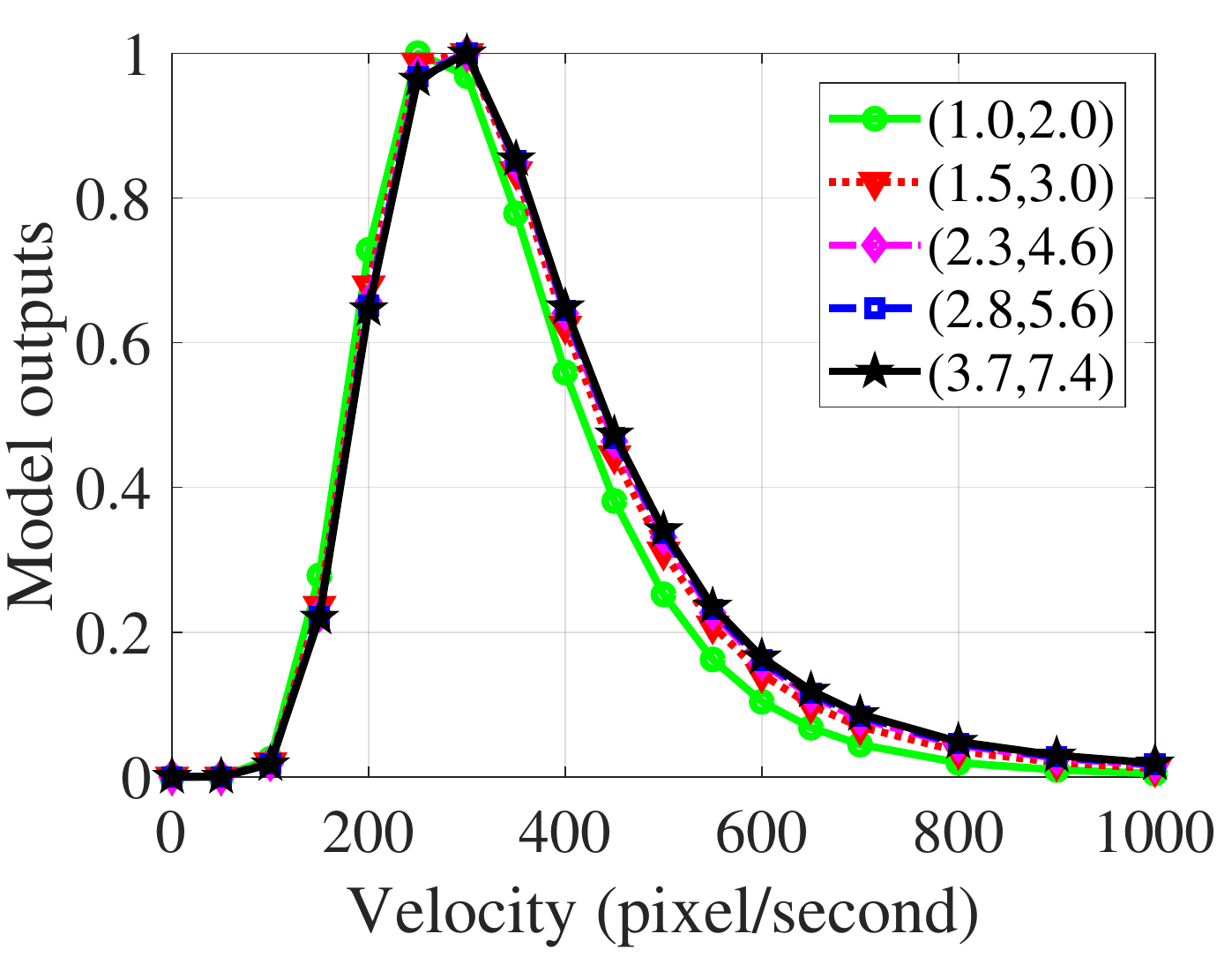}
		\label{Parameter-Sensitivity-Experiment-3-DS-STMD-Velocity-Tuning}}
	\hfil
	\subfloat[]{\includegraphics[width=0.23\textwidth]{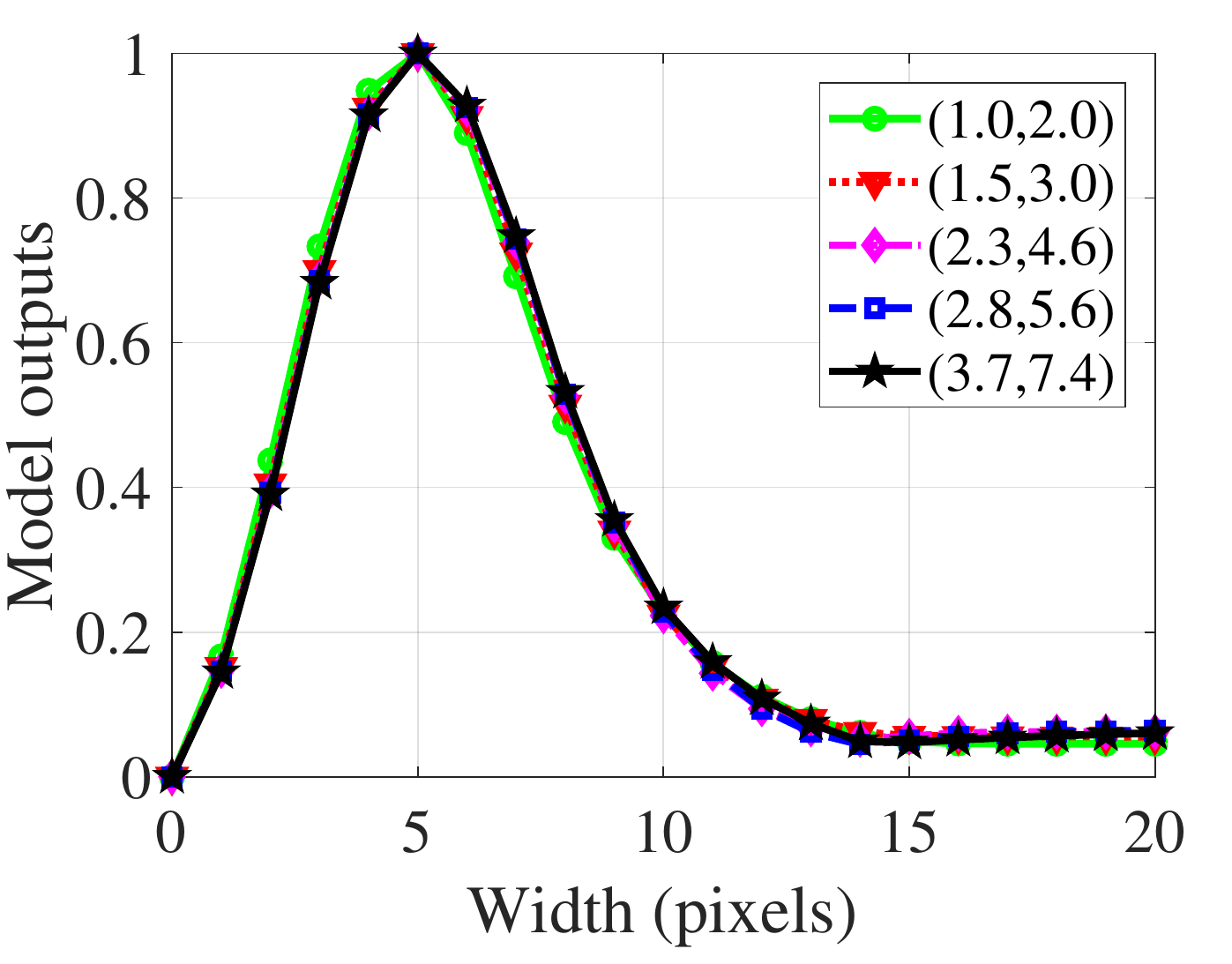}
		\label{Parameter-Sensitivity-Experiment-3-DS-STMD-Width-Tuning}}
	\hfil
	\subfloat[]{\includegraphics[width=0.23\textwidth]{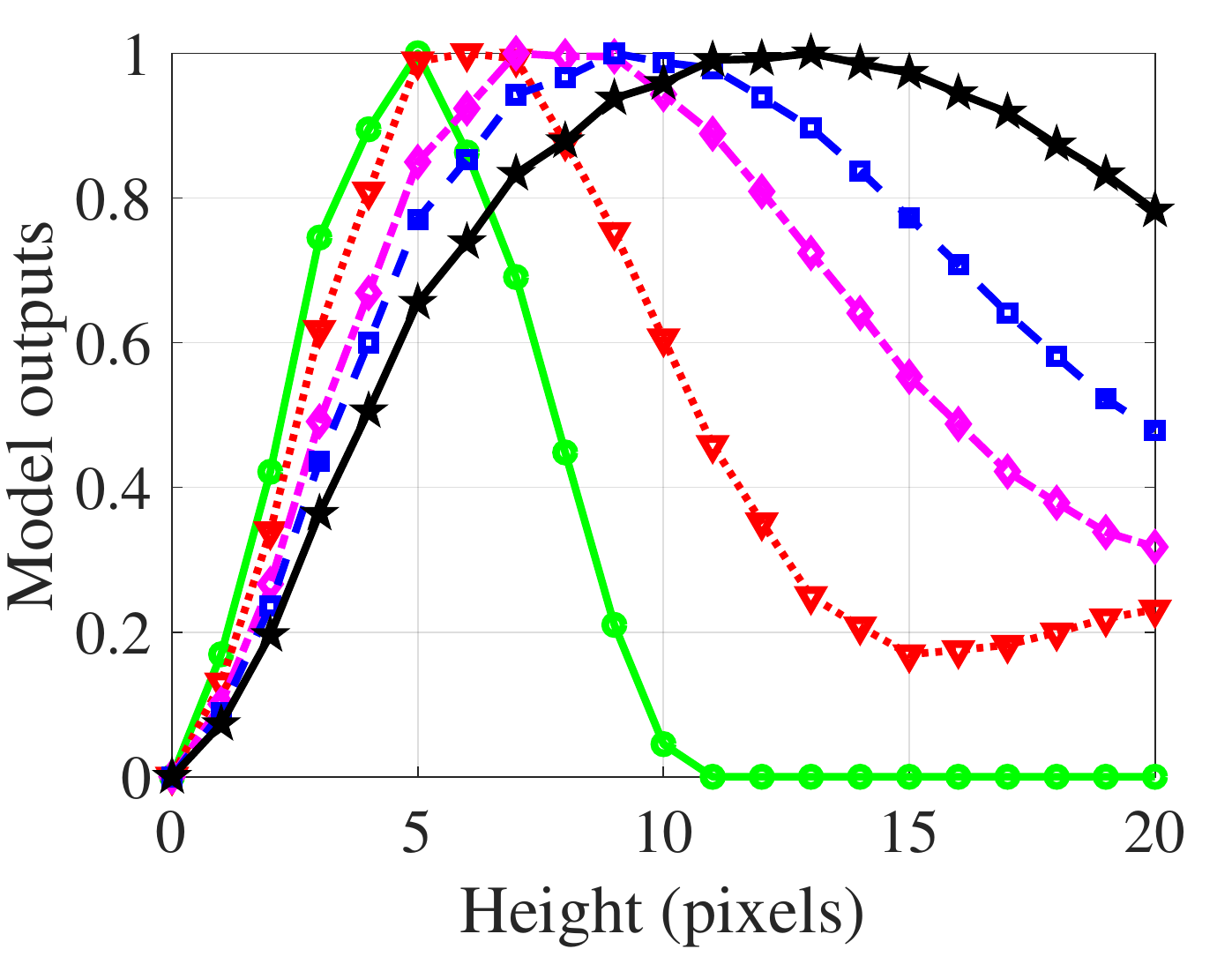}
		\label{Parameter-Sensitivity-Experiment-3-DS-STMD-Height-Tuning-P6-S}}
	
	\caption{Tuning properties of the proposed neural network under different parameter $(\sigma_4, \sigma_5)$. In this experiment, $(\sigma_4, \sigma_5)$ is set as $(1.0,2.0)$, $(1.5,3.0)$, $(2.3,4.6)$, $(2.8,5.6)$, $(3.7,7.4)$ while the other parameters are fixed. In each subplot, the horizontal axis represents one of the target parameters (Weber Contrast, velocity, width, and height) while the vertical axis denotes normalized model outputs. (a) Weber Contrast tuning curves. (b) Velocity tuning curves. (c) Width tuning curves. (d) Height tuning curves.}
	\label{Parameter-Sensitivity-Experiment-3-DS-STMD}
\end{figure*}

\begin{figure}[t!]
	\centering
	\includegraphics[width=0.35\textwidth]{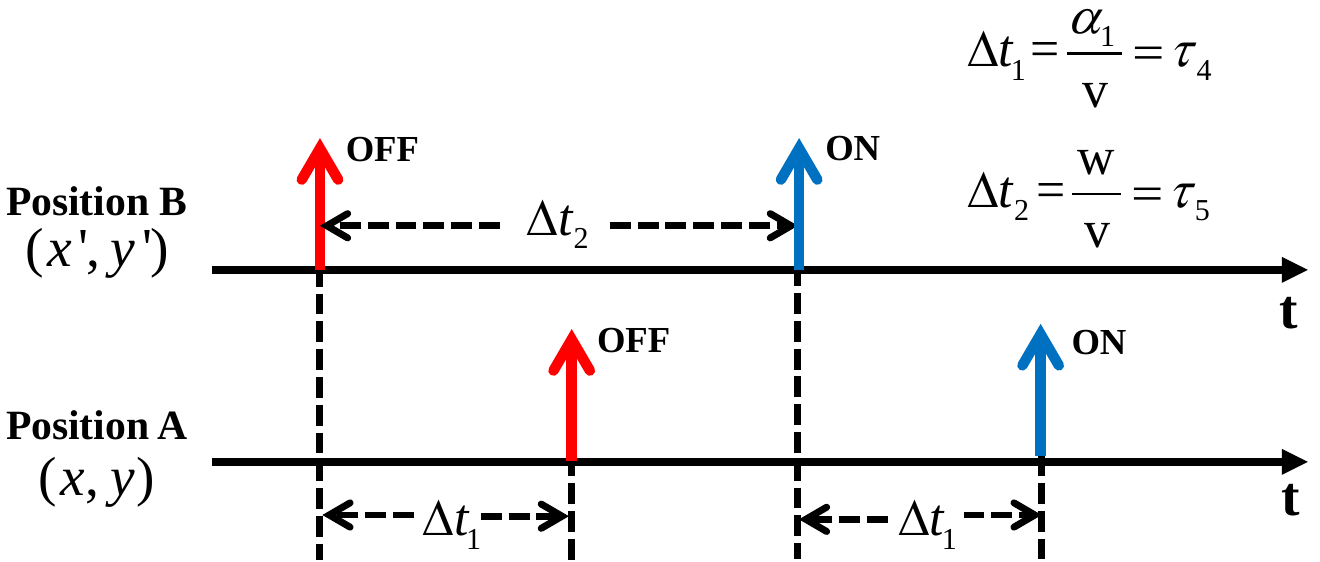}
	\caption{Schematic illustration of the luminance changes of the position A and B when a dark target successively passes position B $(x',y')$ and A $(x,y)$. The red arrow denotes luminance decrease signal (OFF signal) while the blue arrow represents luminance increase signal (ON signal). Let $\alpha_1$, $w$ and $v$ stand for the distance between position A and B, target width and velocity, respectively. Then we have $\Delta t_1 = \frac{\alpha_1}{v}, \Delta t_2 = \frac{w}{v}$.}
	\label{Schematic-of-Luminance-Changes}
\end{figure}

In the last section, we have demonstrated that the proposed neural network shows four basic properties, i.e., Weber Contrast sensitivity, velocity selectivity, width selectivity and height selectivity. In this section, we further evaluate the impacts of three sets of parameters, including $(n_4, \tau_4)$, $(n_5, \tau_5)$ and $(\sigma_4, \sigma_5)$, on the four basic properties. These three sets of parameters are defined in (\ref{ESTMD-Mdeulla-Lateral-Inhibition-Kernel-W2-2}) and (\ref{DS-STMD-Signal-Correlation}).

We conduct three experiments to assess the effects of these three sets of parameters, respectively. In each experiment, we change one set of parameters while keep other two sets of parameters at their initially assigned value [see Table \ref{Table-Parameter-STMD}], then record corresponding model outputs. In the first experiment, $(n_4, \tau_4)$ is set as 	$(1,5)$, $(2,10)$, $(3,15)$, $(4,20)$, $(5,25)$, $(6,30)$. In the second experiment, $(n_5, \tau_5)$ is set as $(3,15)$, $(4,20)$, $(5,25)$, $(6,30)$, $(7,35)$, $(8,40)$. In the third experiment, $(\sigma_4, \sigma_5)$ is set as $(1.0,2.0)$, $(1.5,3.0)$, $(2.3,4.6)$, $(2.8,5.6)$, $(3.7,7.4)$. The recorded tuning curves of the proposed neural network under different parameter settings, are  presented in Fig. \ref{Parameter-Sensitivity-Experiment-1-DS-STMD}--\ref{Parameter-Sensitivity-Experiment-3-DS-STMD}.

In the first and second experiment, we illustrate that the parameter $(n_4,\tau_4)$ and $(n_5,\tau_5)$ have large impact on the velocity selectivity and width selectivity, but show little effect on the Weber Contrast sensitivity and height selectivity. More precisely, from Fig. \ref{Parameter-Sensitivity-Experiment-1-DS-STMD}(a) and (d), we can see that the increase of $(n_4,\tau_4)$ have not induced any significant changes of the Weber Contrast tuning curve and the height tuning curve. However, with the increase of $(n_4,\tau_4)$, as shown in Fig. \ref{Parameter-Sensitivity-Experiment-1-DS-STMD}(b) and (c), the peak velocity decreases while the peak width increases. In Fig. \ref{Parameter-Sensitivity-Experiment-2-DS-STMD}, the parameter $(n_5,\tau_5)$ has similar effect with $(n_4,\tau_4)$ on the four basic properties.

The reasons for the above results are---in the proposed neural network, $\tau_4$ and $\tau_5$ are positively correlated to $\frac{\alpha_1}{v}$ and $\frac{w}{v}$, respectively, where $\alpha_1$, $v$ and $w$ stand for the distance between position A and B, the peak velocity and the peak width, respectively. Once $\alpha_1$ is given, the increase of $\tau_4$ (or $\tau_5$) will result in the decrease of the peak velocity $v$ and the increase of the peak width $w$. 

We further explain why $\tau_4$ and $\tau_5$ are positively correlated to $\frac{\alpha_1}{v}$ and $\frac{w}{v}$. In Fig. \ref{Schematic-of-Luminance-Changes}, we present the luminance changes of position A and B when a dark small target moves from B to A. In the equation (\ref{DS-STMD-Signal-Correlation}), the DSTMD uses four medulla signals from position A $(x,y)$ and B $(x',y')$ to define the output of STMD neurons. Combining Fig. \ref{Schematic-of-Luminance-Changes} with the equation (\ref{DS-STMD-Signal-Correlation}), we point out that these four medulla signal are: 1) ON signal of position A $(x,y)$, corresponding to $S^{^{Tm3}}(x,y,t)$; 2) ON signal of position B $(x',y')$ with time delay order $n_4$ and time delay length $\tau_4$, corresponding to $S^{^{Mi1}}_{_{D(n_{_4},\tau_{_4})}}(x',y',t)$; 3) OFF signal of position A $(x,y)$ with time delay order $n_5$ and time delay length $\tau_5$, corresponding to $S^{^{Tm1}}_{_{D(n_{_5},\tau_{_5})}}(x,y,t)$; 4) OFF signal of position B $(x',y')$ with time delay order $n_6$ and time delay length $\tau_6$, corresponding to $S^{^{Tm1}}_{_{D(n_{_6},\tau_{_6})}}(x',y',t)$. In the DSTMD, we set $\tau_4$, $\tau_5$ and $\tau_6$ as $\Delta t_1$, $\Delta t_2$ and $\Delta t_1 + \Delta t_2$, respectively. Since $\Delta t_1 = \frac{\alpha_1}{v}$ and $\Delta t_2 = \frac{w}{v}$, then we have $\tau_4 = \frac{\alpha_1}{v}$ and $\tau_5 = \frac{w}{v}$. That is, $\tau_4$ and $\tau_5$ are positively correlated to $\frac{\alpha_1}{v}$ and $\frac{w}{v}$, respectively.

In the third experiment, we demonstrate that the parameter $(\sigma_4, \sigma_5)$ has large impact on the height selectivity, but shows little effect on the other three properties. As it can be seen from Fig. \ref{Parameter-Sensitivity-Experiment-3-DS-STMD}(a)-(c), the tuning curves have little changes with the increase of $(\sigma_4, \sigma_5)$; in contrast, the peak height of the height tuning curve increases, as presented in Fig. \ref{Parameter-Sensitivity-Experiment-3-DS-STMD}(d). Here, we point out that the peak height is positively correlated to the size of the excitatory region of the lateral inhibition mechanism (see Fig. \ref{Schematic-Correlation-Mechanism-and-Lateral-Inhibition-Mechanism}(b)). In the proposed neural network, the size of the excitatory region is determined by $\sigma_4$ and $\sigma_5$, where the higher $(\sigma_4, \sigma_5)$ means the larger excitatory region, i.e, the larger peak height.

\subsection{Direction Selectivity and Motion Direction Estimation}
\label{Direction-Selcetivity-and-Motion-Direction-Estimation}
\begin{figure}[!t]
	\centering
	\includegraphics[width=0.35\textwidth]{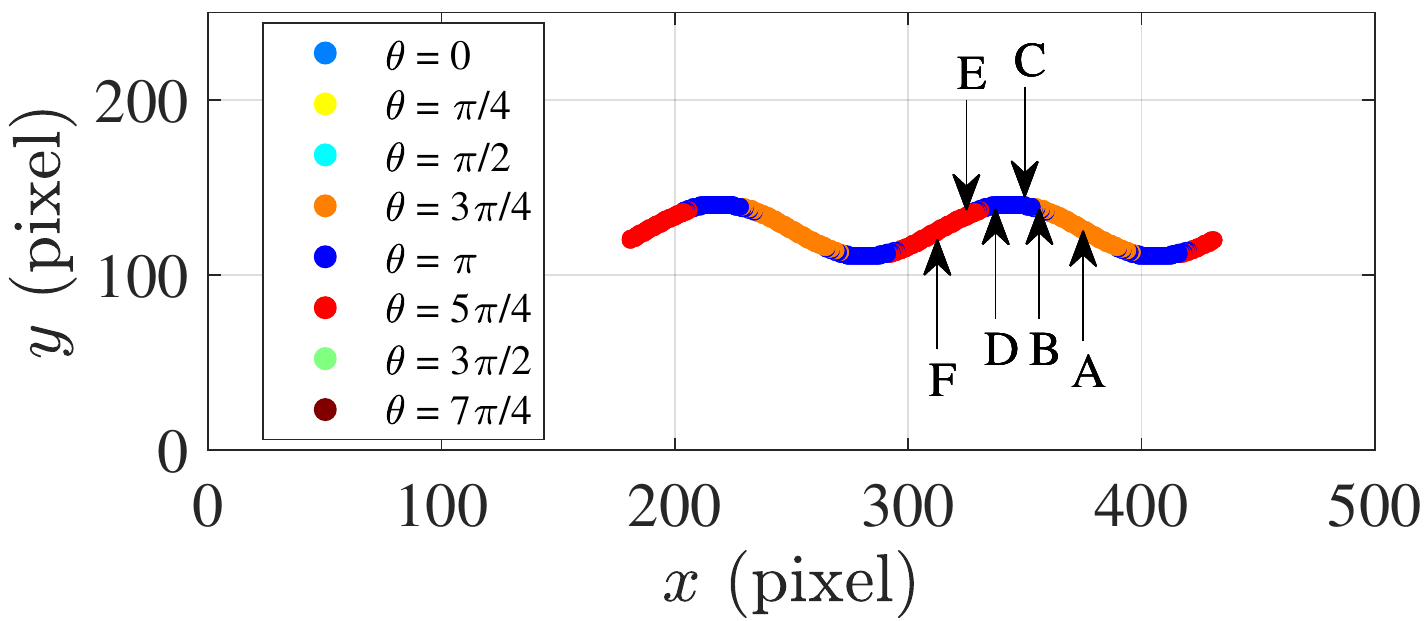}
	\caption{Motion trace of the small target where color denotes the direction of the strongest output of the proposed neural network.}
	\label{Direction-Selectivity-Target-Motion-Trace}
\end{figure}

In this section, we illustrate how the proposed neural network encode motion directions of small targets. In the experiment, an image sequence which displays a small target moving against the white background, is used as the network input. The luminance and size of the small target are set as $0$ and $5 \times 5$ pixels, respectively. The coordinate of the small target at time $t$ is $(500 - 250 \cdot \frac{t+300}{1000}, 125+15 \cdot \sin(4\pi \frac{t+300}{1000})), t \in [0, 1000]$ ms. Fig. \ref{Direction-Selectivity-Target-Motion-Trace} presents the motion trace of the small target. The motion direction of the small target varies between $142.98^{\circ}$ and $217.01^{\circ}$ when it moves along this motion trace.

\begin{figure*}[!t]
	\vspace{-10pt}
	\centering
	\subfloat[]{\includegraphics[width=0.125\textwidth]{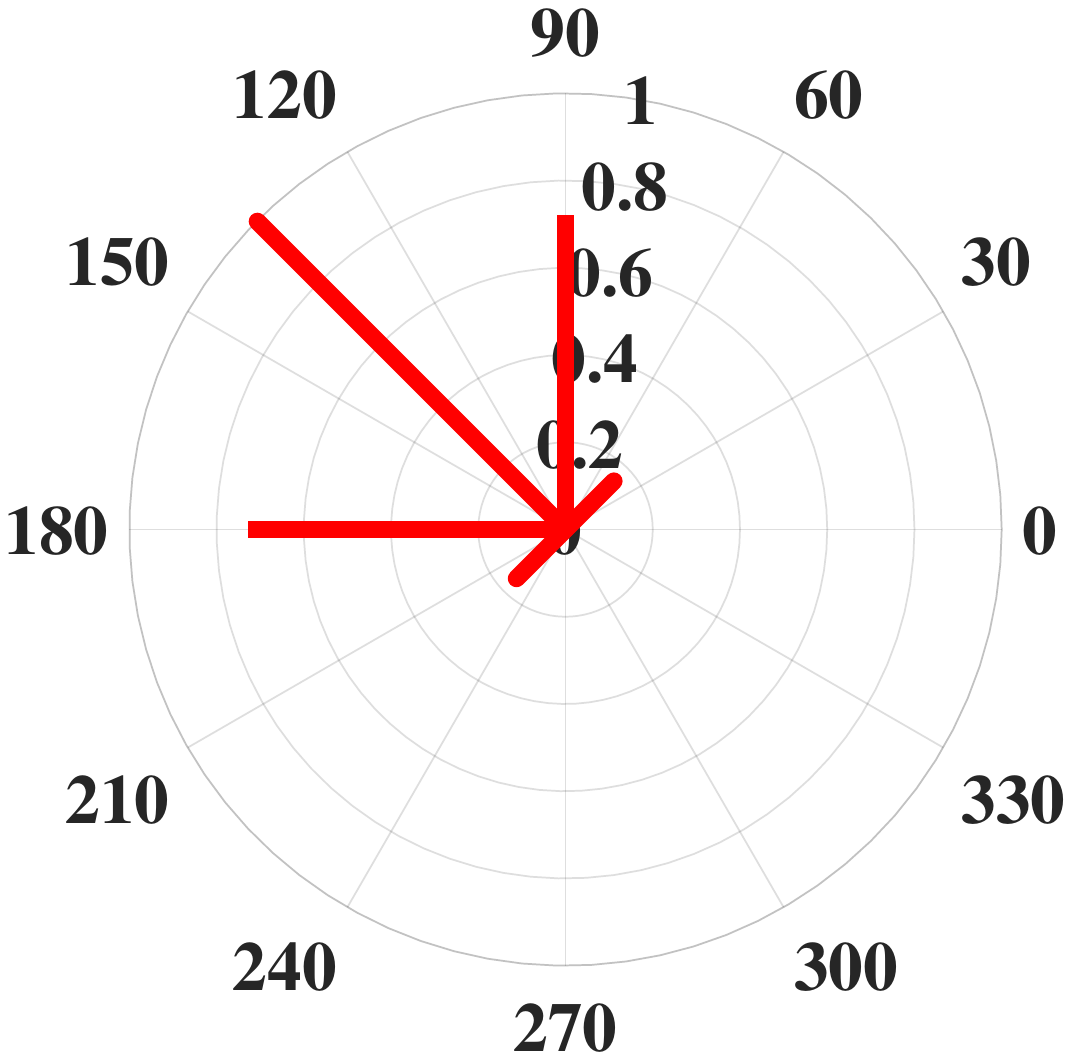}
		\label{Direction-Selectivity-Neural-Responses-Polar-X-123-Y-373}}
	\hfil
	\subfloat[]{\includegraphics[width=0.125\textwidth]{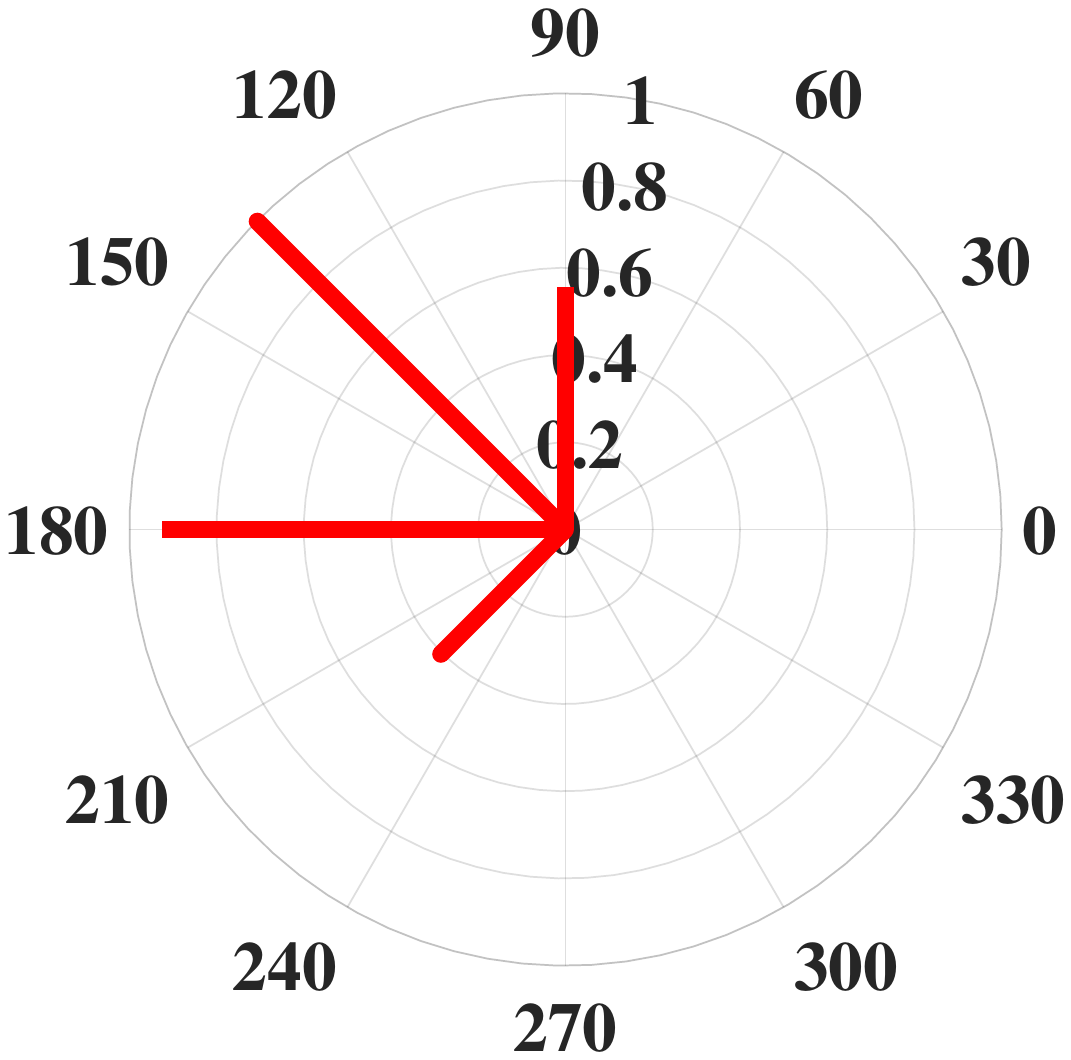}
		\label{Direction-Selectivity-Neural-Responses-Polar-X-115-Y-360}}
	\hfil
	\subfloat[]{\includegraphics[width=0.125\textwidth]{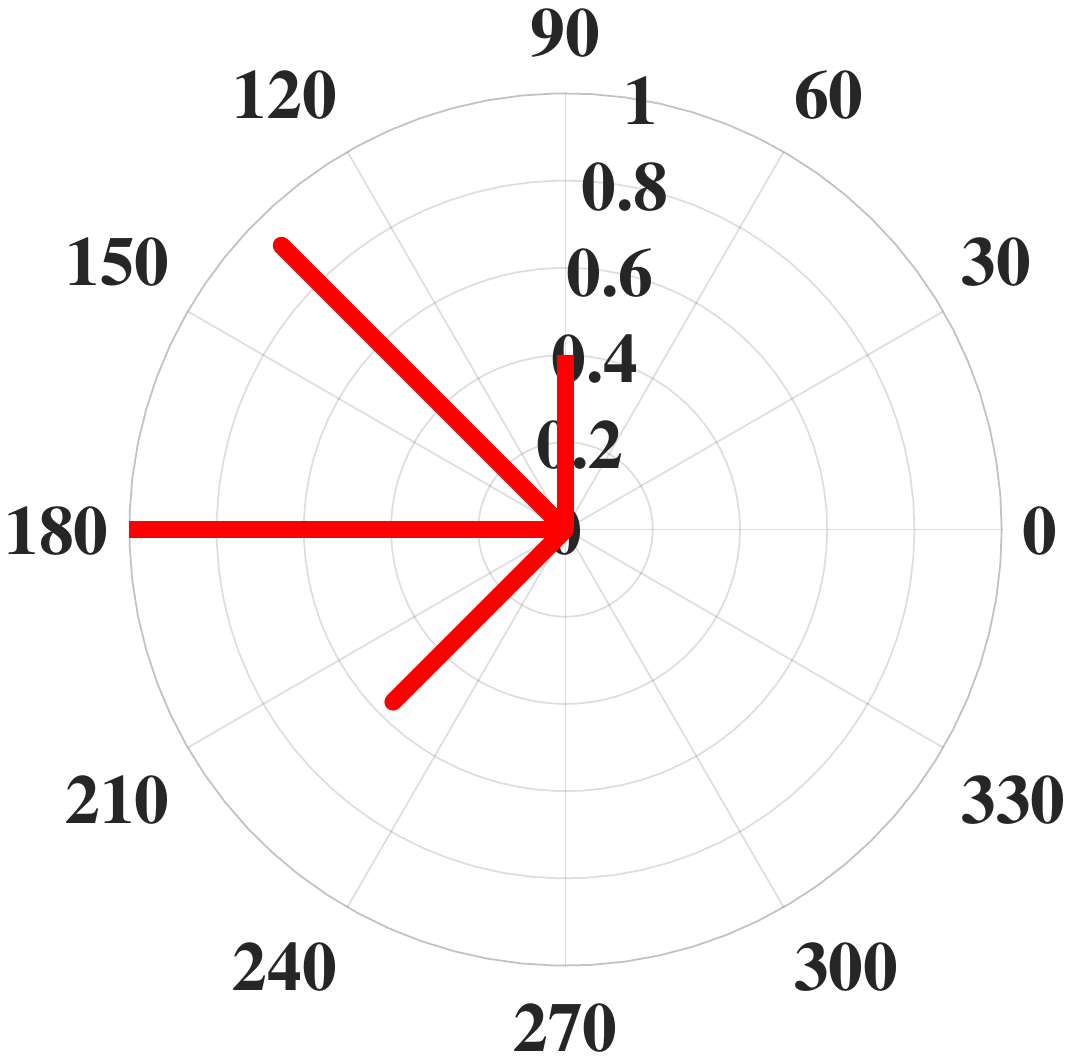}
		\label{Direction-Selectivity-Neural-Responses-Polar-X-111-Y-350}}
	\hfil
	\subfloat[]{\includegraphics[width=0.125\textwidth]{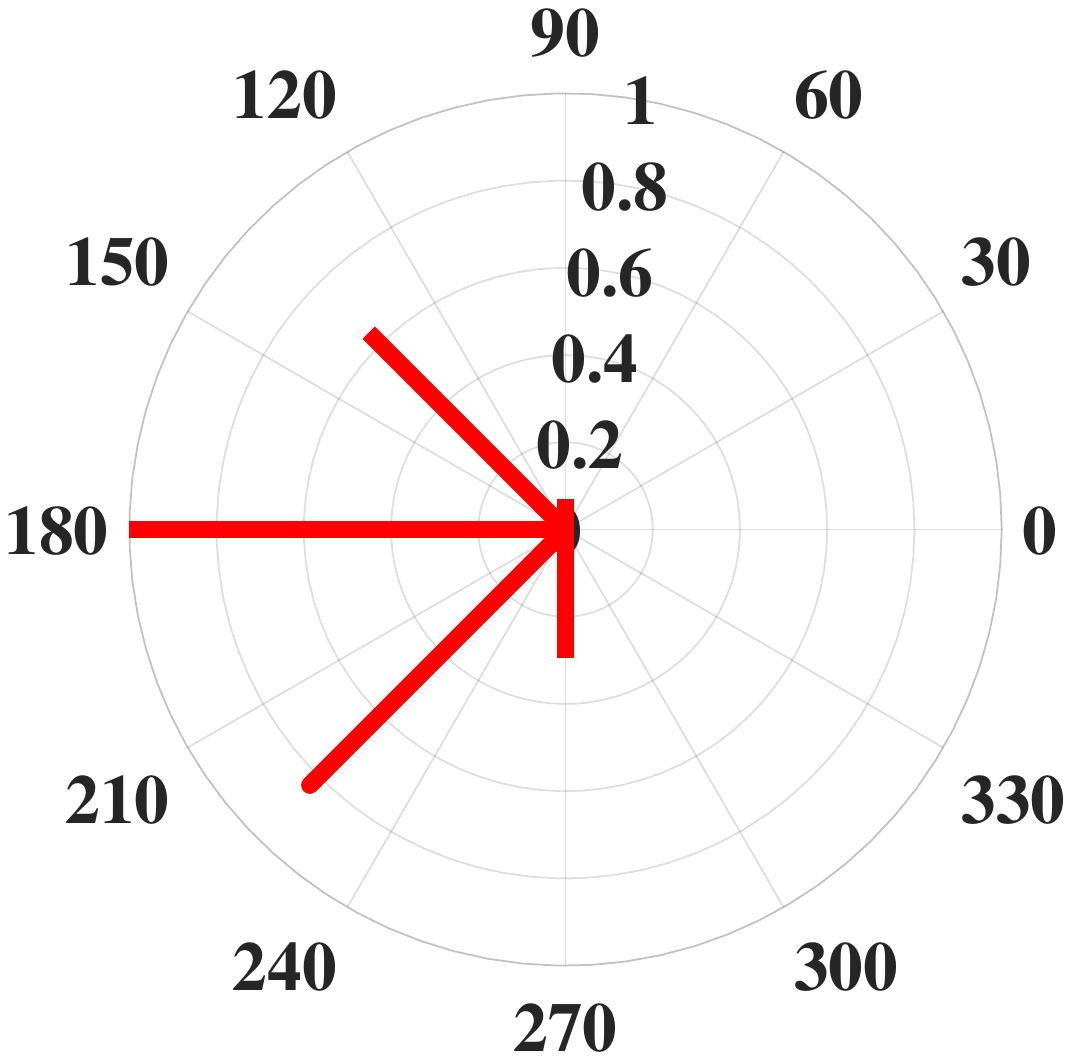}
		\label{Direction-Selectivity-Neural-Responses-Polar-X-111-Y-343}}
	\hfil
	\subfloat[]{\includegraphics[width=0.125\textwidth]{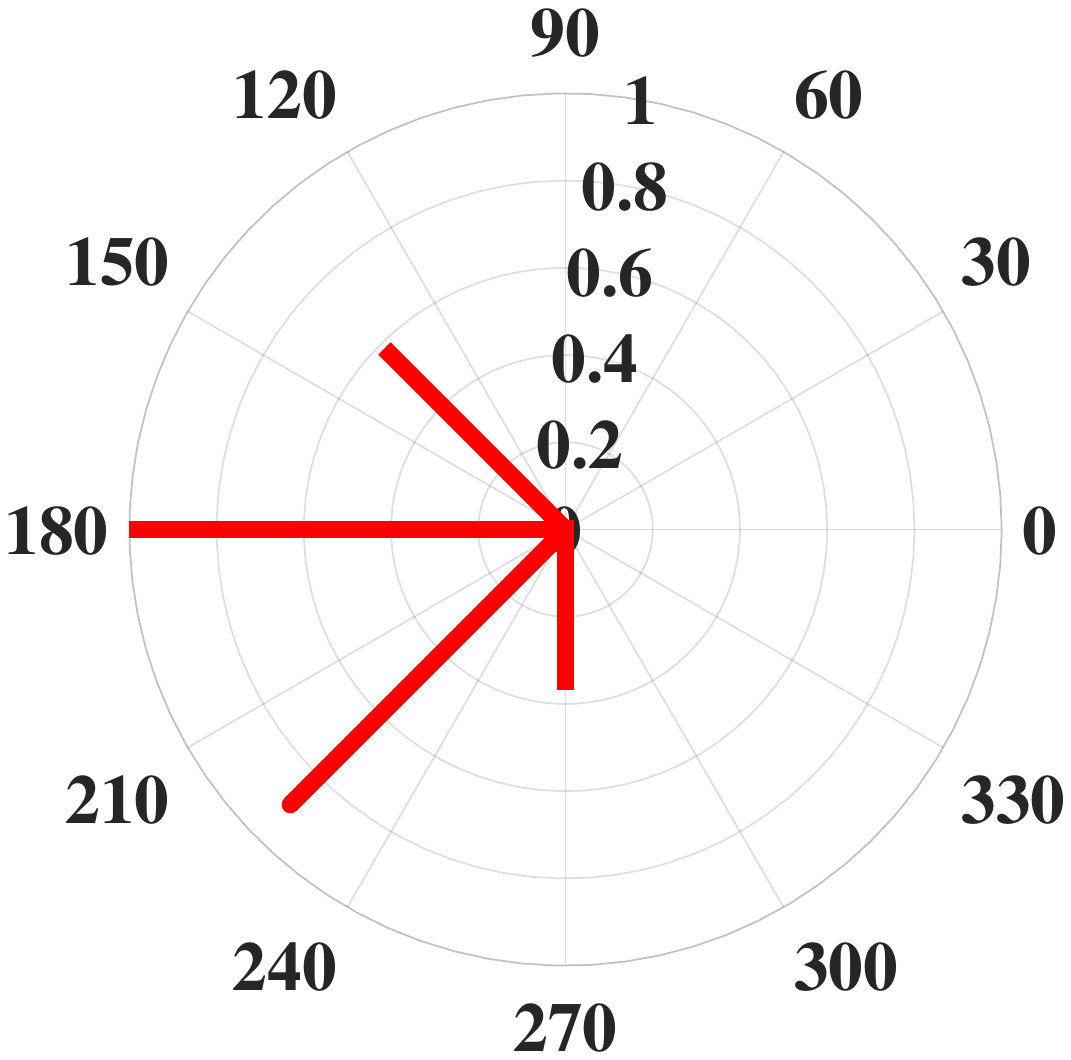}
		\label{Direction-Selectivity-Neural-Responses-Polar-X-112-Y-335}}
	\hfil
	\subfloat[]{\includegraphics[width=0.125\textwidth]{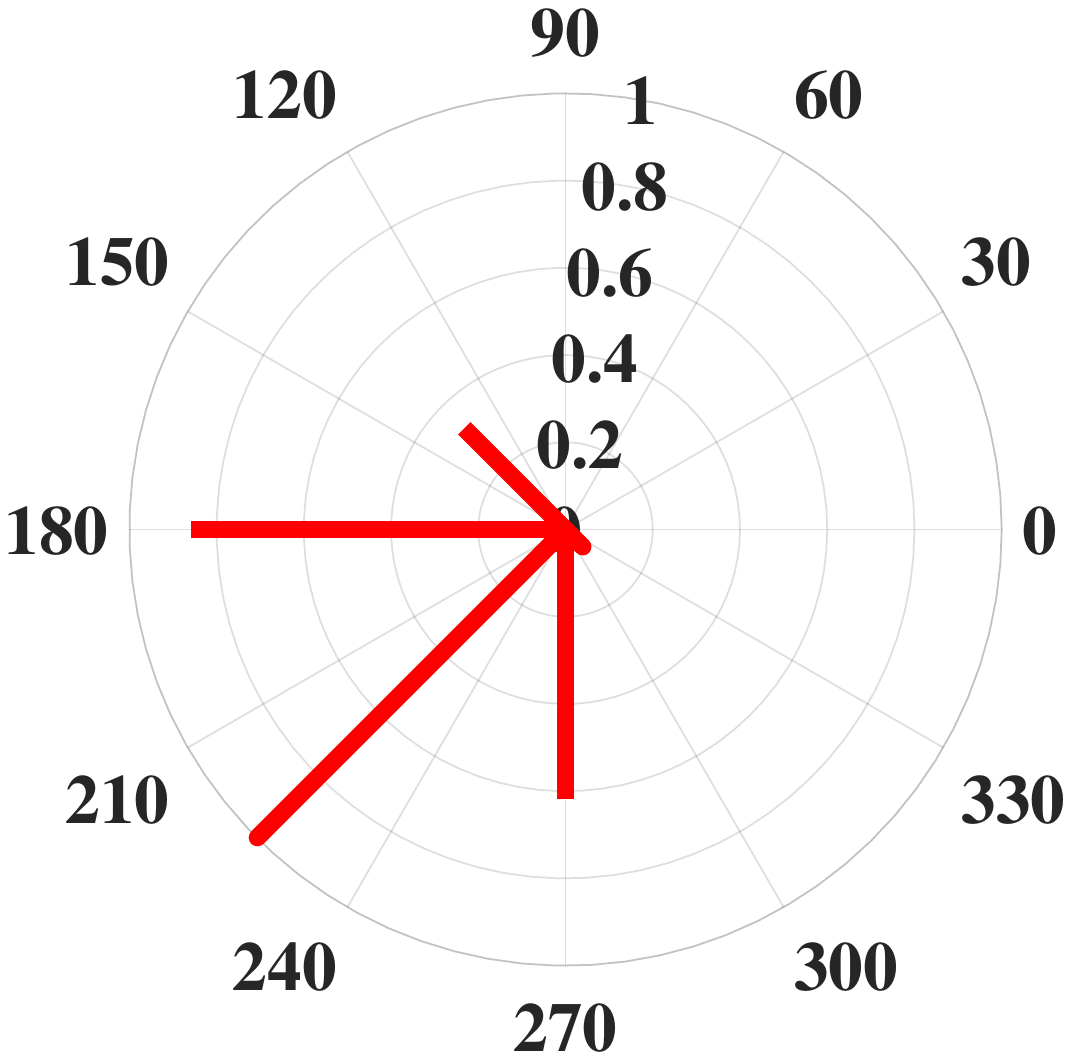}
		\label{Direction-Selectivity-Neural-Responses-Polar-X-121-Y-319}}
	\caption{(a)-(f) Normalized DSTMD outputs at the position A--F. In each subplot, the angular coordinate represents the preferred motion direction of the DSTMD while the radial coordinate denotes the strength of the DSTMD output tuned to this preferred direction.}
	\label{Direction-Tuning-Normalized-DS-STMD-Outputs-Polar}
\end{figure*}

\begin{figure*}[!t]
	\vspace{-10pt}
	\centering
	\subfloat[]{\includegraphics[width=0.125\textwidth]{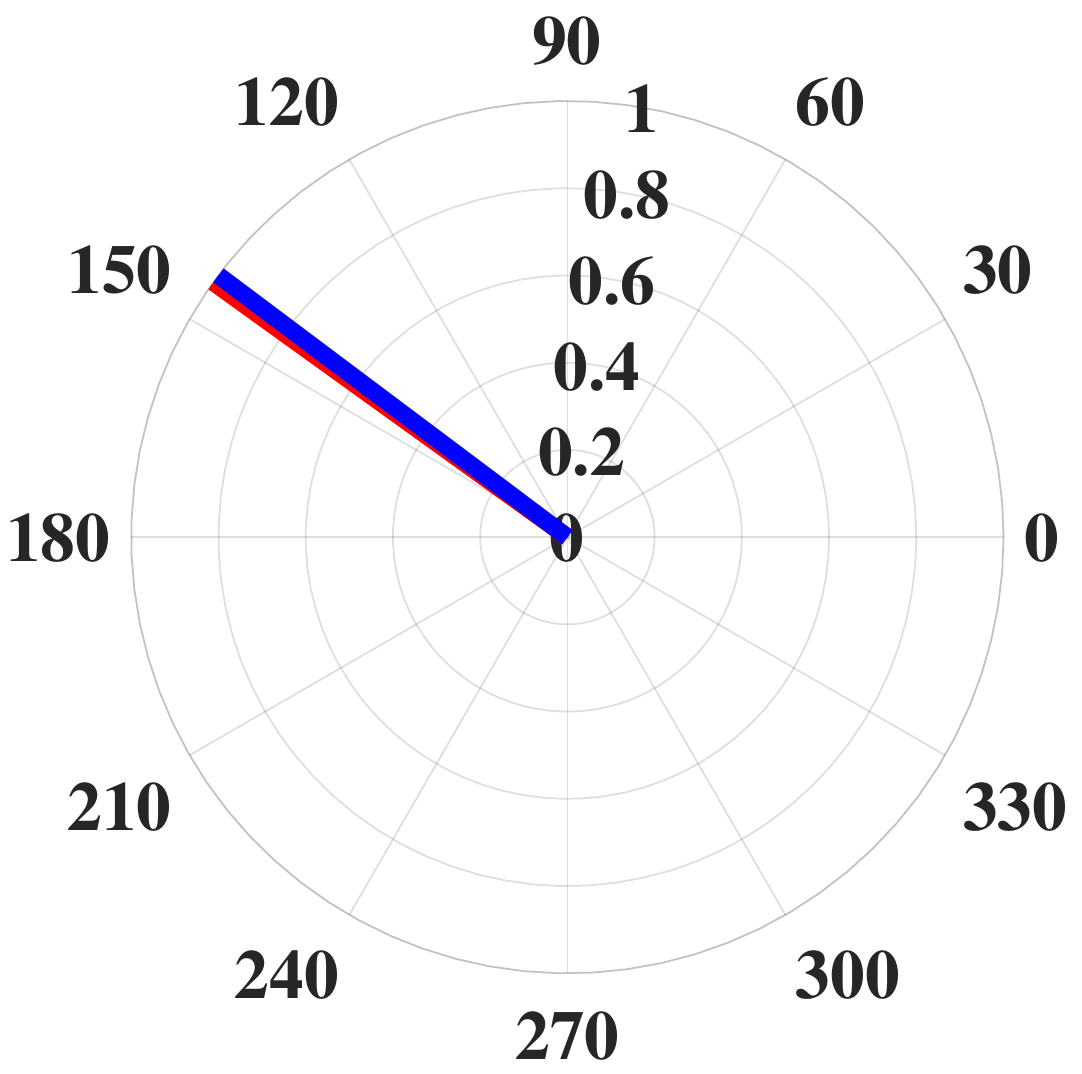}		\label{Direction-Selectivity-Angle-X-123-Y-373}}
	\hfil
	\subfloat[]{\includegraphics[width=0.125\textwidth]{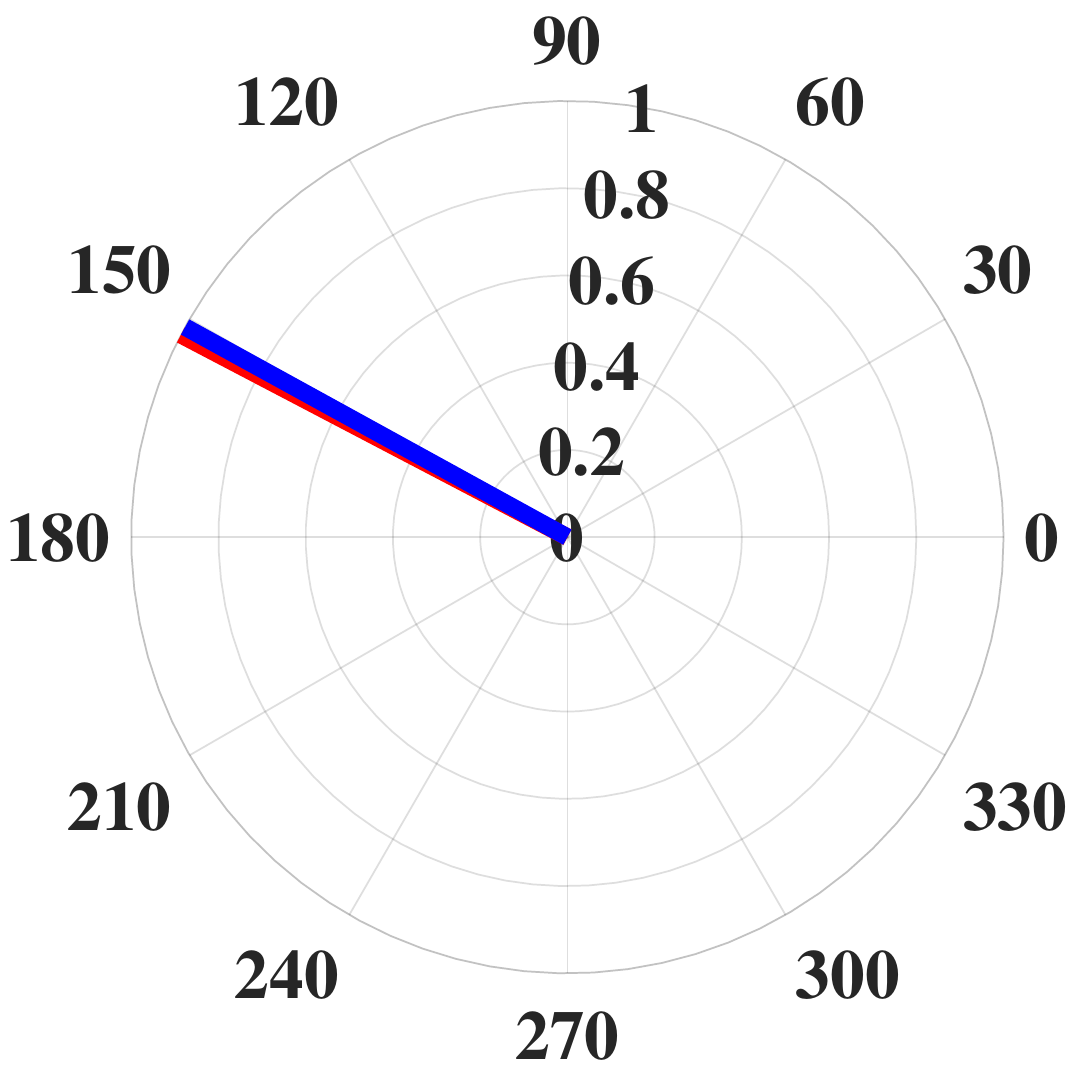}
		\label{Direction-Selectivity-Angle-X-115-Y-360}}
	\hfil
	\subfloat[]{\includegraphics[width=0.125\textwidth]{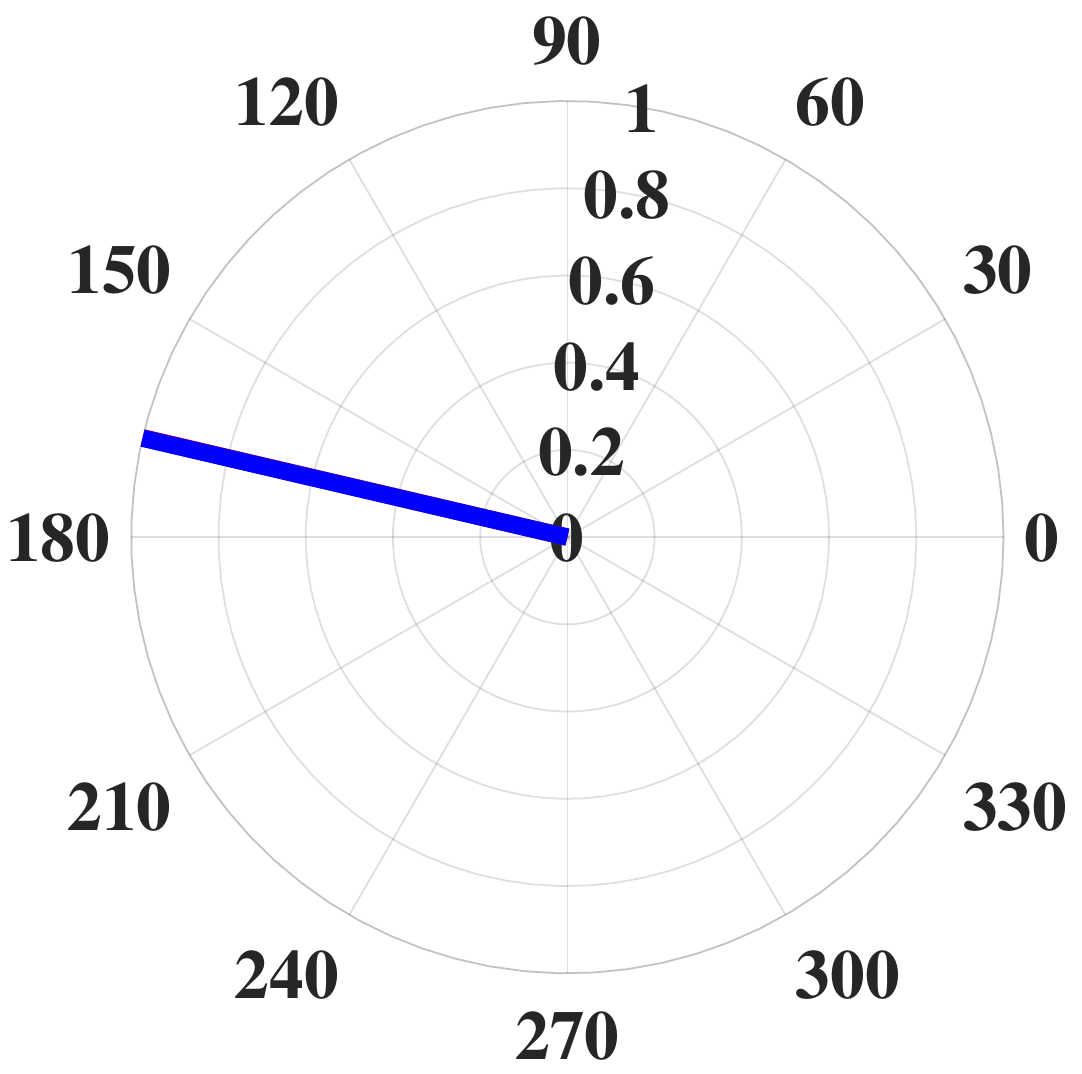}
		\label{Direction-Selectivity-Angle-X-111-Y-350}}
	\hfil
	\subfloat[]{\includegraphics[width=0.125\textwidth]{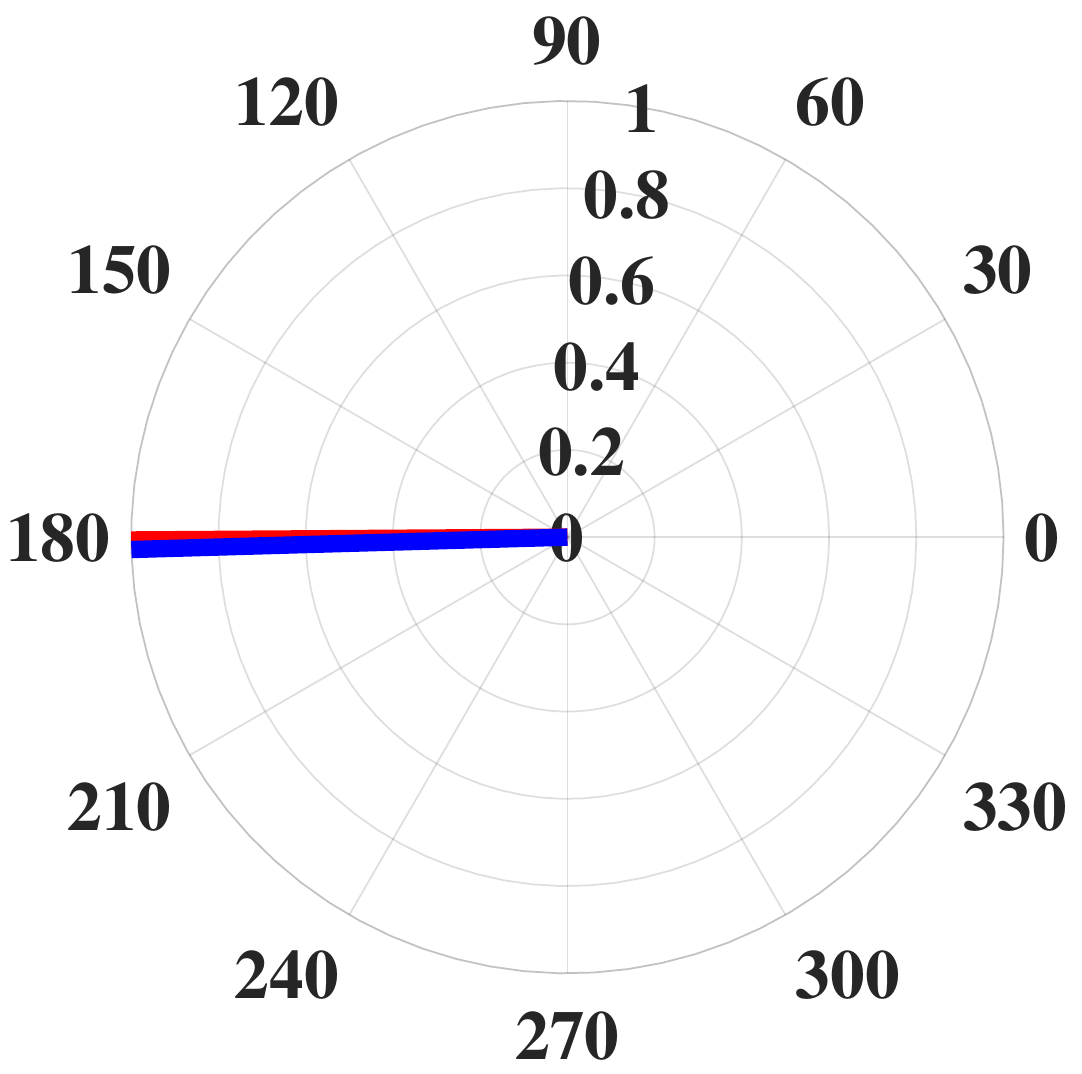}
		\label{Direction-Selectivity-Angle-X-111-Y-343}}
	\hfil
	\subfloat[]{\includegraphics[width=0.125\textwidth]{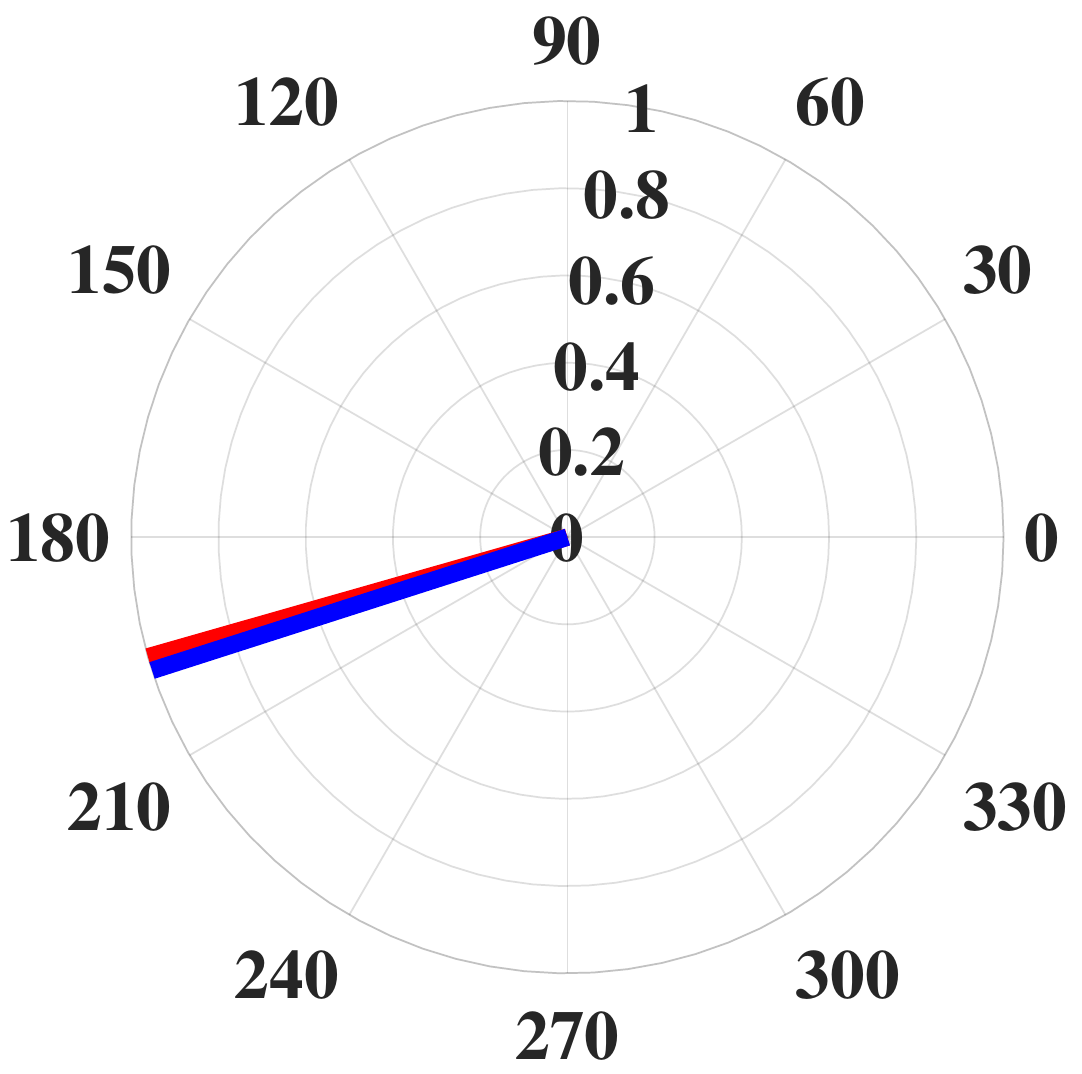}
		\label{Direction-Selectivity-Angle-X-112-Y-335}}
	\hfil
	\subfloat[]{\includegraphics[width=0.125\textwidth]{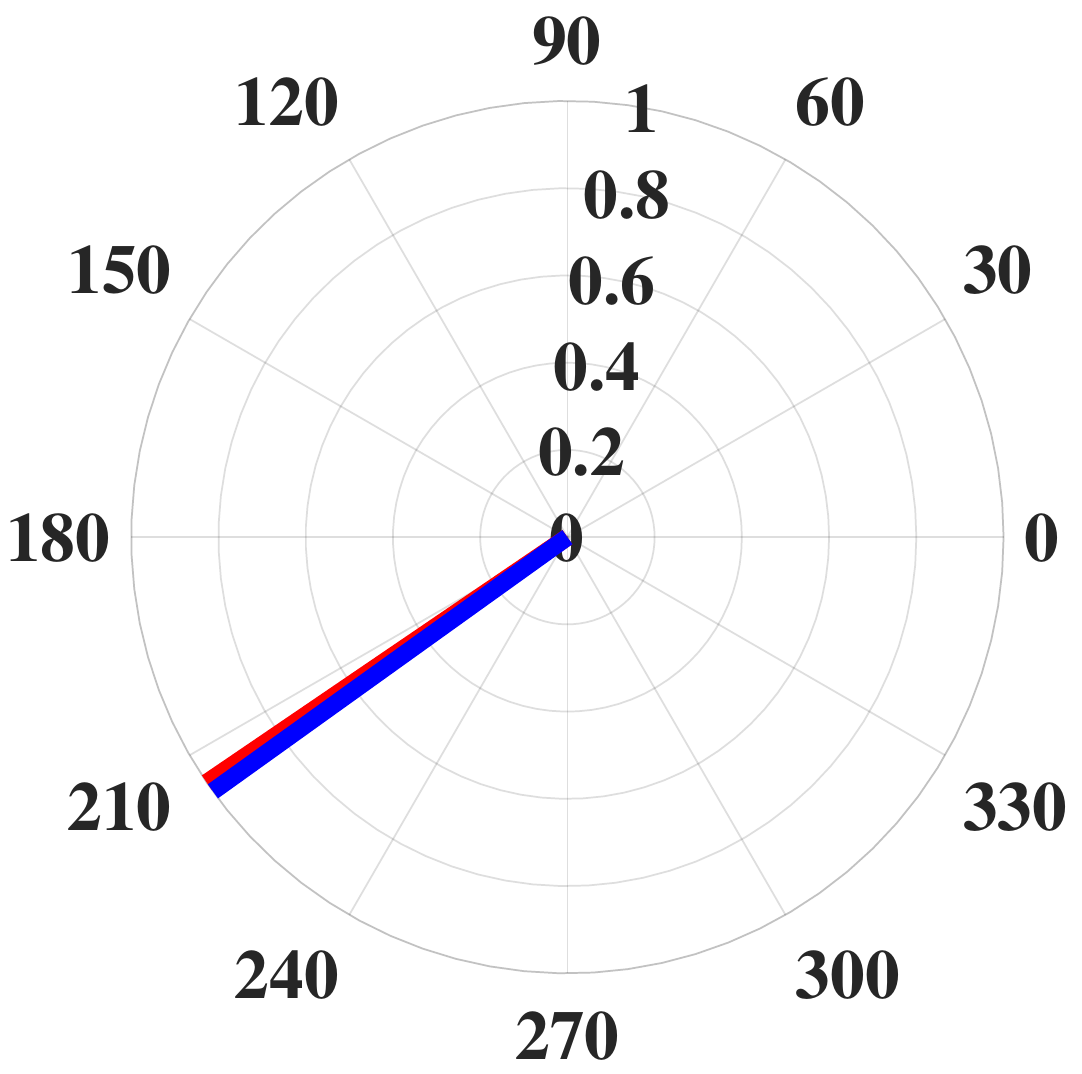}
		\label{Direction-Selectivity-Angle-X-121-Y-319}}
	\caption{(a)-(f) Estimated motion direction (red) and actual motion direction (blue) at the position A--F. In each subplot, the red line is highly overlapped with the blue line. That is, the estimated motion direction is quite close to the actual motion direction.}
	\label{Direction-Tuning-Normalized-Estimation-Actual-Polar}
\end{figure*}

We select six positions on the motion trace (A--F, in Fig. \ref{Direction-Selectivity-Target-Motion-Trace}). The outputs of the DSTMD at these six positions are normalized, then shown in polar coordinate (see Fig. \ref{Direction-Tuning-Normalized-DS-STMD-Outputs-Polar}). In each subplot of Fig. \ref{Direction-Tuning-Normalized-DS-STMD-Outputs-Polar}, we can see that the smaller difference between the preferred direction $\theta$ and the actual motion direction (shown in Fig. \ref{Direction-Tuning-Normalized-Estimation-Actual-Polar}), the stronger DSTMD output tuned to this direction $\theta$. These directionally selective outputs are used to encode the motion direction of the small target by the population vector algorithm. Fig. \ref{Direction-Tuning-Normalized-Estimation-Actual-Polar} and Table \ref{Estimated-Motion-Direction-Actual-Direction} show the estimated motion direction and the actual motion direction at the six positions. As can be seen, the difference between the estimated direction and actual direction is smaller than $2^{\circ}$ at these six positions. We further estimate the motion direction of the small target at each position of the motion trace. The maximal difference between the estimated motion direction and actual motion direction is $3.17^{\circ}$. Above results indicate that the proposed neural network provides a good estimation for the motion direction of the small target.

\begin{table}[t]
	\vspace{-5pt}
	\renewcommand{\arraystretch}{1.3}
	\caption{Estimated Motion Direction and Actual \hspace{100pt} Motion Direction at the Six Positions}
	\label{Estimated-Motion-Direction-Actual-Direction}
	\centering
	\begin{tabular}{|c|c|c|c|}
		\hline
		Position & Estimated & Actual  & Difference  \\	
		\hline
		A  & $144.25^{\circ}$ & $143.12^{\circ}$ & $1.13^{\circ}$ \\
		\hline
		B  & $152.36^{\circ}$ & $151.21^{\circ}$ & $1.15^{\circ}$ \\
		\hline
		C & $166.83^{\circ}$ & $166.88^{\circ}$ & $0.05^{\circ}$ \\
		\hline
		D & $180.37^{\circ}$ & $181.63^{\circ}$ & $1.26^{\circ}$ \\
		\hline
		E & $195.93^{\circ}$ & $197.80^{\circ}$ & $1.87^{\circ}$ \\
		\hline
		F & $214.24^{\circ}$ & $215.53^{\circ}$ & $1.29^{\circ}$ \\
		\hline
	\end{tabular}
\end{table}

\subsection{Target Detection in Cluttered Backgrounds}
\label{Target-Detection-in-Cluttered-Backgrounds}
In this section, we test the ability of the proposed neural network for detecting small targets against cluttered backgrounds. 
For a given detection threshold $\gamma$, if there is a position $(x_0,y_0)$, time $t_0$ and direction $\theta_0$ which satisfy the DSTMD output $E(x_0,y_0,t_0;\theta_0)>\gamma$, then we believe that a small target is detected at position $(x_0,y_0)$ and time $t_0$.
Two metrics are defined to evaluate the detection performance. That is,
\begin{align}
D_R & = \frac{\text{number of true detections}}{\text{number of actual targets}} \\
F_A & = \frac{\text{number of false detections}}{\text{number of images}}
\end{align}
where $D_R$ and $F_A$ denote the detection rate and false alarm rate, respectively. The detected result is considered correct if the pixel distance between the ground truth and the result is within a threshold ($5$ pixels).

\begin{figure}[!t]
	\vspace{-5pt}
	\centering
	\includegraphics[width=0.25\textwidth]{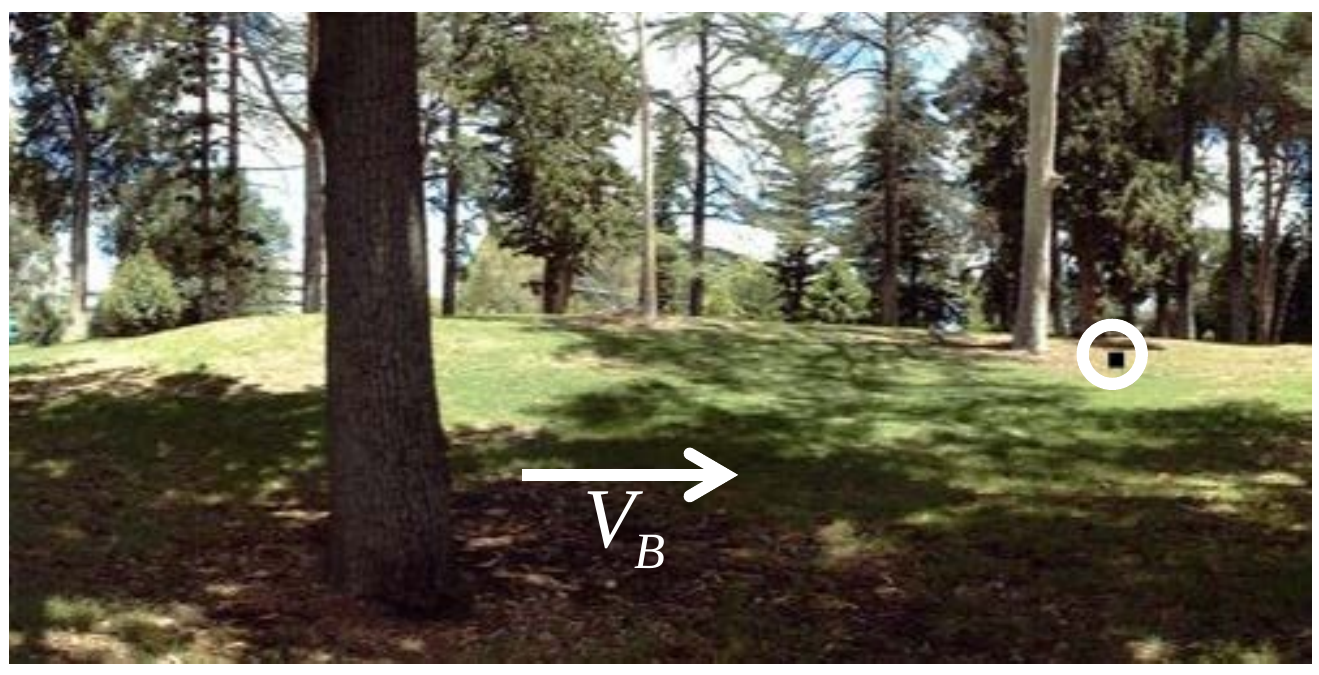}
	\caption{Representative frame of the input image sequence. The small target is highlighted by the white circle. The white arrow $V_B$ denotes the motion direction of the background.}
	\label{Curvilinear-Motion-Original-Image}
\end{figure}

\begin{figure*}[!t]
	\vspace{-10pt}
	\centering
	\subfloat[]{\includegraphics[width=0.25\textwidth]{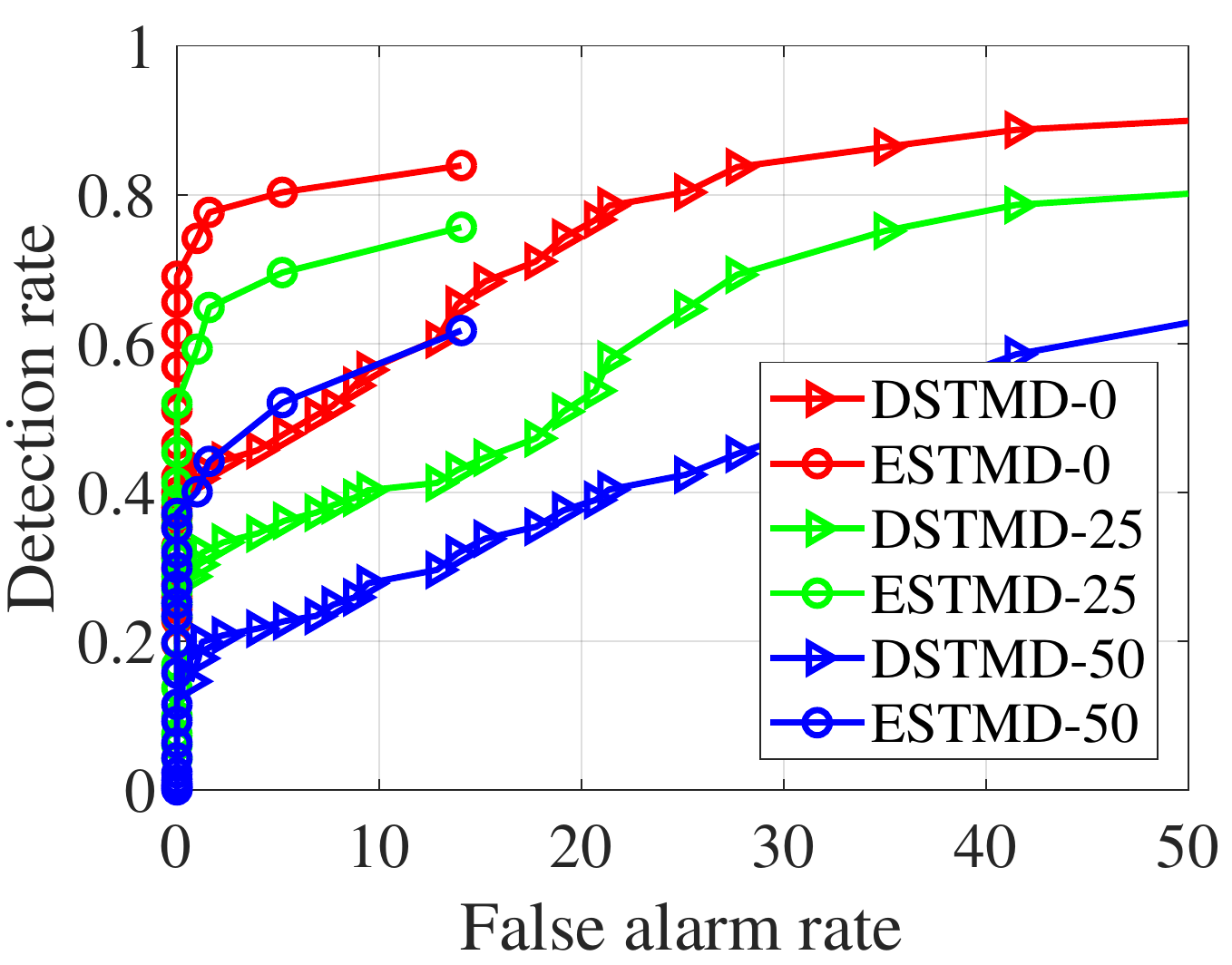}
		\label{CB-1-LDTB-DR-FA}}
	\hfil
	\subfloat[]{\includegraphics[width=0.25\textwidth]{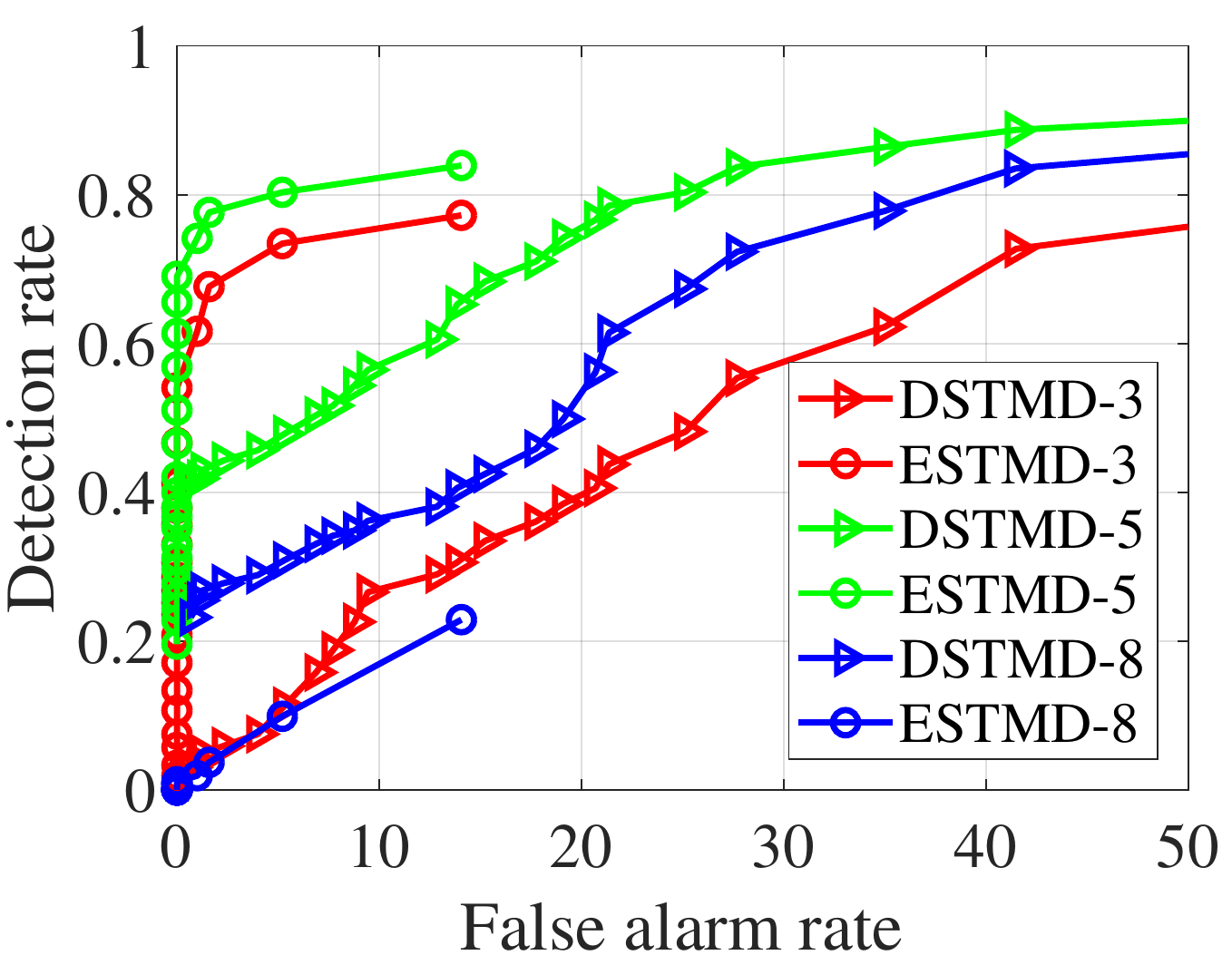}
		\label{CB-1-Size-DR-FA}}
	\hfil
	\subfloat[]{\includegraphics[width=0.25\textwidth]{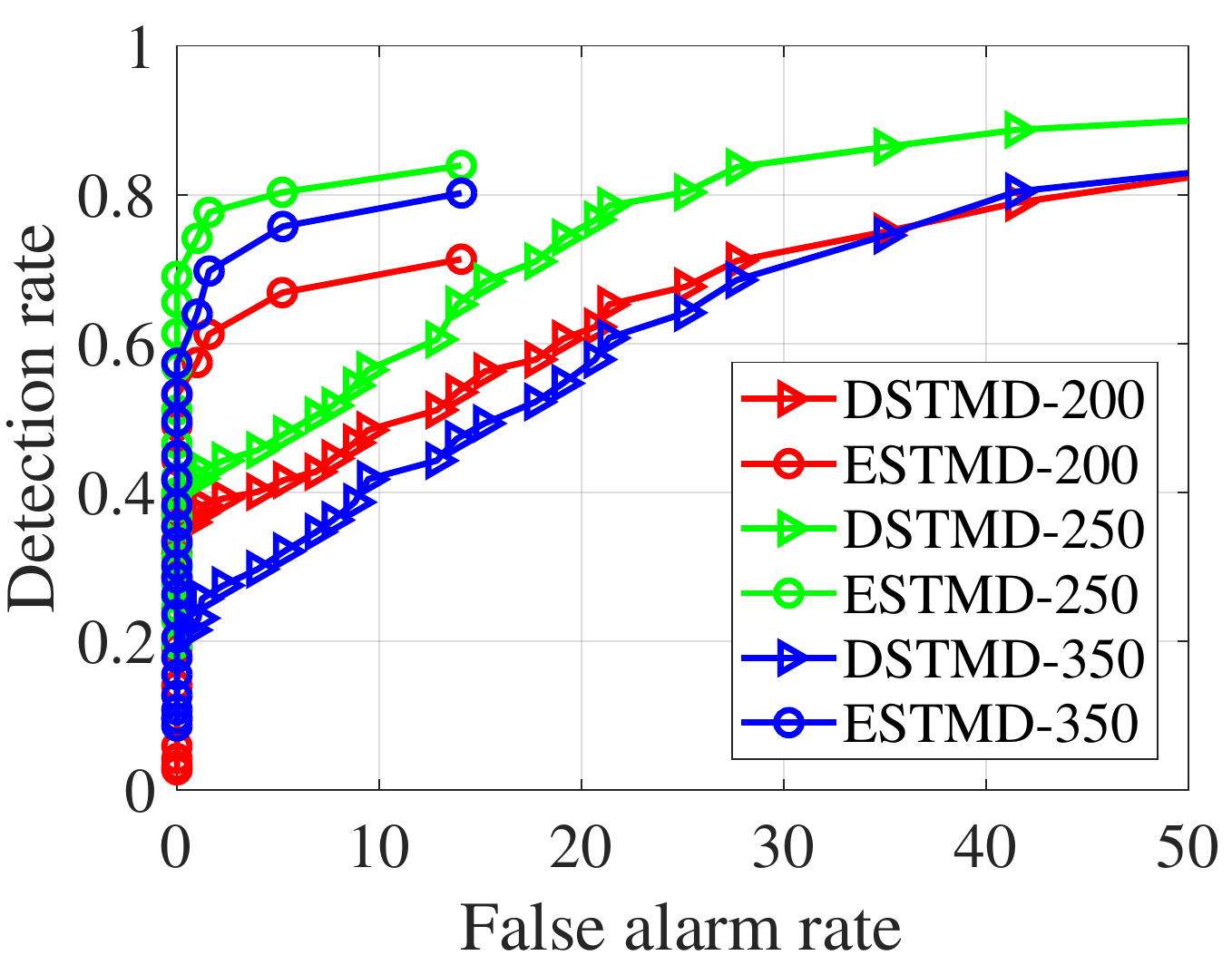}
		\label{CB-1-Velocity-DR-FA}}
	
	\caption{ROC curves of the first three experiments with respect to target luminance, sizes and velocities. (a) Experiment $1$, different target luminance. Legend 'ESTMD-$0$' and 'DSTMD-$0$' represent the ROC curves of the ESTMD and DSTMD when target luminance equals to $0$, respectively. Similar explanations for other legends. (b) Experiment $2$, different target sizes. Legend 'ESTMD-$3$' and 'DSTMD-$3$' represent the ROC curves of the ESTMD and DSTMD when target size equals to $3 \times 3$ pixels, respectively. Similar explanations for other legends. (c) Experiment $3$, different horizontal velocities ($V_{_T}^x$). Legend 'ESTMD-$200$' and 'DSTMD-$200$' represent the ROC curves of the ESTMD and DSTMD when the horizontal velocity $V_{_T}^x$ equals to $200$ pixel/s, respectively. Similar explanations for other legends.}
	\label{CB-1-LDTB-Size-Velocity-DR-FA}
\end{figure*}

In the first three experiments, we investigate the influences of three target parameters (size, luminance and velocity) on the detection performance. In each experiment, we change one of the target parameters while fix the other two parameters, then record the detection performance of the models under this parameter setting. The parameter settings of the first three experiments are shown in Table \ref{Parameter-Selection-For-Target-CB-1}. All input image sequences are produced using the same background image where a representative frame is given in Fig. \ref{Curvilinear-Motion-Original-Image}. In all input image sequences, the background is moving from left to right and its velocity is set as $250$ pixel/s. A small target is moving against the cluttered background, and its coordinate at time $t$ is $(500 - V_{_T}^x \cdot \frac{t+300}{1000}, 125+15 \cdot \sin(4\pi \frac{t+300}{1000})), t \in [0, 1000]$ ms where $V_{_T}^x$ denotes the horizontal velocity. The receiver operating characteristic (ROC) curves of the three experiments with respect to target luminance, size and horizontal velocity $V_{_T}^x$, are displayed in Fig. \ref{CB-1-LDTB-Size-Velocity-DR-FA}.

\begin{table}[!t]
	\vspace{-5pt}
	\renewcommand{\arraystretch}{1.3}
	\caption{Settings of the Parameters Including Target Luminance, Size and Horizontal Velocity for the First Three Experiments}
	\label{Parameter-Selection-For-Target-CB-1}
	\centering
	\begin{tabular}{|c|c|c|c|}
		\hline
		& Luminance & Size & Velocity ($V_{_T}^x$) \\	
		\hline
		Experiment $1$ & $0, 25, 50$ & $5 \times 5$ & $250$ \\
		\hline
		Experiment $2$ & $0$ & $3 \times 3, 5 \times 5, 8 \times 8$ & $250$ \\
		\hline
		Experiment $3$ & $0$ & $5 \times 5$ & $200, 250, 350$ \\
		\hline
	\end{tabular}
\end{table}

\begin{figure*}[!t]
	\vspace{-10pt}
	\centering
	\subfloat[]{\includegraphics[width=0.25\textwidth]{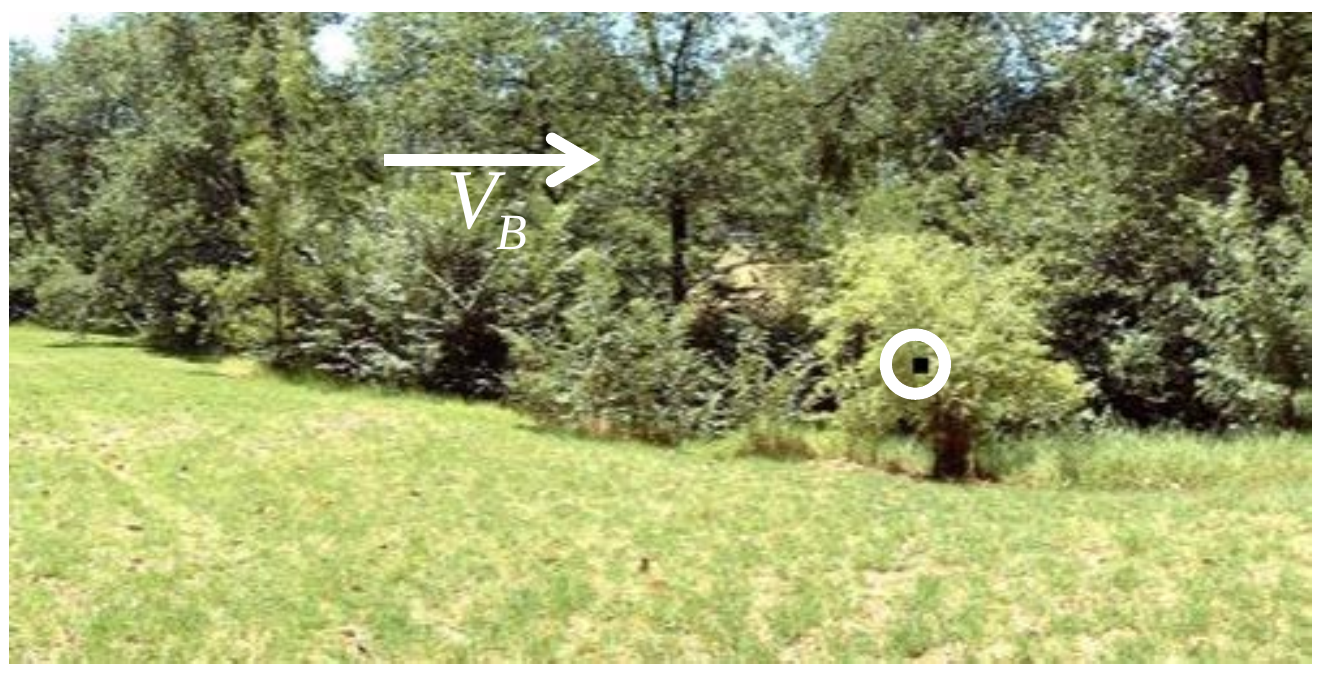}
		\label{CB-2-Frame-600}}
	\hfil
	\subfloat[]{\includegraphics[width=0.23\textwidth]{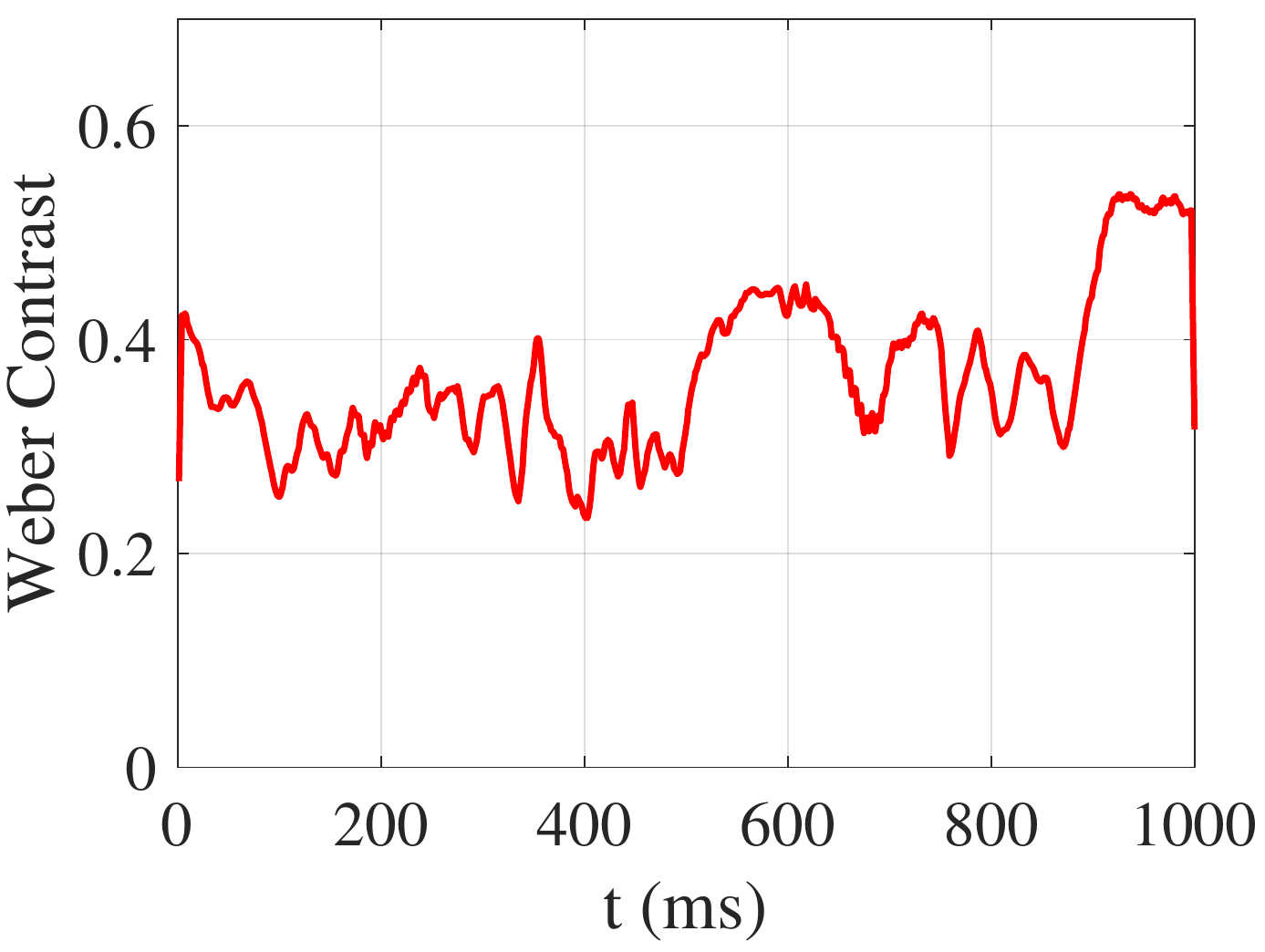}
		\label{CB-2-LDTB}}
	\hfil
	\subfloat[]{\includegraphics[width=0.23\textwidth]{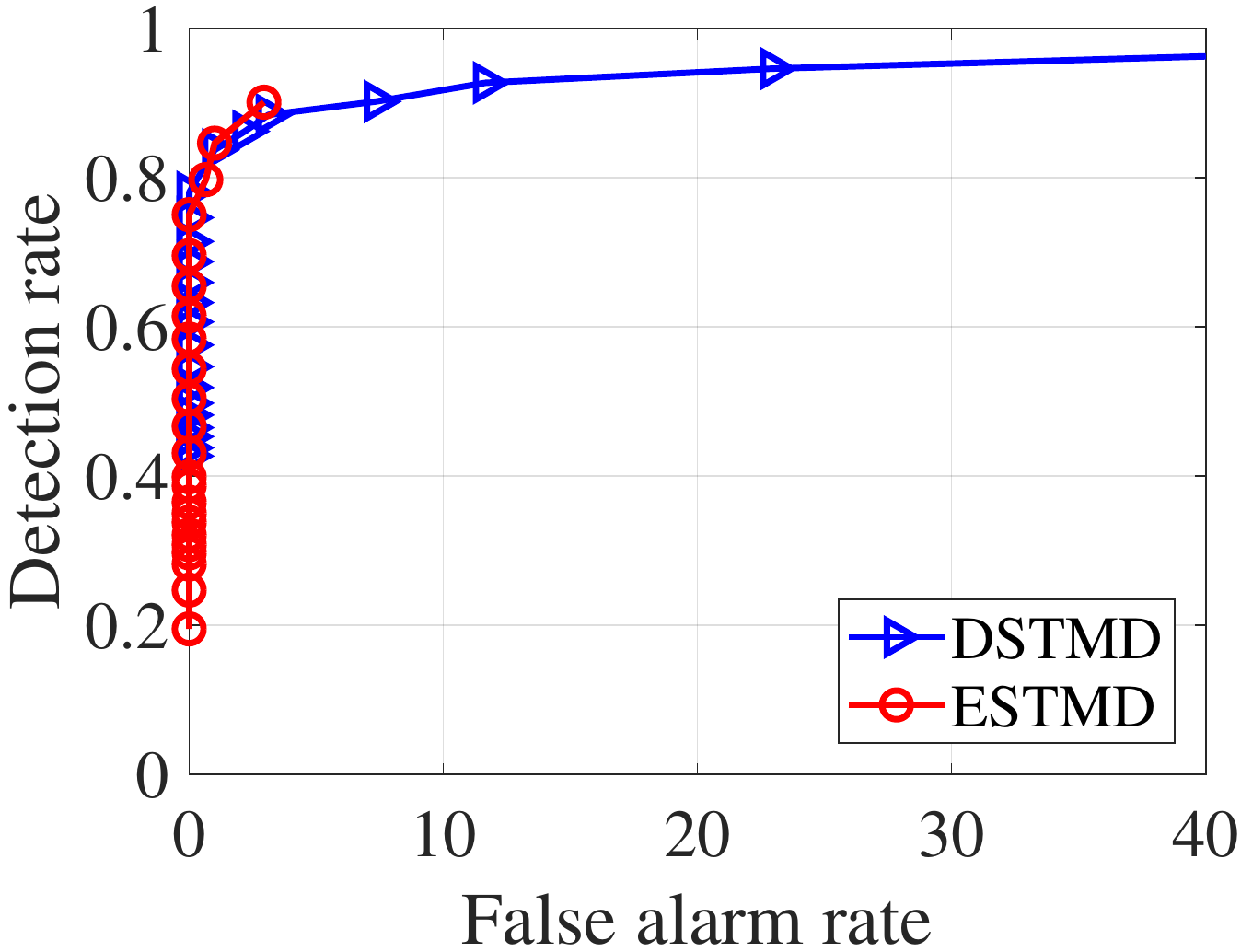}
		\label{CB-2-DR-FA}}
	\hspace{5cm}
	\subfloat[]{\includegraphics[width=0.125\textwidth]{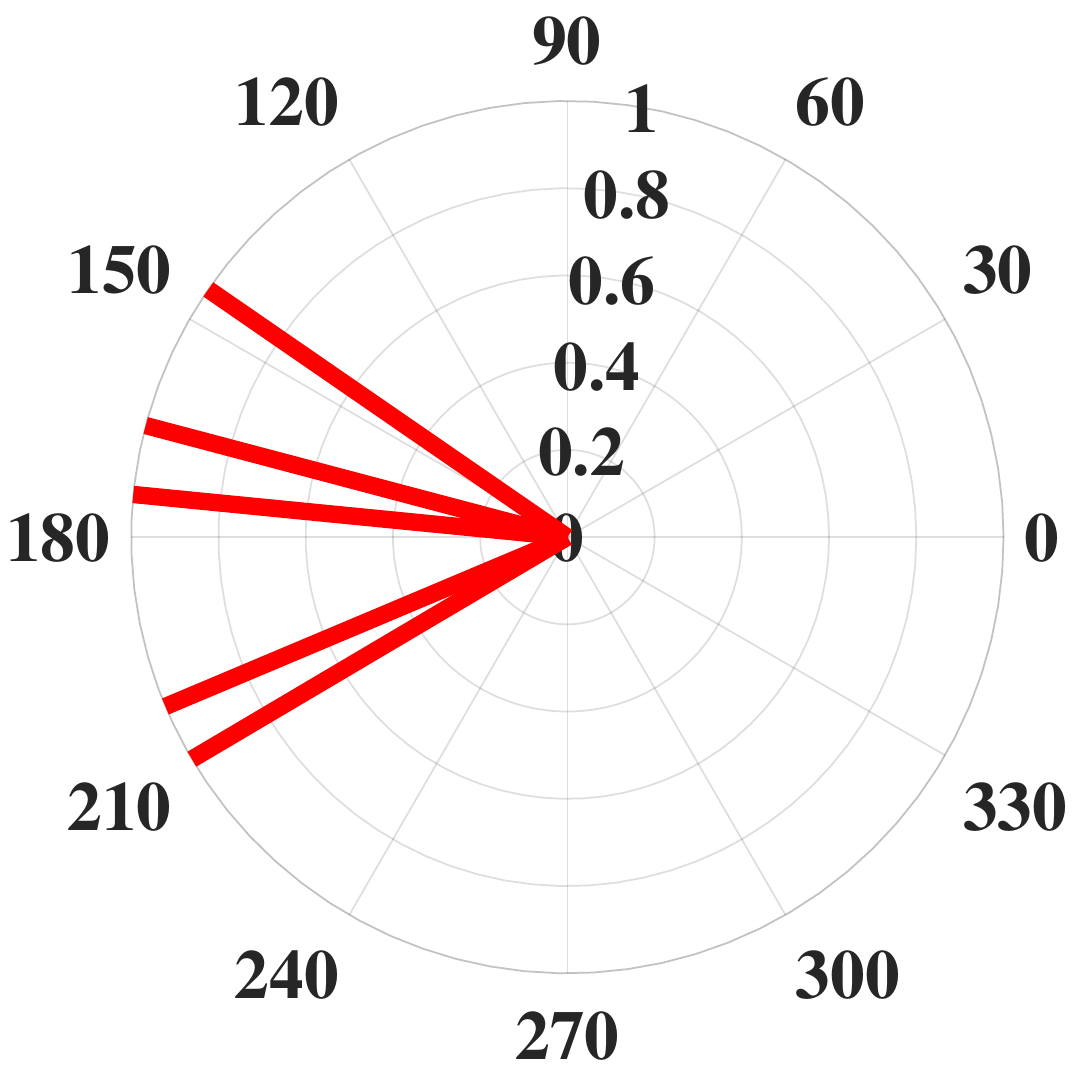}
		\label{CB-2-Estimated-Motion-Direction-510-570-600-630-700}}
	\hfil
	\subfloat[]{\includegraphics[width=0.125\textwidth]{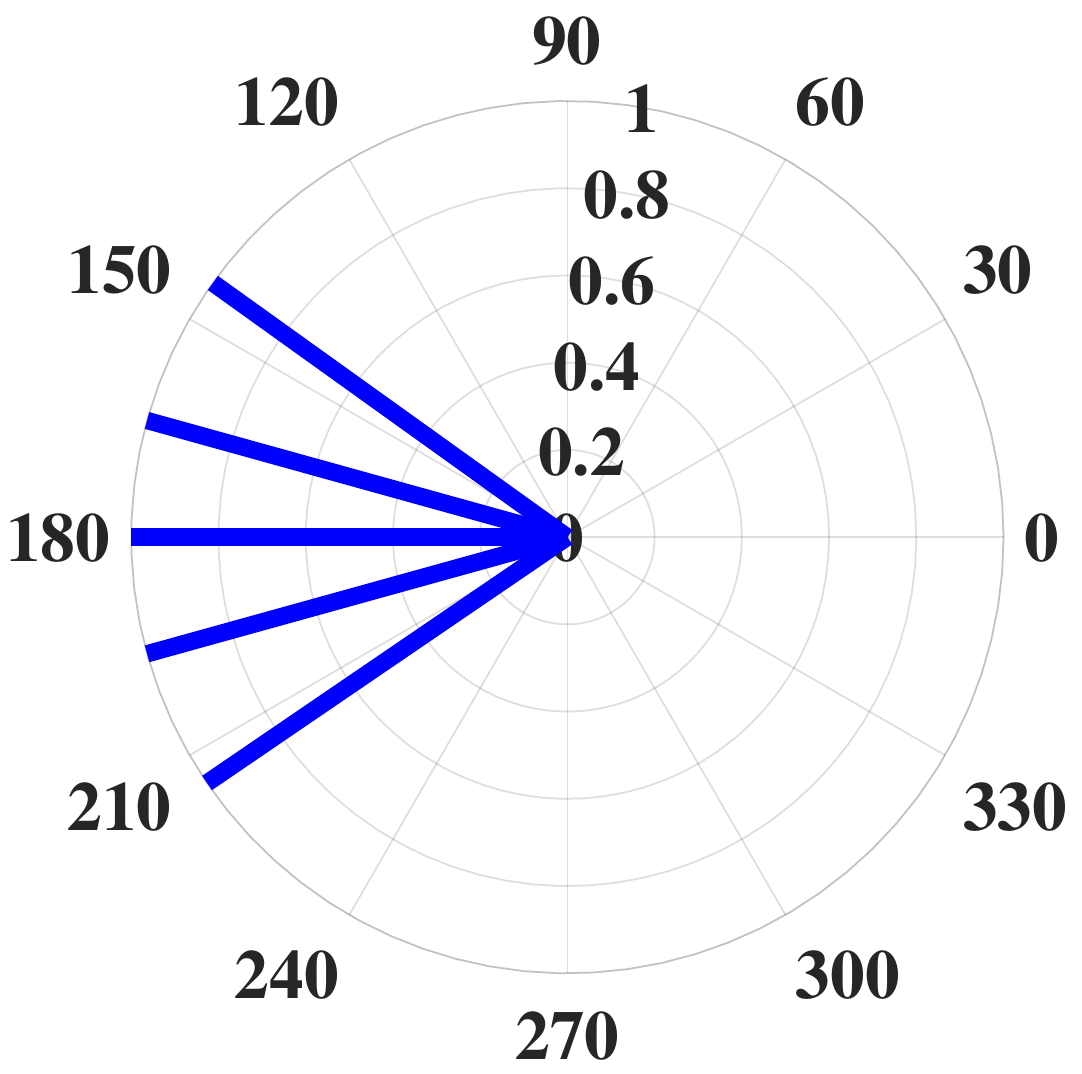}
		\label{CB-2-Actual-Motion-Direction-510-570-600-630-700}}
	\hfil
	\subfloat[]{\includegraphics[width=0.125\textwidth]{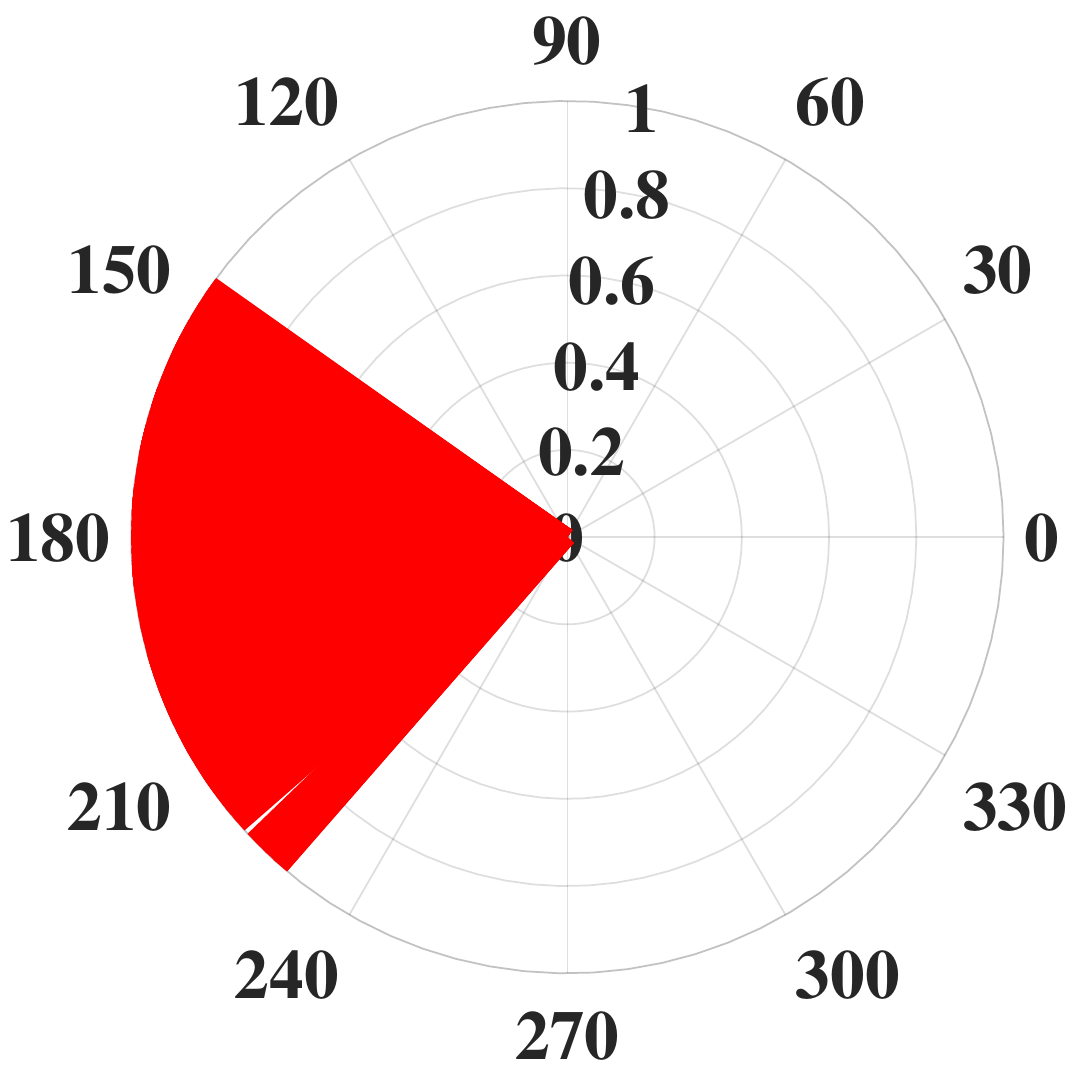}
		\label{CB-2-Estimated-Direction-500-700-1}}
	\hfil
	\subfloat[]{\includegraphics[width=0.125\textwidth]{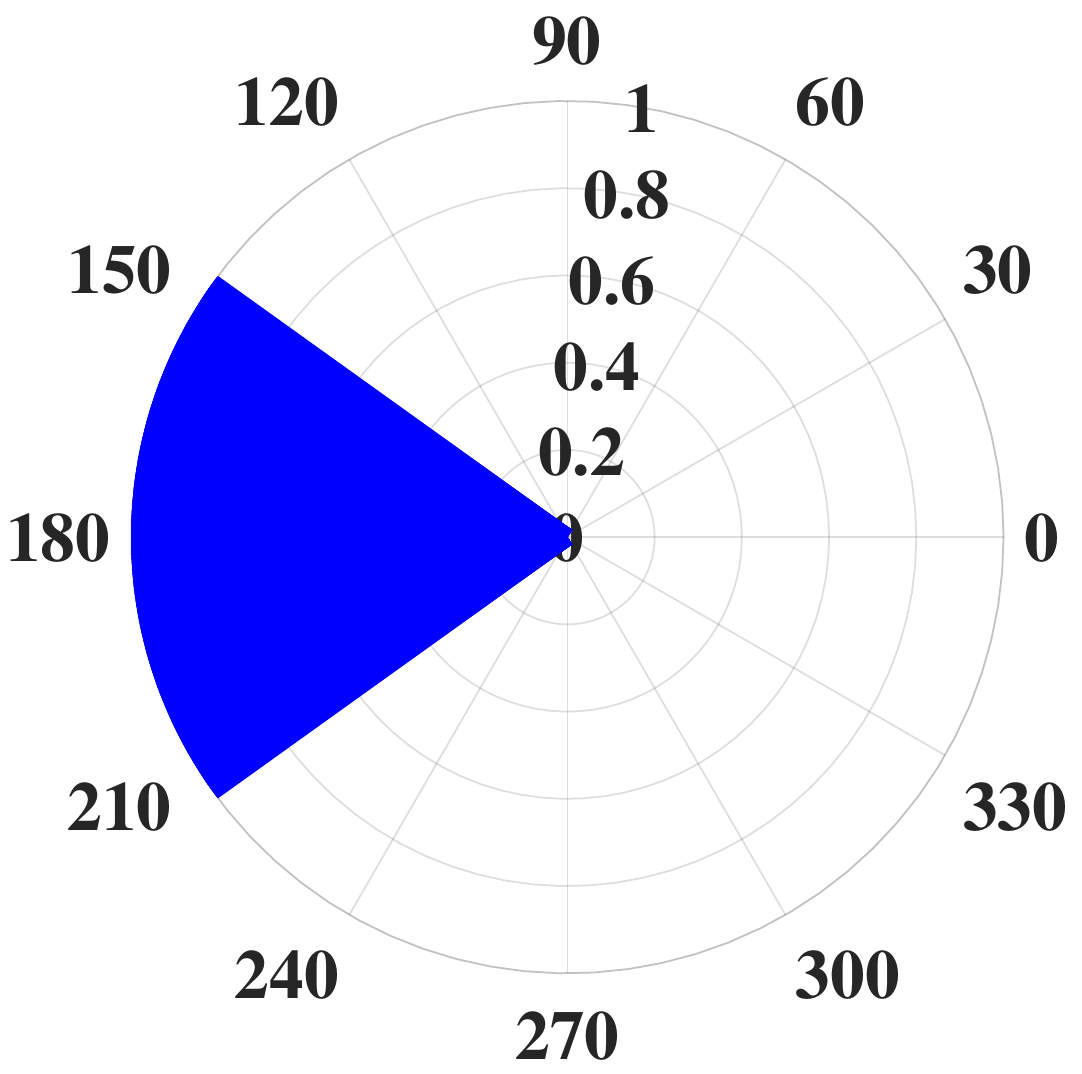}
		\label{CB-2-Actual-Direction-500-700-1}}
	
	\caption{Experiment $4$. (a) Representative frame of the input image sequence. (b) Weber Contrast of the small target during time period $t \in [0, 1000]$ ms. (c) ROC curves of the ESTMD and DSTMD. (d) Motion directions detected by the DSTMD in the sample $510$, $570$, $600$, $630$, $700$ frames. No motion direction detected by the ESTMD. (e) Actual motion directions in the sample $510$, $570$, $600$, $630$, $700$ frames. (f) Motion directions detected by the DSTMD from the $500$th to the $700$th frame. (g) Actual motion directions from the $500$th to the $700$th frame.}
	\label{CB-2-LDTB-DR-FA}
\end{figure*}

In Fig. \ref{CB-1-LDTB-Size-Velocity-DR-FA}(a), we can see that the lower target luminance is, the better ESTMD and DSTMD perform. This is because the decrease of target luminance can induce the increase of Weber Contrast (see Fig. S3 in the supplementary material). Note that the ESTMD and DSTMD all show Weber Contrast sensitivity, so the higher Weber Contrast can elicit the stronger model output. From Fig. \ref{CB-1-LDTB-Size-Velocity-DR-FA}(b) and \ref{CB-1-LDTB-Size-Velocity-DR-FA}(c), we can see that when the false alarm rate is given, the target size of  $5 \times 5$ (or the velocity of $250$) has higher detection rate compared to the target size of $3 \times 3$ and $8 \times 8$ (or the velocity of $200$ and $350$). This is because the ESTMD and DSTMD all exhibit size and velocity selectivities. They show the strongest output to the target whose size (or velocity) equals to $5 \times 5$ pixels (or $250$ pixel/s), but weaker outputs to the object whose size (or velocity) is higher or lower than $5 \times 5$ pixels (or $250$ pixel/s).

\begin{figure*}[!t]
	\vspace{-15pt}
	\centering
	\subfloat[]{\includegraphics[width=0.25\textwidth]{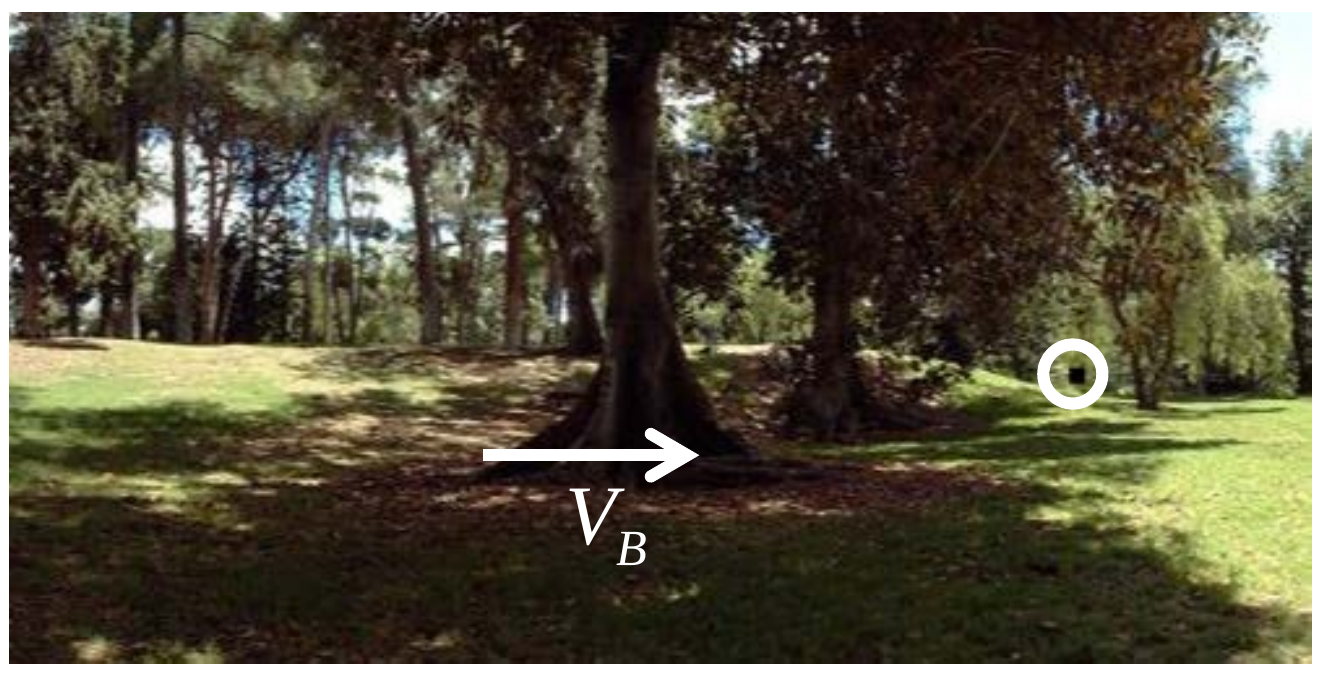}
		\label{CB-3-Frame-360}}
	\hfil
	\subfloat[]{\includegraphics[width=0.23\textwidth]{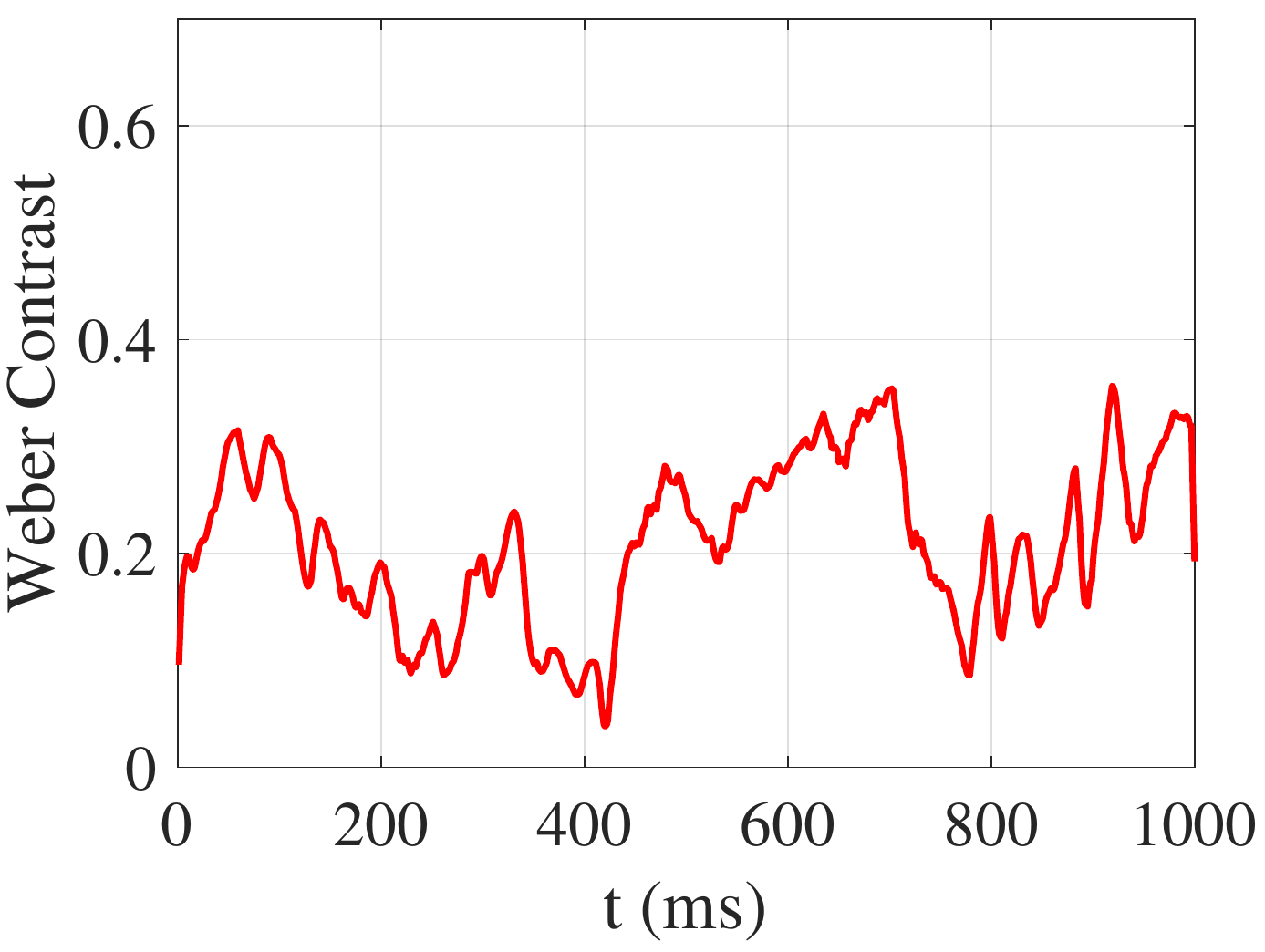}
		\label{CB-3-LDTB}}
	\hfil
	\subfloat[]{\includegraphics[width=0.23\textwidth]{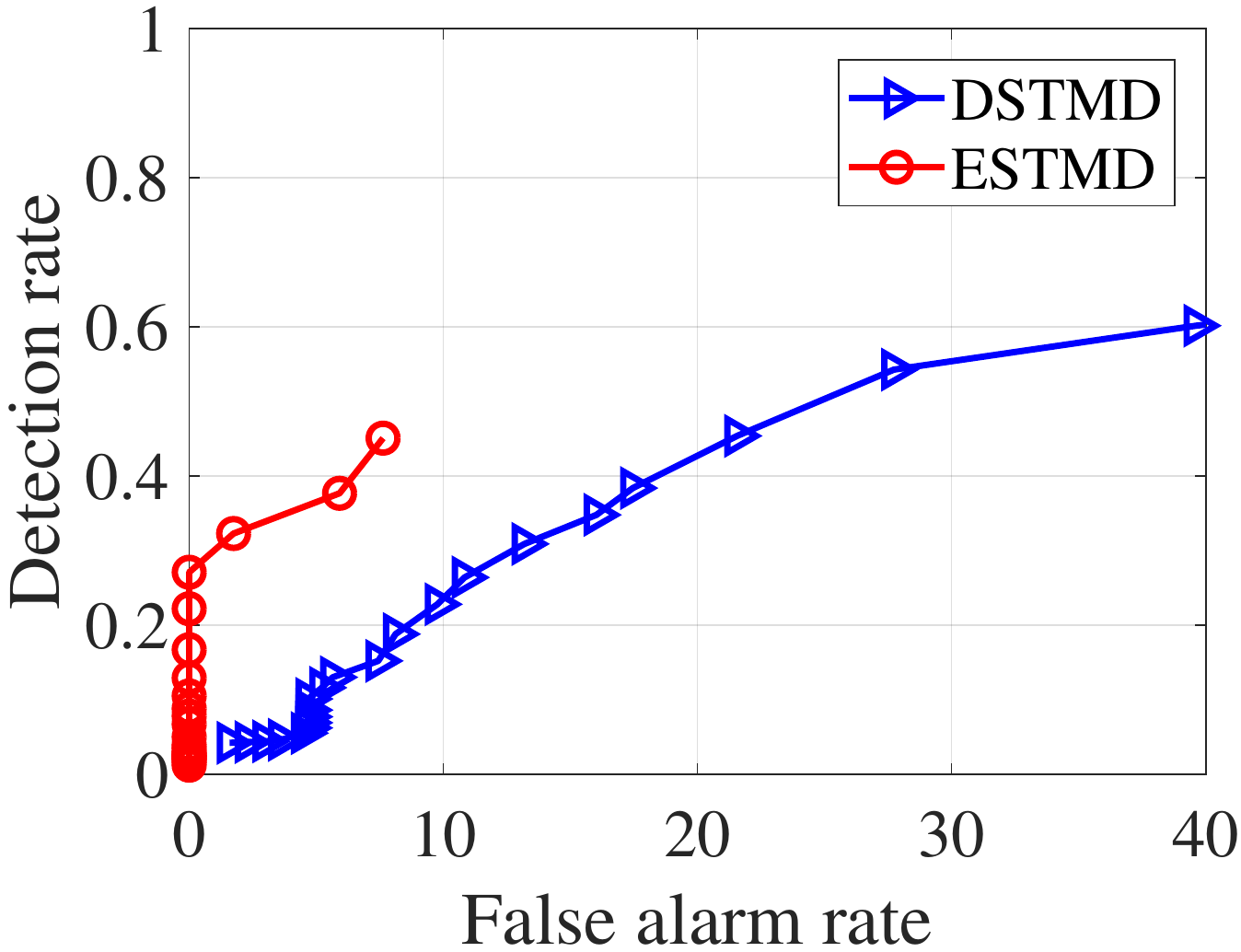}
		\label{CB-3-DR-FA}}
	\hspace{5cm}
	\subfloat[]{\includegraphics[width=0.125\textwidth]{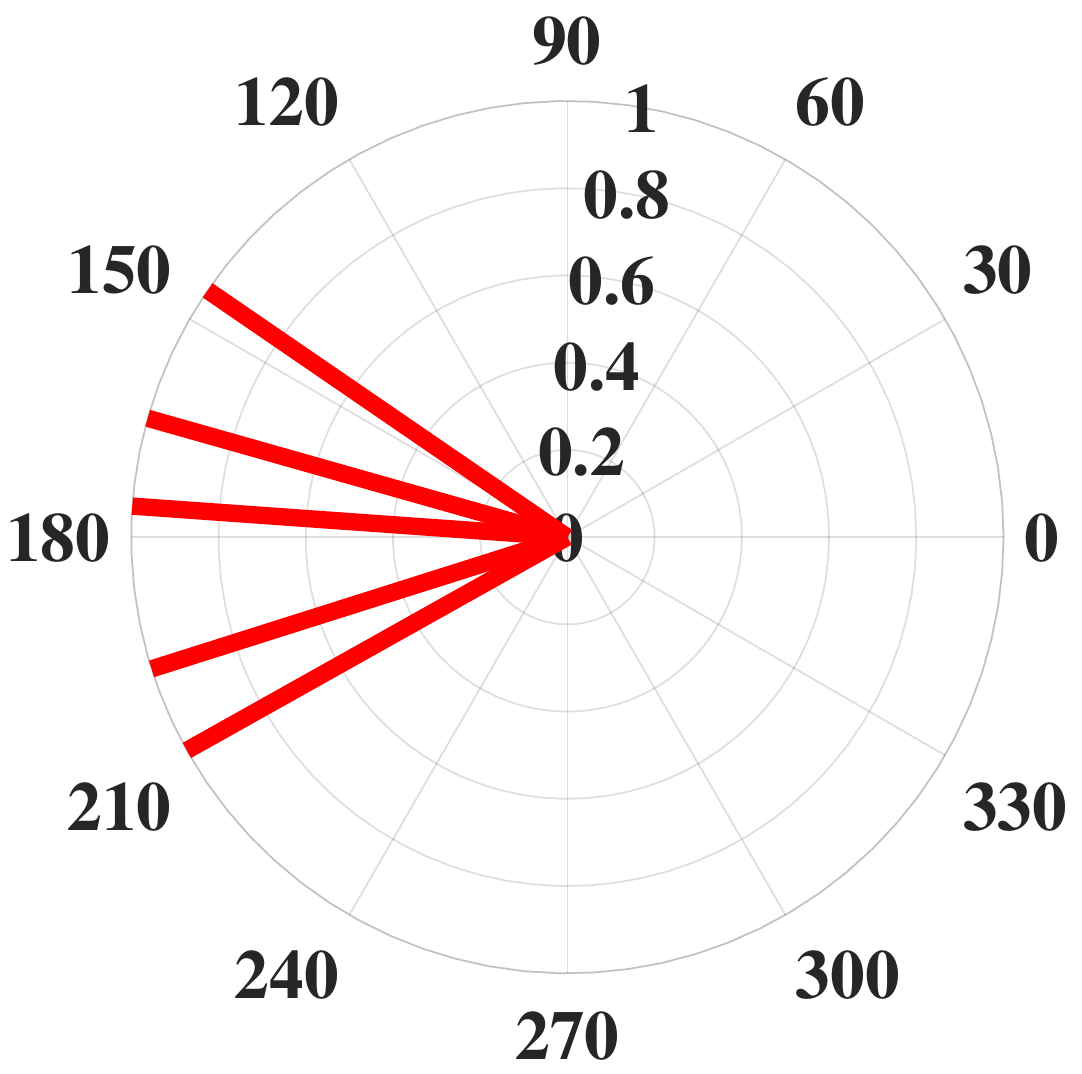}
		\label{CB-3-Estimated-Motion-Direction-510-570-600-630-700}}
	\hfil
	\subfloat[]{\includegraphics[width=0.125\textwidth]{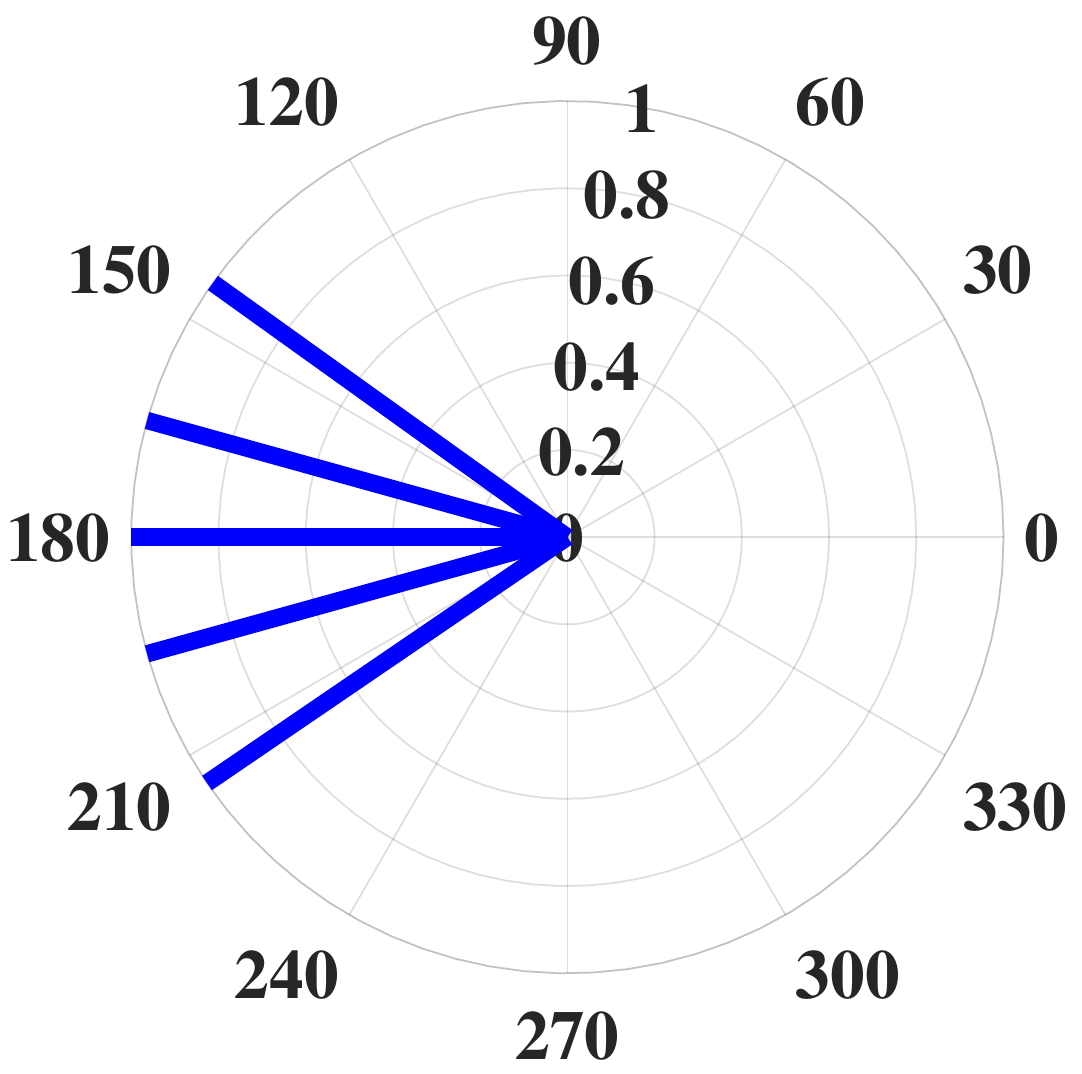}
		\label{CB-3-Actual-Motion-Direction-510-570-600-630-700}}
	\hfil
	\subfloat[]{\includegraphics[width=0.125\textwidth]{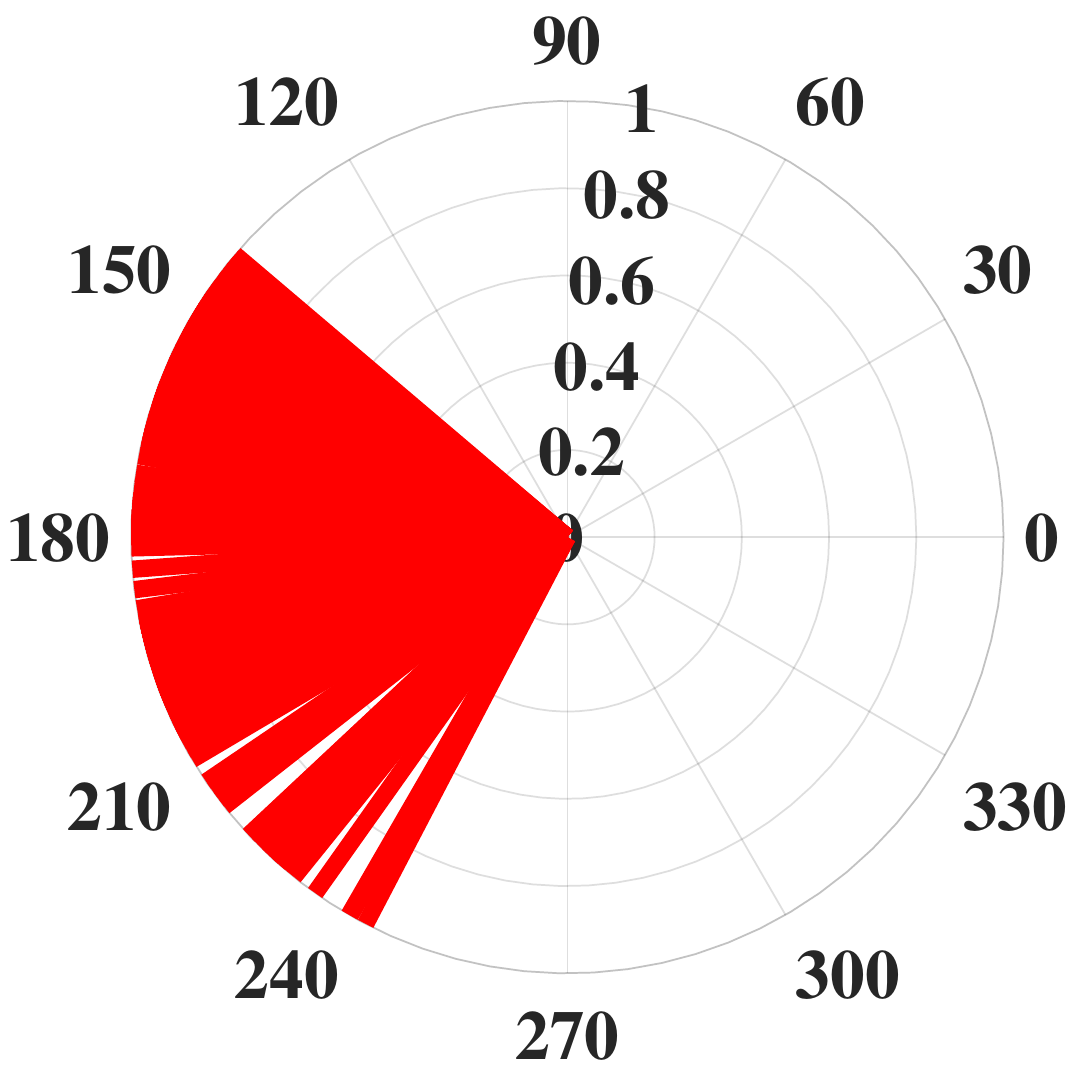}
		\label{CB-3-Estimated-Direction-500-700-1}}
	\hfil
	\subfloat[]{\includegraphics[width=0.125\textwidth]{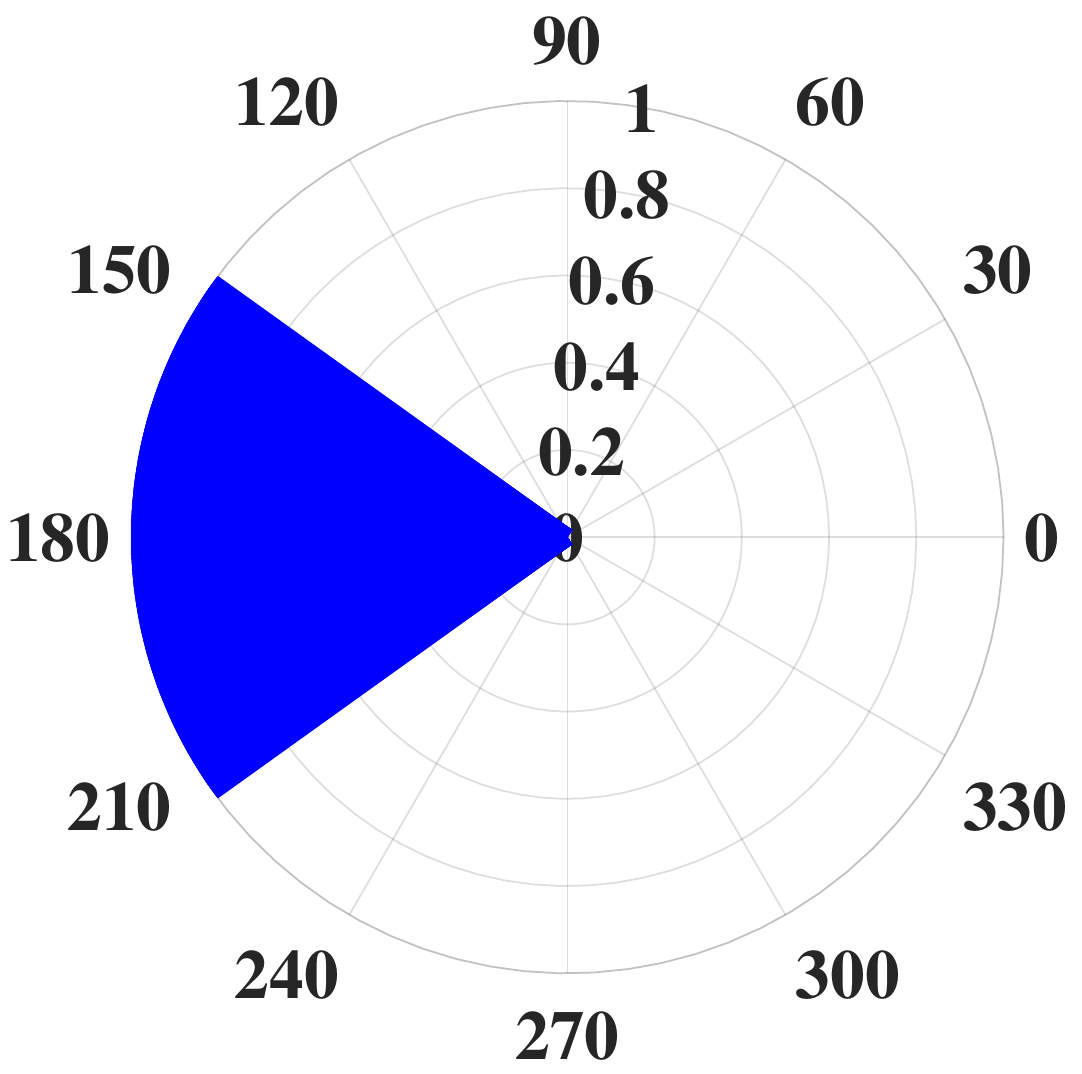}
		\label{CB-3-Actual-Direction-500-700-1}}
	
	\caption{Experiment $5$. (a) Representative frame of the input image sequence. (b) Weber Contrast of the small target during time period $t \in [0, 1000]$ ms. (c) ROC curves of the ESTMD and DSTMD. (d) Motion directions detected by the DSTMD in the sample $510$, $570$, $600$, $630$, $700$ frames. No motion direction detected by the ESTMD. (e) Actual motion directions in the sample $510$, $570$, $600$, $630$, $700$ frames. (f) Motion directions detected by the DSTMD from the $500$th to the $700$th frame. (g) Actual motion directions from the $500$th to the $700$th frame.}
	\label{CB-3-LDTB-DR-FA}
\end{figure*}

In the fourth and fifth experiment, we evaluate the performance of the proposed neural network in  different backgrounds. Two input image sequences with different backgrounds are displayed in Fig. \ref{CB-2-LDTB-DR-FA}(a) and Fig. \ref{CB-3-LDTB-DR-FA}(a), respectively. In these two image sequences, the backgrounds are all moving from left to right and their velocities are set as $250$ pixel/s. A small target whose luminance and size are set as $0$ and $5 \times 5$ pixels, is moving against the cluttered backgrounds. The coordinate of the small target at time $t$ is $(500 - 250 \cdot \frac{t+300}{1000}, 125+15 \cdot \sin(4\pi \frac{t+300}{1000})), t \in [0, 1000]$ ms. 

Fig. \ref{CB-2-LDTB-DR-FA}(c) and Fig. \ref{CB-3-LDTB-DR-FA}(c) demonstrate the ROC curves for the two image sequences, respectively. As can be seen, the detection rates of the DSTMD (or ESTMD) in Fig. \ref{CB-3-LDTB-DR-FA}(c) are much lower than those in Fig. \ref{CB-2-LDTB-DR-FA}(c). There are two reasons for the above result: 1) the background in Fig. \ref{CB-3-LDTB-DR-FA}(a) is more cluttered compared to Fig. \ref{CB-2-LDTB-DR-FA}(a), which means that it contains more small-target-like background features and 2) the Weber Contrast in Fig. \ref{CB-3-LDTB-DR-FA}(b) is much lower than that in Fig. \ref{CB-2-LDTB-DR-FA}(b), suggesting that the models exhibit much weaker outputs to the small target in the fifth experiment. 

Fig. \ref{CB-2-LDTB-DR-FA}(d) displays the motion directions detected by the DSTMD in the sample $510$, $570$, $600$, $630$, $700$ frames while Fig. \ref{CB-2-LDTB-DR-FA}(f) illustrates the motion directions detected by the DSTMD from the $500$th to the $700$th frame. As it is shown, these detected motion directions are quite close to the actual motion directions in Fig. \ref{CB-2-LDTB-DR-FA}(e) and \ref{CB-2-LDTB-DR-FA}(g). No motion direction is detected by the ESTMD, because it is not directionally selective. Similar results can be seen in Fig. \ref{CB-3-LDTB-DR-FA}(d)-(g).

The proposed neural network is further tested on a set of real videos. The experimental results are presented in the supplementary materials, due to the page limit.

\section{Further Discussions}
\label{Further-Discussions}
In the above sections, the presented neural network (DSTMD) demonstrated a reliable ability to detect small targets and motion directions against complex backgrounds. Nowadays, for vision-based mobile robots, their visual sensors are becoming more reliable while computation ability is more powerful. These make it possible for mobile robots, such as unmanned aerial vehicle (UAV), equipped with the presented neural network to detect small moving targets in the distance in the real world.

In the insects' visual system, numerous neurons work together to extract different cues from the real world. For example, the LMCs extract motion information while the amacrine cells capture contrast information from input visual signals \cite{vogt2007first,zheng2006feedback}. Integrating these two types of information may contribute to the improvement of  detection performance of the STMD neurons in cluttered backgrounds. In the future, the cooperation of these specialized neurons needs to be taken into consideration.

A number of bio-inspired neural networks based on firing-rate methods, spiking neural networks or convolutional neural networks \cite{hwu2017self,beyeler2015gpu,browning2009cortical}, have been used for target detection, tracking and navigation. Although these neural networks perform well, they cannot distinguish small target motion from large object motion. Detecting target motion is relatively easy, but distinguishing different target motion in terms of the targets' sizes is more challenging and difficult. For example, a naturally cluttered background always contains small targets such as insects, and large objects such as bushes, trees or rocks. Due to the camera motion, these large objects are moving with the background. In this case, the above-mentioned neural networks can detect both small and large object motion, but cannot distinguish them.

In engineering, small target motion detection can be performed by infrared detection methods \cite{gao2013infrared}. However, these infrared methods always require significant temperature differences between objects of interest (such as rockets and jets) and the background. This largely limits their application, because such significant temperature difference is rare in the natural world. Different from the infrared methods, the presented neural network uses normal images as input and  provides a vision-based method for small moving target detection.

\section{Conclusion}
\label{Conclusion}
In this paper, we proposed a visual neural network (DSTMD) to simulate the directionally selective STMD neurons. Direction selectivity is obtained by correlating signals from two positions while size selectivity is introduced by the second-order lateral inhibition mechanism.  Motion directions of detected targets are estimated by the population vector algorithm. Systematic experiments showed that the presented STMD-based neural network can detect not only small moving targets, but also motion directions against complex backgrounds. In the future work, various visual neurons which extract different cues simultaneously, will be integrated together to improve detection performance.


\section*{Acknowledgment}
The authors would like to thank the reviewers for their valuable comments. They would also like to thank M. Daniels for proofreading this manuscript.

\ifCLASSOPTIONcaptionsoff
  \newpage
\fi


\bibliographystyle{IEEEtran}
\bibliography{IEEEabrv,Reference}

\begin{thebibliography}{10}
\providecommand{\url}[1]{#1}
\csname url@samestyle\endcsname
\providecommand{\newblock}{\relax}
\providecommand{\bibinfo}[2]{#2}
\providecommand{\BIBentrySTDinterwordspacing}{\spaceskip=0pt\relax}
\providecommand{\BIBentryALTinterwordstretchfactor}{4}
\providecommand{\BIBentryALTinterwordspacing}{\spaceskip=\fontdimen2\font plus
\BIBentryALTinterwordstretchfactor\fontdimen3\font minus
  \fontdimen4\font\relax}
\providecommand{\BIBforeignlanguage}[2]{{%
\expandafter\ifx\csname l@#1\endcsname\relax
\typeout{** WARNING: IEEEtran.bst: No hyphenation pattern has been}%
\typeout{** loaded for the language `#1'. Using the pattern for}%
\typeout{** the default language instead.}%
\else
\language=\csname l@#1\endcsname
\fi
#2}}
\providecommand{\BIBdecl}{\relax}
\BIBdecl

\bibitem{yue2006collision}
S.~Yue and F.~C. Rind, ``Collision detection in complex dynamic scenes using an
  lgmd-based visual neural network with feature enhancement,'' \emph{IEEE
  Trans. Neural Netw.}, vol.~17, no.~3, pp. 705--716, May 2006.

\bibitem{nordstrom2006insect}
K.~Nordstr{\"o}m, P.~D. Barnett, and D.~C. O'Carroll, ``Insect detection of
  small targets moving in visual clutter,'' \emph{PLoS Biol.}, vol.~4, no.~3,
  p. e54, Feb. 2006.

\bibitem{olberg2000prey}
R.~Olberg, A.~Worthington, and K.~Venator, ``Prey pursuit and interception in
  dragonflies,'' \emph{J. Comp. Physiol. A}, vol. 186, no.~2, pp. 155--162,
  Feb. 2000.

\bibitem{nordstrom2012neural}
K.~Nordstr{\"o}m, ``Neural specializations for small target detection in
  insects,'' \emph{Curr. Opin. Neurobiol.}, vol.~22, no.~2, pp. 272--278, Apr.
  2012.

\bibitem{nordstrom2006small}
K.~Nordstr{\"o}m and D.~C. O'Carroll, ``Small object detection neurons in
  female hoverflies,'' \emph{Proc. R. Soc. Lond., B, Biol. Sci.}, vol. 273, no.
  1591, pp. 1211--1216, May 2006.

\bibitem{o1993feature}
D.~O'Carroll, ``Feature-detecting neurons in dragonflies,'' \emph{Nature}, vol.
  362, no. 6420, pp. 541--543, Apr. 1993.

\bibitem{barnett2007retinotopic}
P.~D. Barnett, K.~Nordstr{\"o}m, and D.~C. O'Carroll, ``Retinotopic
  organization of small-field-target-detecting neurons in the insect visual
  system,'' \emph{Curr. Biol.}, vol.~17, no.~7, pp. 569--578, Apr. 2007.

\bibitem{gonzalez2013eight}
P.~T. Gonzalez-Bellido, H.~Peng, J.~Yang, A.~P. Georgopoulos, and R.~M. Olberg,
  ``Eight pairs of descending visual neurons in the dragonfly give wing motor
  centers accurate population vector of prey direction,'' \emph{Proc. Natl.
  Acad. Sci. U.S.A.}, vol. 110, no.~2, pp. 696--701, Jan. 2013.

\bibitem{wiederman2008model}
S.~D. Wiederman, P.~A. Shoemaker, and D.~C. O'Carroll, ``A model for the
  detection of moving targets in visual clutter inspired by insect
  physiology,'' \emph{PLoS One}, vol.~3, no.~7, p. e2784, Jul. 2008.

\bibitem{wiederman2013biologically}
S.~D. Wiederman and D.~C. O’Carroll, ``Biologically inspired feature
  detection using cascaded correlations of off and on channels,'' \emph{J.
  Artif. Intell. Soft Comput. Res.}, vol.~3, no.~1, pp. 5--14, Dec. 2013.

\bibitem{bagheri2017autonomous}
Z.~M. Bagheri, B.~S. Cazzolato, S.~Grainger, D.~C. O’Carroll, and S.~D.
  Wiederman, ``An autonomous robot inspired by insect neurophysiology pursues
  moving features in natural environments,'' \emph{J. Neural Eng.}, vol.~14,
  no.~4, p. 046030, Jul. 2017.

\bibitem{bagheri2017performance}
Z.~M. Bagheri, S.~D. Wiederman, B.~S. Cazzolato, S.~Grainger, and D.~C.
  O’Carroll, ``Performance of an insect-inspired target tracker in natural
  conditions,'' \emph{Bioinspir. \& Biomim.}, vol.~12, no.~2, p. 025006, Feb.
  2017.

\bibitem{bagheri2015properties}
Z.~M. Bagheri, S.~D. Wiederman, B.~S. Cazzolato, S.~Grainger, and D.~C.
  O'Carroll, ``Properties of neuronal facilitation that improve target tracking
  in natural pursuit simulations,'' \emph{J. R. Soc. Interface}, vol.~12, no.
  108, p. 20150083, Jun. 2015.

\bibitem{judge1997locust}
S.~Judge and F.~Rind, ``The locust dcmd, a movement-detecting neurone tightly
  tuned to collision trajectories,'' \emph{J. Exp. Biol.}, vol. 200, no.~16,
  pp. 2209--2216, Aug. 1997.

\bibitem{rind1992orthopteran}
F.~C. Rind and P.~J. Simmons, ``Orthopteran dcmd neuron: a reevaluation of
  responses to moving objects. i. selective responses to approaching objects,''
  \emph{J. Neurophysiol.}, vol.~68, no.~5, pp. 1654--1666, Nov. 1992.

\bibitem{simmons1997responses}
P.~J. Simmons and F.~C. Rind, ``Responses to object approach by a wide field
  visual neurone, the lgmd2 of the locust: characterization and image cues,''
  \emph{J. Comp. Physiol. A}, vol. 180, no.~3, pp. 203--214, Feb. 1997.

\bibitem{lee2015spatio}
Y.-J. Lee, H.~O. J{\"o}nsson, and K.~Nordstr{\"o}m, ``Spatio-temporal dynamics
  of impulse responses to figure motion in optic flow neurons,'' \emph{PLoS
  One}, vol.~10, no.~5, pp. 1--16, May 2015.

\bibitem{borst1990direction}
A.~Borst and M.~Egelhaaf, ``Direction selectivity of blowfly motion-sensitive
  neurons is computed in a two-stage process,'' \emph{Proc. Natl. Acad. Sci.
  U.S.A.}, vol.~87, no.~23, pp. 9363--9367, Dec. 1990.

\bibitem{borst1995mechanisms}
A.~Borst, M.~Egelhaaf, and J.~Haag, ``Mechanisms of dendritic integration
  underlying gain control in fly motion-sensitive interneurons,'' \emph{J.
  Comput. Neurosci.}, vol.~2, no.~1, pp. 5--18, Mar. 1995.

\bibitem{rind1996neural}
F.~C. Rind and D.~Bramwell, ``Neural network based on the input organization of
  an identified neuron signaling impending collision,'' \emph{J.
  Neurophysiol.}, vol.~75, no.~3, pp. 967--985, Mar. 1996.

\bibitem{yue2013redundant}
S.~Yue and F.~C. Rind, ``Redundant neural vision systems-competing for
  collision recognition roles,'' \emph{IEEE Trans. Auton. Mental Develop.},
  vol.~5, no.~2, pp. 173--186, Apr. 2013.

\bibitem{yue2006bio}
S.~Yue, F.~C. Rind, M.~S. Keil, J.~Cuadri, and R.~Stafford, ``A bio-inspired
  visual collision detection mechanism for cars: Optimisation of a model of a
  locust neuron to a novel environment,'' \emph{Neurocomputing}, vol.~69,
  no.~13, pp. 1591--1598, Feb. 2006.

\bibitem{LGMD2-BMVC}
Q.~Fu, C.~Hu, and S.~Yue, ``Bio-inspired collision detector with enhanced
  selectivity for ground robotic vision system,'' in \emph{Proc. Brit. Mach.
  Vis. Conf.}, York, U.K., 2016.

\bibitem{hu2016bio}
C.~Hu, F.~Arvin, C.~Xiong, and S.~Yue, ``A bio-inspired embedded vision system
  for autonomous micro-robots: the lgmd case,'' \emph{IEEE Trans. Cogn.
  Develop. Syst.}, vol.~9, no.~3, pp. 241--254, Sep. 2016.

\bibitem{hassenstein1956systemtheoretische}
B.~Hassenstein and W.~Reichardt, ``Systemtheoretische analyse der zeit-,
  reihenfolgen-und vorzeichenauswertung bei der bewegungsperzeption des
  r{\"u}sselk{\"a}fers chlorophanus,'' \emph{Zeitschrift f{\"u}r Naturforschung
  B}, vol.~11, no. 9-10, pp. 513--524, Oct. 1956.

\bibitem{franceschini1989directionally}
N.~Franceschini, A.~Riehle, and A.~Le~Nestour, ``Directionally selective motion
  detection by insect neurons,'' in \emph{Facets of vision}.\hskip 1em plus
  0.5em minus 0.4em\relax Berlin, Heidelberg: Springer, 1989, pp. 360--390.

\bibitem{eichner2011internal}
H.~Eichner, M.~Joesch, B.~Schnell, D.~F. Reiff, and A.~Borst, ``Internal
  structure of the fly elementary motion detector,'' \emph{Neuron}, vol.~70,
  no.~6, pp. 1155--1164, Jun. 2011.

\bibitem{clark2011defining}
D.~A. Clark, L.~Bursztyn, M.~A. Horowitz, M.~J. Schnitzer, and T.~R. Clandinin,
  ``Defining the computational structure of the motion detector in
  drosophila,'' \emph{Neuron}, vol.~70, no.~6, pp. 1165--1177, Jun. 2011.

\bibitem{warrant2016matched}
E.~J. Warrant, ``Matched filtering and the ecology of vision in insects,'' in
  \emph{The Ecology of Animal Senses}.\hskip 1em plus 0.5em minus 0.4em\relax
  Springer, 2016, pp. 143--167.

\bibitem{freifeld2013gabaergic}
L.~Freifeld, D.~A. Clark, M.~J. Schnitzer, M.~A. Horowitz, and T.~R. Clandinin,
  ``Gabaergic lateral interactions tune the early stages of visual processing
  in drosophila,'' \emph{Neuron}, vol.~78, no.~6, pp. 1075--1089, Jun. 2013.

\bibitem{behnia2014processing}
R.~Behnia, D.~A. Clark, A.~G. Carter, T.~R. Clandinin, and C.~Desplan,
  ``Processing properties of on and off pathways for drosophila motion
  detection,'' \emph{Nature}, vol. 512, no. 7515, p. 427, Aug. 2014.

\bibitem{yang2016subcellular}
H.~H. Yang, F.~St-Pierre, X.~Sun, X.~Ding, M.~Z. Lin, and T.~R. Clandinin,
  ``Subcellular imaging of voltage and calcium signals reveals neural
  processing in vivo,'' \emph{Cell}, vol. 166, no.~1, pp. 245--257, Jun. 2016.

\bibitem{bolzon2009local}
D.~M. Bolzon, K.~Nordstr{\"o}m, and D.~C. O'Carroll, ``Local and large-range
  inhibition in feature detection,'' \emph{J. Neurosci.}, vol.~29, no.~45, pp.
  14\,143--14\,150, Nov. 2009.

\bibitem{tuthill2016four}
J.~C. Tuthill and B.~G. Borghuis, ``Four to foxtrot: how visual motion is
  computed in the fly brain,'' \emph{Neuron}, vol.~89, no.~4, pp. 677--680,
  Feb. 2016.

\bibitem{escobar2013mathematical}
M.-J. Escobar, D.~Pezo, and P.~Orio, ``Mathematical analysis and modeling of
  motion direction selectivity in the retina,'' \emph{J. Phys. Paris}, vol.
  107, no.~5, pp. 349--359, Nov. 2013.

\bibitem{straw2008vision}
A.~D. Straw, ``Vision egg: an open-source library for realtime visual stimulus
  generation,'' \emph{Front. Neuroinf.}, vol.~2, p.~4, Nov. 2008.

\bibitem{vogt2007first}
N.~Vogt and C.~Desplan, ``The first steps in drosophila motion detection,''
  \emph{Neuron}, vol.~56, no.~1, pp. 5--7, Oct. 2007.

\bibitem{zheng2006feedback}
L.~Zheng, G.~G. de~Polavieja, V.~Wolfram, M.~H. Asyali, R.~C. Hardie, and
  M.~Juusola, ``Feedback network controls photoreceptor output at the layer of
  first visual synapses in drosophila,'' \emph{J. Gen. Physiol.}, vol. 127,
  no.~5, pp. 495--510, Apr. 2006.

\bibitem{hwu2017self}
T.~Hwu, J.~Isbell, N.~Oros, and J.~Krichmar, ``A self-driving robot using deep
  convolutional neural networks on neuromorphic hardware,'' in \emph{Proc. Int.
  Joint Conf. Neural Netw.}, 2017, pp. 635--641.

\bibitem{beyeler2015gpu}
M.~Beyeler, N.~Oros, N.~Dutt, and J.~L. Krichmar, ``A gpu-accelerated cortical
  neural network model for visually guided robot navigation,'' \emph{Neural
  Netw.}, vol.~72, pp. 75--87, Dec. 2015.

\bibitem{browning2009cortical}
N.~A. Browning, S.~Grossberg, and E.~Mingolla, ``Cortical dynamics of
  navigation and steering in natural scenes: Motion-based object segmentation,
  heading, and obstacle avoidance,'' \emph{Neural Netw.}, vol.~22, no.~10, pp.
  1383--1398, Dec. 2009.

\bibitem{gao2013infrared}
C.~Gao, D.~Meng, Y.~Yang, Y.~Wang, X.~Zhou, and A.~G. Hauptmann, ``Infrared
  patch-image model for small target detection in a single image,'' \emph{IEEE
  Trans. Image Process.}, vol.~22, no.~12, pp. 4996--5009, Sep. 2013.

\end{thebibliography}

\end{document}